\documentclass{article} 
\usepackage[table]{xcolor}

\usepackage{iclr2023_conference,times}

\usepackage{amsmath,amsfonts,bm}

\def\eg{\emph{e.g}\onedot} 
\def\ie{\emph{i.e}\onedot}

\def\etc{\emph{etc}\onedot} \def\vs{\emph{vs}\onedot}

\def\eqref#1{equation~\ref{#1}}

\def\1{\bm{1}}

\def\vs{{\bm{s}}}

\DeclareMathAlphabet{\mathsfit}{\encodingdefault}{\sfdefault}{m}{sl}
\SetMathAlphabet{\mathsfit}{bold}{\encodingdefault}{\sfdefault}{bx}{n}

\definecolor{citecolor}{HTML}{2980b9}
\definecolor{linkcolor}{HTML}{c0392b}
\usepackage[pagebackref=true,breaklinks=true,colorlinks,bookmarks=false,citecolor=citecolor,linkcolor=linkcolor]{hyperref}

\usepackage{url}
\usepackage{subcaption}
\usepackage{arydshln}
\usepackage{graphicx}
 \usepackage[capitalize]{cleveref}

\usepackage{afterpage}
\usepackage{cleveref}
\usepackage{xspace}

\usepackage{enumerate}

\usepackage[ruled,vlined,linesnumbered]{algorithm2e} 

\usepackage{booktabs, siunitx}
\usepackage{authblk} %

\usepackage{multirow}
\usepackage{rotating}
\usepackage{booktabs}
\usepackage{tabularx}
\usepackage{mysymbols}

\title{Unified-IO: A unified model for \\vision, language, and multi-modal tasks}

\author{%
  \textbf{Jiasen Lu$^{\dagger}$\thanks{Equal contribution. Correspondence to \texttt{jiasenl@allenai.org}}, Christopher Clark$^{\dagger*}$, Rowan Zellers$^{\dagger\diamond}$},\\ \textbf{Roozbeh Mottaghi$^{\dagger\diamond}$, Aniruddha Kembhavi$^{\dagger\diamond}$}\\
  \vspace{0.1in}
  $^\dagger$Allen Institute for AI, $^\diamond$University of Washington, Seattle
  }
  
\newcommand{\boldheader}[1]{\noindent\textbf{#1.}}

\newcommand{\uio}{\mbox{\sc{Unified-IO}}}
\newcommand{\vqgan}{\mbox{\sc{VQ-GAN}}}
\newcommand{\dvae}{\mbox{\sc{D-VAE}}}

\usepackage{wrapfig}
\usepackage{array}

\iclrfinalcopy 
\begin{document}

\maketitle
\begin{abstract}
We propose \uio, a model that performs a large variety of AI tasks spanning classical computer vision tasks, including pose estimation, object detection, depth estimation and image generation, vision-and-language tasks such as region captioning and referring expression, to natural language processing tasks such as question answering and paraphrasing. Developing a single unified model for such a large variety of tasks poses unique challenges due to the heterogeneous inputs and outputs pertaining to each task, including RGB images, per-pixel maps, binary masks, bounding boxes, and language. We achieve this unification by homogenizing every supported input and output into a sequence of discrete vocabulary tokens. This common representation across all tasks allows us to train a single transformer-based architecture, jointly on over 90 diverse datasets in the vision and language fields. \uio\ is the first model capable of performing all 7 tasks on the GRIT benchmark and produces strong results across 16 diverse benchmarks like NYUv2-Depth, ImageNet, VQA2.0, OK-VQA, Swig, VizWizGround, BoolQ, and SciTail, with no task-specific fine-tuning. 
Code and demos for \uio\ are available at: \href{https://unified-io.allenai.org}{\texttt{unified-io.allenai.org}}

\end{abstract}

\section{Introduction}



We present \uio, the first neural model to jointly perform a large and diverse set of AI tasks spanning classical computer vision (such as object detection, segmentation, and depth estimation), image synthesis (such as image generation and image in-painting), vision-and-language (like visual question answering, image captioning, and referring expression) and NLP (such as question answering and paraphrasing). Unified general-purpose models avoid the need for task-specific design, learn and perform a wide range of tasks with a single architecture, can utilize large, diverse data corpora, can effectively transfer concept knowledge across tasks, and even perform tasks unknown and unobserved at design and training time.

\begin{figure}[t]
    \centering
    \includegraphics[width=\textwidth]{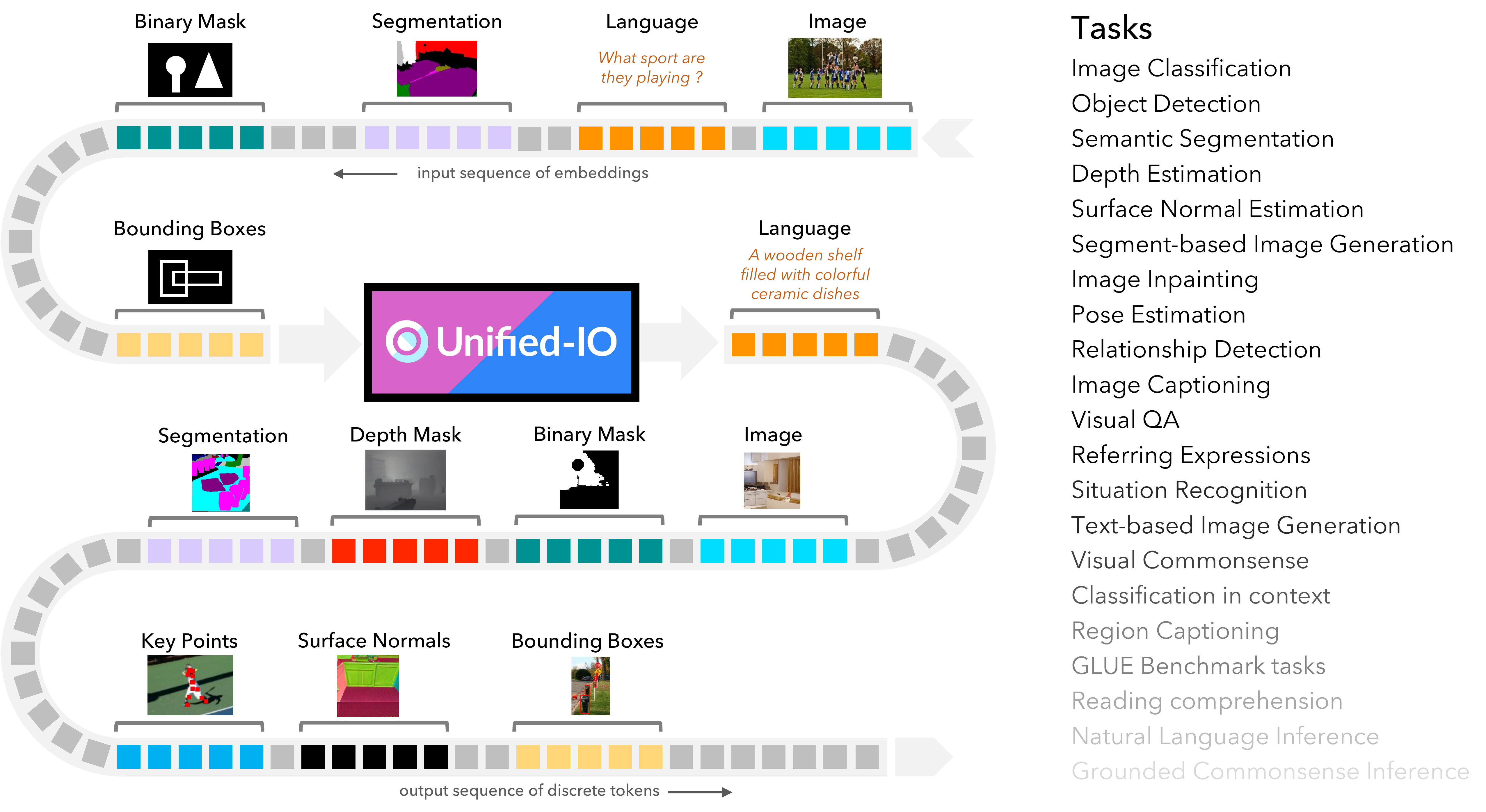}
    \captionof{figure}{\textbf{\uio} is a single sequence-to-sequence model that performs a variety of tasks in computer vision and NLP using a unified architecture without a need for either task or modality-specific branches. This broad unification is achieved by homogenizing every task’s input and output into a sequence of discrete vocabulary tokens. \uio\ supports modalities as diverse as images, masks, keypoints, boxes, and text, and tasks as varied as depth estimation, inpainting, semantic segmentation, captioning, and reading comprehension.}
\end{figure}

Building unified models for computer vision has proven to be quite challenging since vision tasks have incredibly diverse input and output representations. For instance, object detection produces bounding boxes around objects in an image, segmentation produces binary masks outlining regions in an image, visual question answering produces an answer as text, and depth estimation produces a map detailing the distance of each pixel from the camera. This heterogeneity makes it very challenging to architect a single model for all these tasks. In contrast, while the landscape of natural language processing (NLP) tasks, datasets, and benchmarks is large and diverse, their inputs and desired outputs can often be uniformly represented as sequences of tokens. Sequence to sequence (Seq2Seq) architectures~\citep{2020t5,gpt3}, specifically designed to accept and produce such sequences of tokens, are thus widely applicable to many tasks. Unified models employing such architectures have been central to much recent progress in NLP.

Unified models for computer vision typically use a shared visual backbone to produce visual embeddings but then employ individual branches for each of the desired tasks. These include models like Mask R-CNN~\citep{He17Mask} for classical visual tasks that use an ImageNet pre-trained encoder followed by branches for detection and segmentation, trained in a fully supervised manner. In the vision and language (V\&L) domain, CNN backbones feed visual features to transformer architectures that also combine language, followed by task-specific heads for visual question answering, referring expression, visual commonsense reasoning, etc.~\citep{lu2019vilbert,li2019visualbert,tan2019lxmert}. A more recent trend has seen the emergence of unified architectures that do away with task-specific heads and instead introduce modality-specific heads~\citep{hu2021unit,cho2021vlt5,Gupta2021GPV, wang2022OFA} -- for instance, a single language decoder that serves multiple tasks requiring language output like captioning and classification. However, most progress in unified models continues to be centered around V\&L tasks, owing to the simplicity of building shared language decoders, and is often limited to supporting just a handful of tasks.

\uio\ is a Seq2Seq model capable of performing a variety of tasks using a unified architecture without a need for either task or even modality-specific branches. This broad unification is achieved by homogenizing every task's output into a sequence of discrete tokens. Dense structured outputs such as images, segmentation masks and depth maps are converted to sequences using a vector quantization variational auto-encoder (VQ-VAE) \citep{esser2021taming}, sparse structured outputs such as bounding boxes, and human joint locations are transcribed into sequences of coordinate tokens, and language outputs are converted to sequences using byte-pair encoding. This unification enables Unified-IO to jointly train on over 90 datasets spanning computer vision, V\&L, and NLP tasks with a single streamlined transformer encoder-decoder architecture \citep{2020t5}.

Our jointly trained \uio\ is the first model to support all 7 tasks in the General Robust Image Task (GRIT) Benchmark~\citep{gupta2022grit} and obtains the top overall score of 64.3 when averaging across all tasks, handily beating the second best model by 32.0. 
We further evaluate \uio\ on 16 diverse benchmarks across computer vision and NLP, without any fine-tuning towards any individual benchmark, and find that it performs remarkably well compared to specialized (or fine-tuned) state-of-the-art models. 
\section{Vision, Language and Multi-Modal Tasks}
\label{sect:02_tasks}

\begin{table}[t]
\label{tasks}
\renewcommand{\dashlinedash}{2pt}
\renewcommand{\dashlinegap}{2pt}
\newcommand{\band}{\rowcolor{gray!10}}
\newcommand{\checkcol}{p1.5cm}
\newcolumntype{M}{>{\centering\arraybackslash}p{0.9cm}}
\renewcommand{\indent}{\hspace{0.2cm}}
\setlength\tabcolsep{3pt}
\renewcommand{\arraystretch}{1.25}
\center
  \resizebox{\textwidth}{!}{
  \begin{tabular}{l l c c c c:  M M M M  :  M M M M}
  \toprule
  & \multirow{2}{1.3cm}{Example Source} & \multicolumn{4}{c}{Size} & \multicolumn{4}{c}{Input Modalities} & \multicolumn{4}{c}{Output Modalities} \\ 
    \cmidrule(r){3-6}
    \cmidrule(r){7-10}
    \cmidrule(r){11-14}
  & & Datasets & Size & Percent & \multicolumn{1}{c}{Rate}  & Text & Image & Sparse & \multicolumn{1}{c}{Dense} & Text & Image & Sparse & \multicolumn{1}{c}{Dense} \\
  \midrule
\band \textbf{Image Synthesis}  &  & \textbf{14} & \textbf{56m} & \textbf{43.0} & \textbf{18.7} & \checkmark & \checkmark & \checkmark & \checkmark & - & \checkmark & - & - \\
\indent Image Synthesis from Text & \textit{RedCaps} & 9 & 55m & 41.9 & 16.7 & \checkmark & - & - & - & - & \checkmark & - & - \\
\indent Image Inpainting & \textit{VG} & 3 & 1.2m & 0.9 & 1.5 & \checkmark & \checkmark & \checkmark & - & - & \checkmark & - & - \\
\indent Image Synthesis from Seg. & \textit{LVIS} & 2 & 220k & 0.2 & 0.6 & \checkmark & - & - & \checkmark & - & \checkmark & - & - \\
\band \textbf{Sparse Labelling}  &  & \textbf{10} & \textbf{8.2m} & \textbf{6.3} & \textbf{12.5} & \checkmark & \checkmark & \checkmark & - & - & - & \checkmark & - \\
\indent Object Detection & \textit{Open Images} & 3 & 1.9m & 1.5 & 3.6 & - & \checkmark & - & - & - & - & \checkmark & - \\
\indent Object Localization & \textit{VG} & 3 & 6m & 4.6 & 7.1 & \checkmark & \checkmark & - & - & - & - & \checkmark & - \\
\indent Keypoint Estimation & \textit{COCO} & 1 & 140k & 0.1 & 0.7 & - & \checkmark & \checkmark & - & - & - & \checkmark & - \\
\indent Referring Expression & \textit{RefCoco} & 3 & 130k & 0.1 & 1.1 & \checkmark & \checkmark & - & - & - & - & \checkmark & - \\
\band \textbf{Dense Labelling}  &  & \textbf{6} & \textbf{2.4m} & \textbf{1.8} & \textbf{6.2} & \checkmark & \checkmark & - & - & - & - & - & \checkmark \\
\indent Depth Estimation & \textit{NYU Depth} & 1 & 48k & 0.1 & 0.4 & - & \checkmark & - & - & - & - & - & \checkmark \\
\indent Surface Normal Estimation & \textit{Framenet} & 2 & 210k & 0.2 & 1.1 & - & \checkmark & - & - & - & - & - & \checkmark \\
\indent Object Segmentation & \textit{LVIS} & 3 & 2.1m & 1.6 & 4.7 & \checkmark & \checkmark & - & - & - & - & - & \checkmark \\
\band \textbf{Image Classification}  &  & \textbf{9} & \textbf{22m} & \textbf{16.8} & \textbf{12.5} & - & \checkmark & \checkmark & - & \checkmark & - & - & - \\
\indent Image Classification & \textit{ImageNet} & 6 & 16m & 12.2 & 8.1 & \checkmark & \checkmark & - & - & \checkmark & - & - & - \\
\indent Object Categorization & \textit{COCO} & 3 & 6m & 4.6 & 4.4 & - & \checkmark & \checkmark & - & \checkmark & - & - & - \\
\band \textbf{Image Captioning}  &  & \textbf{7} & \textbf{31m} & \textbf{23.7} & \textbf{12.5} & - & \checkmark & \checkmark & - & \checkmark & - & - & - \\
\indent Webly Supervised Captioning & \textit{CC12M} & 3 & 26m & 19.7 & 8.8 & - & \checkmark & - & - & \checkmark & - & - & - \\
\indent Supervised Captioning & \textit{VizWiz} & 3 & 1.4m & 1.1 & 1.7 & - & \checkmark & - & - & \checkmark & - & - & - \\
\indent Region Captioning & \textit{VG} & 1 & 3.8m & 2.9 & 2.0 & - & \checkmark & \checkmark & - & \checkmark & - & - & - \\
\band \textbf{Vision \& Language}  &  & \textbf{16} & \textbf{4m} & \textbf{3.0} & \textbf{12.5} & \checkmark & \checkmark & \checkmark & - & \checkmark & - & - & \checkmark \\
\indent Visual Question Answering & \textit{VQA 2.0} & 13 & 3.3m & 2.5 & 10.4 & \checkmark & \checkmark & \checkmark & - & \checkmark & - & - & - \\
\indent Relationship Detection & \textit{VG} & 2 & 640k & 0.5 & 1.9 & - & \checkmark & \checkmark & - & \checkmark & - & - & - \\
\indent Grounded VQA & \textit{VizWiz} & 1 & 6.5k & 0.1 & 0.1 & \checkmark & \checkmark & - & - & \checkmark & - & - & \checkmark \\
\band \textbf{NLP}  &  & \textbf{31} & \textbf{7.1m} & \textbf{5.4} & \textbf{12.5} & \checkmark & - & - & - & \checkmark & - & - & - \\
\indent Text Classification & \textit{MNLI} & 17 & 1.6m & 1.2 & 4.8 & \checkmark & - & - & - & \checkmark & - & - & - \\
\indent Question Answering & \textit{SQuAD} & 13 & 1.7m & 1.3 & 5.2 & \checkmark & - & - & - & \checkmark & - & - & - \\
\indent Text Summarization & \textit{Gigaword} & 1 & 3.8m & 2.9 & 2.5 & \checkmark & - & - & - & \checkmark & - & - & - \\
\band \textbf{Language Modelling} &  & \textbf{2} & - & - & \textbf{12.5} & \checkmark & - & - & - & \checkmark & - & - & - \\
\indent Masked Language Modelling & \textit{C4} & 2 & - & - & 12.5 & \checkmark & - & - & - & \checkmark & - & - & - \\
\band \textbf{All Tasks} &  & \textbf{95} & \textbf{130m} & \textbf{100} & \textbf{100} & \checkmark & \checkmark & \checkmark & \checkmark & \checkmark & \checkmark & \checkmark & \checkmark \\

 \bottomrule
\end{tabular}}

\caption{\small{
Tasks \uio\ learns to complete. From left to right, columns show an example of one of the sources used for the task, the number of datasets, total number and percent of examples relative to the entire training corpus, and sample rate during multi-task training. 
Subsequent columns show what modalities are required for the tasks, and highlighted rows show aggregated statistics for groups of similar tasks.
}}
\vspace{-0.15in}
\label{tab:tasks}
\end{table}


\uio\ is designed to handle a wide range of language, vision and language, and classic vision tasks in a unified way. To fully test this capability, we gather 95 vision, language, and multi-modal datasets from 62 publicly available data sources as targets for our model to learn during multi-task training. These datasets cover a wide range of tasks, skills, and modalities.

We categorize the input and output modalities of each task into 4 different types: \texttt{Text} -- natural language tokens; \texttt{Image} -- RGB images; \texttt{Sparse} -- a small number of location coordinates within the image; \texttt{Dense} -- per-pixel labels such as depth maps, surface normal maps, \etc. 
We group related datasets into 8 groups and 22 tasks to facilitate our training and analysis:

\boldheader{Image Synthesis} Given a text description, partially occluded image and inpainting target, or segmentation map containing a semantic class for some pixels, generate a matching image. Data sources with image and text pairs~\citep{redcaps}, bounding boxes~\citep{visual_genome} or semantic segmentation~\citep{lvis} can be used to build these tasks.    

\boldheader{Sparse Labelling} Given an image and a natural language query, identify the target regions or keypoint locations that are being referred to. Tasks include object detection \citep{open_images}, object localization \citep{rhodes2017fast}, human pose estimation \citep{coco} and referring expression \citep{referitgame}.

\boldheader{Dense Labelling} Given an image, produce per-pixel labels for that image. Labels include the distance of that pixel to the camera \citep{nyu_depth}, surface orientation \citep{bae2021estimating} or semantic class \citep{coco}. 

\boldheader{Image Classification} Given an image and optionally a target bounding box, generate a class name or tag of that image or target region. This group includes image classification \citep{imagenet_cvpr09} and object categorization \citep{pinz2006object} datasets.

\boldheader{Image Captioning} Given an image and optionally a bounding box, generate a natural language description of that image or target region. We include both crowd-sourced \citep{coco_captions} and webly supervised \citep{cc12cm} captions. 

\boldheader{Vision \& Language} A broad category for other tasks that require jointly reason over image content and a natural language query. There are many popular vision and language datasets, and we categories these datasets into 3 tasks -- visual question answering \citep{VQA}; relationship detection \citep{lu2016visual} and grounded VQA \citep{whiz_viz_answer_grounded_vqa}.

\boldheader{NLP} Tasks with text as the only input and output modalities, including text classification \citep{mnli}, question answering \citep{rajpurkar2016squad} and text summarization \citep{graff2003english}. 

\boldheader{Language Modeling} The masking language modeling pre-training task (See Section~\ref{sect:training}) using text from C4 \citep{2020t5} and Wikipedia \citep{wikidump}, which we include to ensure the knowledge gained from language pre-training is not lost during multi-task training. Other pre-training tasks are not included because the relevant datasets are already used in other supervised tasks (\eg, for captioning or classification).

Table~\ref{tab:tasks} shows the details of tasks and groups. We list an example dataset source, number of datasets, number of examples, percent of the total number of examples, and sampling rate during training (Section \ref{sect:training}) for each group and task. Subsequent columns show what modalities are required for the inputs and outputs. 
We defer additional task details, inference details, the complete list of datasets and visualizations to the Appendix~\ref{sect:task_details}.

\section{Unified-IO}

%
%
%
Our goal is to build a single unified model that can support a diverse set of tasks across computer vision and language with little to no need for task-specific customizations and parameters. Such unified architectures can be applied to new tasks with little to no knowledge of the underlying machinery, enable general pre-training to benefit many diverse downstream applications, be jointly trained on a large number of tasks, and better allows knowledge to be shared between tasks.

\subsection{Unified Task Representations}
\label{sect:task-representation}
Supporting a variety of modalities such as images, language, boxes, binary masks, segmentation masks, \etc without task-specific heads requires representing these modalities in a shared and unified space. To do this, we discretize the text, images, and other structured outputs in our tasks and represent them with tokens drawn from a unified and finite vocabulary.

\begin{figure}
    \centering
    \includegraphics[width=\textwidth]{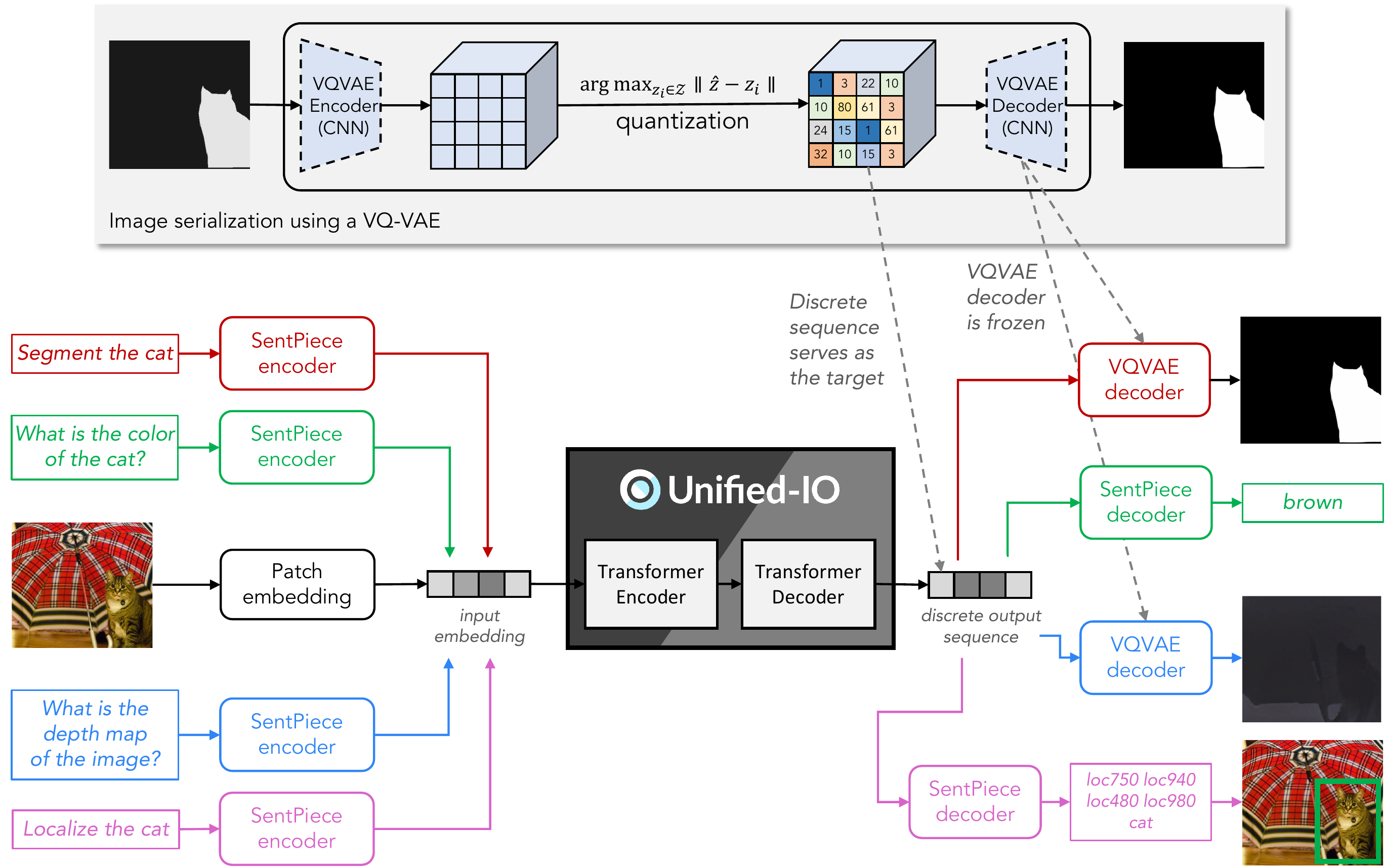}
    \captionof{figure}{\textbf{Unified-IO.} A schematic of the model with four demonstrative tasks: object segmentation, visual question answering, depth estimation and object localization.}
\end{figure}

\boldheader{Text representation} Following \citet{2020t5}, text inputs and outputs are tokenized using SentencePiece \citep{kudo2018sentencepiece}. Following past works such as \cite{McCann2018decaNLP, 2020t5, Gupta2021GPV, wang2022OFA} we also specify each task with a natural language prompt (excluding some tasks like VQA, which are fully specified by their text inputs) in order to indicate what task should be performed. For example, ``\textit{What is the depth map of the image?}'' for depth estimation or ``\textit{What region does ``cat" describe?}'' for object localization.

\boldheader{Images and dense structures representation} A variety of tasks in computer vision requires the model to produce high-dimensional outputs such as images (\eg, image in-painting) or per-pixel labels (\eg, depth estimation). 
To handle these modalities, we first convert per-pixel labels into RGB images. For depth, we construct a grayscale image by normalizing the depth map. For surface normal estimation, we convert the $x/y/z$ orientations into $r/g/b$ values. For segmentation, we map each instance present in the image to a unique color. We randomly select colors for each instance and specify the color-to-class mapping in the text instead of using universal color-to-class mapping. This avoids requiring a fixed list of classes and avoids having colors that may only be marginally different due to the presence of a large number of classes. 

Then we encode these images as discrete tokens using a \vqgan. 
In particular, we use the imagenet-pretrained \vqgan\ from \citet{esser2021taming} with $256 \times 256$ resolution, compression ratio of $16$, and $16384$ codebook size.
The \vqgan\ codebook is added to the vocabulary as additional tokens that can be generated by the decoder. During training, the tokens for the target image are used as targets. During inference, the \vqgan\ decoder is used to convert the generated image tokens into an output image.

\boldheader{Sparse structures representation} 
We encode sparse structures such as bounding boxes or human joints by adding 1000 special tokens to the vocabulary to represent discretized image coordinates~\citep{chen2021pix2seq}. 
Points are then encoded with a sequence of two such tokens, one for the $x$ and one for the $y$ coordinates, and boxes are encoded using a sequence of four tokens, two for the upper right corner and two for the lower left corner. Labeled boxes are encoded as a box followed by a text class label, and joints are encoded as a sequence of points followed by a text visibility label. 
This allows us to handle a wide variety of tasks that use these elements in their inputs or output (see Appendix~\ref{sect:task_details} for examples).




\subsection{Unified Architecture} 

Universally representing a wide variety of tasks as input and output sequences of discrete tokens enables us to employ architectures that have been proven successful in natural language processing. In \uio, we propose a pure transformer model largely following the design of T5 \citep{2020t5}. In particular, \uio~is an encoder-decoder architecture where both the encoder and decoder are composed of stacked transformer layers, which in turn are composed of self-attention transformers, cross-attention transformers (in the decoder), and feed-forward neural networks. The layers are applied residually, and layer norms are applied before each transformer and feed-forward network. See \cite{2020t5} for details.

We make a few architectural changes to adapt the T5 architecture to our setting. First, to handle input images, we reshape the image into a sequence of patches that are embedded with linear projection similar to \cite{dosovitskiy2020vit}. Second, we expand the vocabulary to include the location tokens and the image tokens used in the VQ-GAN. Third, we extend the 1-d relative embedding \citep{dosovitskiy2020vit} to 2-d with a fixed number of learned embeddings. We also add absolute position embedding to the token embedding following \cite{devlin2018bert}, since the absolute position information is essential to image tasks. 

We use a maximum of 256 and 128 text tokens for inputs and outputs respectively, and a maximum length of 576 (\ie $24 \times 24$ patch encoding from a $384 \times 384$ image) for image inputs and 256 (\ie $16 \times 16$ latent codes from a $256 \times 256$ image) for image outputs. 
In this work, we present four versions of \uio~ ranging from 71 million to 2.9 billion parameters, as detailed in Table~\ref{tab:model_size}.

\subsection{Training}
\label{sect:training}

\begin{table}[t]
\center
  \resizebox{1.0\textwidth}{!}{
  \begin{tabular}{l S[table-format=3.0] S[table-format=3.0] S[table-format=3.0] S[table-format=3.0] S[table-format=3.0] S[table-format=3.0]}
\toprule
{Model} & {Encoder Layers} & {Decoder Layers} & {Model Dims} & {MLP Dims} & {Heads} & {Total Params} \\
\midrule
\uio$_{\texttt{SMALL}}$ & 8 & 8 & 512 & 1024 & 6 & 71M\\
\uio$_{\texttt{BASE}}$ & 12 & 12 & 768 & 2048 & 12 & 241M\\
\uio$_{\texttt{LARGE}}$ & 24 & 24 & 1024 & 2816 & 16 & 776M \\
\uio$_{\texttt{XL}}$ & 24 & 24 & 2048 & 5120 & 32 & 2925M \\
\bottomrule
\end{tabular}}
\caption{\small{Size variant of \uio. Both encoder and decoder are based on T5 implementation \citep{2020t5}. Parameters of VQ-GAN \citep{esser2021taming} are not included in the total parameter count.}}
\label{tab:model_size}
\end{table}

\uio~is trained in two stages -- A pre-training stage that uses unsupervised losses from text, image, and paired image-text data, and a massive multi-task stage where the model is jointly trained on a large variety of tasks. Since our goal is to examine whether a single unified model can solve a variety of tasks simultaneously, we \textbf{do not perform task-specific fine-tuning} although prior work \citep{Lu202012in1MV, wang2022OFA} shows it can further improve task performance. 

\boldheader{Pre-training} To learn good representations from large-scale webly supervised image and text data, we consider two pre-training tasks: \textit{text span denoising} and \textit{masked image denoising}. The text span denoising task follows \cite{2020t5} -- randomly corrupt 15\% of the tokens and replace the consecutive corrupted tokens with a unique mask token. The masked image denoising task follows \cite{bao2021beit} and \cite{he2022masked} -- randomly masked 75\% of the image patches, and the goal is to recover the whole image. When another modality is present, \ie image or text, the model can use information from that modality to complete the tasks.

We construct the pre-training dataset by incorporating publicly available language data (i.e., plain texts from Common Crawl), vision data (i.e., raw images from different datasets), and V\&L data (i.e., image caption and image label pairs).
For V\&L data, we add a simple prompt ``\textit{An image of}'' at the beginning of caption or categories to indicate it is multi-modal data \citep{wang2021simvlm}. 

We pre-train \uio\ on this combination of datasets with an in-batch mixing strategy. We equally sample data with the text and image denoising objective. For text denoising, half of the samples are from pure text data, \ie C4 and Wikipedia. The other half is constructed from image and class data, such as Imagenet21k \citep{ridnik2021imagenet} or image and caption data, such as YFCC15M \citep{radford2021learning}. For image denoising, we also use the same caption and class data and some image-only data from datasets for our vision tasks. We sample from datasets in proportion to dataset size. See Appendix~\ref{sect:pretraining_distribution} for details.

\boldheader{Multi-tasking} 
To build a single unified model for diverse vision, language, and V\&L tasks, we construct a massive multi-tasking dataset by ensembling 95 datasets from 62 publicly available data sources. See Section~\ref{sect:02_tasks} for task details and Appendix~\ref{sect:task_details} for dataset visualizations. 

We jointly train \uio\ on this large set of datasets by mixing examples from these datasets within each batch. We equally sample each group ($1/8$) except for image synthesis ($3/16$) and dense labeling ($1/16$) since dense labeling has significantly fewer data and image synthesis has significantly more data than other groups. 
Within each group, we sample datasets proportional to the square root of their size to better expose the model to underrepresented tasks. Due to the large variance in dataset size, some tasks are still rarely sampled (\eg  depth estimation only has a $0.43\%$ chance of being sampled). See Appendix~\ref{sect:multi_task_distribution} for details and visualizations.

\subsection{Implementation Details}
The total vocabulary size is 49536, with 32152 language tokens, 1000 location tokens, and 16384 vision tokens. 
During training, we random sub-sample 128 image patches for pre-training state and 256 image patches (out of 576) for multi-task stage. We do not use dropout. Adafactor \citep{shazeer2018adafactor} optimizer is used to save memory. We use a learning rate of $10^{-2}$ for the first 10,000 steps and then decay at a rate of $1/\sqrt{k}$. We train with $\beta_1 = 0.9$ and $\beta_2 = 1.0-k^{-0.8}$, where $k$ is the step number. We use global norm gradient clipping with 1.0 and find this is crucial to stabilized \texttt{XL} training. 
We train the \texttt{Small}, \texttt{Base} and \texttt{Large} with batch size of 2048 and \texttt{XL} with batch size of 1024 due to memory consideration. 4-way in-layer parallelism and 128-way data parallelism used to scale the 3B model training.
For all models, we train $1000k$ steps -- $500k$ for pre-training and multi-task training respectively. 

\section{Experiments}

We now present results for \uio\ on the GRIT benchmark (Sec \ref{sect:main_results}), evaluation on same concept and new concept (Sec \ref{appendix:same_new_eval}), ablate training data via the GRIT ablation benchmark (Sec \ref{sect:ablation}) and evaluate \uio\ on 16 other benchmarks in computer vision and NLP (Sec \ref{sect:other-results}). Section \ref{appendix:prompt_generalization} shows the prompt generalization on refer expression. Qualitative examples are in Appendix \ref{sect:qual_examples}.

\subsection{Results on GRIT}
\label{sect:main_results}

The General Robust Image Task (GRIT) Benchmark~\citep{gupta2022grit} is an evaluation-only benchmark designed to measure the performance of models across multiple tasks, concepts, and data sources. 
GRIT aims to encourage the building of unified and general purpose vision models and is thus well suited to evaluate \uio. 
GRIT has seven tasks that cover a range of visual skills with varying input and output modalities and formats: categorization, localization, VQA, refer expression, segmentation, keypoint, and surface normal estimation. 

\uio~is the first model to support all seven tasks in GRIT. As seen in Table~\ref{tab:grit_results}, \uio$_\texttt{XL}$ outperforms all prior submissions to GRIT obtaining average accuracy of 64.3 on test. The next best submission is GPV-2~\citep{Kamath2022WeblySC} which obtains 32.0 and can only support 4 out of 7 tasks.
%
\uio$_\texttt{XL}$ also outperforms the multi-task checkpoint of OFA$_\texttt{LARGE}$ \citep{wang2022OFA} on VQA, refer expression and categorization. 

\newcommand{\band}{\rowcolor{gray!10}}
\begin{table}[tb]
\setlength\tabcolsep{3pt}
\renewcommand{\arraystretch}{1.25}
\center
  \resizebox{\textwidth}{!}{
  \begin{tabular}{c l @{\hspace{1.0\tabcolsep}} c  c c c c c  c c  c c c c c c  c c c c c}
\toprule
 &  & \multicolumn{2}{c}{Categorization} &  \multicolumn{2}{c}{Localization} & \multicolumn{2}{c}{VQA} & \multicolumn{2}{c}{Refexp} & \multicolumn{2}{c}{Segmentation} & \multicolumn{2}{c}{Keypoint} & \multicolumn{2}{c}{Normal} & \multicolumn{2}{c}{All} \\
  \cmidrule(r){3-4}
  \cmidrule(r){5-6}
  \cmidrule(r){7-8}
  \cmidrule(r){9-10}
  \cmidrule(r){11-12}
  \cmidrule(r){13-14}
  \cmidrule(r){15-16}
  \cmidrule(r){17-18}
 & & ablation & test & ablation & test & ablation & test & ablation & test & ablation & test & ablation & test & ablation & test & ablation & test \\
 \midrule
    \band \small\texttt{0} & NLL-AngMF [\citenum{bae2021estimating}] & - & - & - & - & - & - & - & - & - & - & - & - & \textbf{49.6} & \textbf{50.5} & 7.2 & 7.1 \\
 \small\texttt{1} & Mask R-CNN [\citenum{He17Mask}] & - & - & 44.7 & 45.1 & - & - & - & - & 26.2 & 26.2 & \textbf{70.8} & \textbf{70.6} & - & - & 20.2 & 20.3 \\
    \band \small\texttt{2} & GPV-1 [\citenum{Gupta2021GPV}] & 33.2 & 33.2 & 42.8 & 42.7 & 50.6 & 49.8 & 25.8 & 26.8 & - & - & - & - & - & - & 21.8 & 21.8 \\
   \small\texttt{3} & CLIP [\citenum{radford2021learning}]& 48.1 & - & - & - & - & - & - & - & - & - & - & - & - &- & 6.9 & - \\
  \band \small\texttt{4} & OFA$_{\texttt{LARGE}}$ [\citenum{wang2022OFA}] & 22.6 & - & - & - & 72.4 & - & 61.7 & - & - & - & - & - & - & - & 22.4 & - \\ 
   \small\texttt{5} & GPV-2 [\citenum{Kamath2022WeblySC}] & 54.7 & 55.1 & 53.6 & 53.6 & 63.5 & 63.2 & 51.5 & 52.1 & - & - & - & - & - & - & 31.9 & 32.0 \\
 \midrule
 \band \small\texttt{6} &\uio$_{\texttt{SMALL}}$ & 42.6 & - & 50.4 & - & 52.9 & - & 51.1 & - & 40.7 & - & 46.5 & - & 33.5 & - & 45.4 & - \\
   \small\texttt{7} &\uio$_{\texttt{BASE}}$ & 53.1 & - & 59.7 & - & 63.0 & - & 68.3 & - & 49.3 & - & 60.2 & - & 37.5 & - & 55.9 & - \\
\band \small\texttt{8} &\uio$_{\texttt{LARGE}}$ & 57.0 & - & 64.2 & - & 67.4 & - & 74.1 & - & 54.0 & - & 67.6 & - & 40.2 & - & 60.7 & -  \\
  \small\texttt{9} &\uio$_{\texttt{XL}}$ & \textbf{61.7} & \textbf{60.8} & \textbf{67.0} & \textbf{67.1} & \textbf{74.5} & \textbf{74.5} & \textbf{78.6} & \textbf{78.9} & \textbf{56.3} & \textbf{56.5} & 68.1 & 67.7 & 45.0 & 44.3 & \textbf{64.5} & \textbf{64.3} \\
\bottomrule
\end{tabular}}
\caption{\small{Comparison of our \uio~models to recent SOTA on GRIT benchmark. \uio~is the first model to support all seven tasks in GRIT. Results of CLIP, OFA obtained from GRIT challenge.}}
\label{tab:grit_results}
\end{table}

Mask R-CNN \citep{He17Mask} is a strong baseline for core vision tasks.
\uio$_\texttt{XL}$ outperforms Mask R-CNN on localization and segmentation. The reason is \uio$_\texttt{XL}$ shows little degradation in performance between same concept and new concept as discussed in Appendix \ref{appendix:same_new_eval}. On keypoint, our model is worse compared to Mask R-CNN (68.1 \vs 70.8). The reason is we have 2-stage inference for keypoint -- first locate the person using the object localization prompt, then find keypoints for each detected region. 

NLL-AngMF~\citep{bae2021estimating} is a SOTA model for surface normal estimation. 
Our model gets strong results compared to NLL-AngMF (44.3 \vs 49.6). 
Since our image tokenizer is only pre-trained on ImageNet without any surface normal data, the upper bound of our method through reconstruction is 59.8 on FrameNet~\citep{referitgame}. This suggests our score could be considerably improved by training a stronger image tokenizer.

\subsection{Evaluation on same concept and new concept}
\label{appendix:same_new_eval}

\begin{table}[h]
\setlength\tabcolsep{4pt}
\renewcommand{\arraystretch}{1.25}
\center
  \resizebox{\textwidth}{!}{
  \begin{tabular}{c l @{\hspace{1.0\tabcolsep}} c  c c c c c  c c  c c c c c c  c c c c c}
\toprule
 &  & \multirow{2}[3]{*}{restricted} & \multirow{2}[3]{*}{params (M)} & \multicolumn{2}{c}{Categorization} &  \multicolumn{2}{c}{Localization} & \multicolumn{2}{c}{VQA} & \multicolumn{2}{c}{Refexp} & \multicolumn{2}{c}{Segmentation} & \multicolumn{2}{c}{Keypoint} & \multicolumn{2}{c}{Normal} \\
  \cmidrule(r){5-6}
  \cmidrule(r){7-8}
  \cmidrule(r){9-10}
  \cmidrule(r){11-12}
  \cmidrule(r){13-14}
  \cmidrule(r){15-16}
  \cmidrule(r){17-18}
 & &  &  & same & new & same & new & same & new & same & new & same & new & same & new & same & new \\
 \midrule
 \band \small\texttt{0} & NLL-AngMF & \checkmark & 72 & - & - & - & - & - & - & - & - & - & - & - & - & \textbf{50.7} & - \\
  \small\texttt{1} & Mask R-CNN  & \checkmark & 58 & - & - & 51.9 & 40.8 & - & - & - & - & 44.9 & 	
0.3 & \textbf{70.9} & - & - & - \\
 \band  \small\texttt{2} & GPV-1 & \checkmark & 236 & 58.7 & 0.8 & 48.3 & 37.8 & 58.4 & 74.0 & 29.7 & 23.1 & - & - & - & - & - & - \\
 \small\texttt{3} & CLIP  &  & 302 & 49.1 & 46.7 & - & - & - & - & - & - & - & - & - & - & - & - \\
 \band \small\texttt{4} & OFA$_{\texttt{LARGE}}$  & & 473 & 28.9 & 15.8 & - & - & 74.9 & 88.6 & 63.4 & 58.5 & - & - & - & - & - & - \\ 
 \small\texttt{5} & GPV-2 & & 370 & \textbf{85.0} & 13.5 & 54.6 & 54.2 & 69.8 & 81.7 & 57.8 & 48.3 & - & - & - & - & - & - \\
 \midrule
\band \small\texttt{6} & \uio$_{\texttt{SMALL}}$ & & 71 & 52.9 & 31.9 & 47.5 & 61.5 & 59.0 & 72.5 & 54.2 & 45.7 & 37.4 & 48.5 & 46.6 & - & 33.6 & - \\
 \small\texttt{7} &\uio$_{\texttt{BASE}}$ & & 241 & 60.3 & 47.5 & 57.9 & 68.4 & 68.0 & 81.8 & 72.5 & 62.2 & 45.8 & 57.2 & 60.2 & - & 37.7 & - \\
 \band \small\texttt{8} & \uio$_{\texttt{LARGE}}$ & & 776 & 63.0 & 52.7 & 63.3 & 70.9 & 72.1 & 84.3 & 79.2 & 66.3 & 50.4 & 62.2 & 67.7 & - & 40.3 & - \\
\small\texttt{9} & \uio$_{\texttt{XL}}$ & & 2925 & 66.1 & \textbf{60.1} & \textbf{65.6} & \textbf{74.4} & \textbf{78.6} &\textbf{ 90.2} & \textbf{83.5} & \textbf{72.4} & \textbf{53.0} & \textbf{64.2} & 68.2 & - & 45.1 & - \\
\bottomrule
\end{tabular}}
\caption{\small{Generalization to new concepts on the GRIT ablation set. 
}}
\label{tab:grit_concept}
\end{table}

GRIT provides a breakdown of metrics into two groups: \emph{same} for samples that only contain concepts seen in the primary training data (a set of common datasets like COCO, ImageNet and Visual Genome), and \emph{new} for samples containing at least one concept unseen in primary training data. Table~\ref{tab:grit_concept} shows results for \uio\ and other leaderboard entries for the ablation set, divided into same and new concepts.

\uio$_\texttt{XL}$ shows little degradation in performance between \emph{same} and \emph{new}, compared to competing entries. On some tasks \uio\ is even able to outperform on the \emph{new} split compared to the \emph{same}. This indicates that the volume of training data used to train \uio\ has a broad coverage of concepts, and provides almost as effective a level of supervision as provided by large standard vision datasets like COCO. Furthermore, since \uio\ is a uniquely unified architecture with no task-specific parameters, it is very likely able to effectively transfer knowledge across different tasks.

In comparison to Mask-RCNN (row 1), GRIT metrics show \uio\ (row 14) is better by a large margin on \emph{new} concepts, i.e., non-COCO examples (74.4 vs 40.8 for localization and 64.2 vs 0.3 on segmentation), but is still superior on the COCO-like examples (65.6 vs 51.9 for localization and 53.0 vs 44.9 on segmentation). \uio\ is also able to beat GPV-2 (row 5) on \emph{new} concepts by large margins across all 4 tasks supported by GPV-2 even though GPV-2 is exposed to these concepts via webly supervised data and is designed to transfer concept knowledge across skills.

\subsection{Ablations on Task Group}
\label{sect:ablation}

\begin{table}[t]
\setlength\tabcolsep{10pt}
\newcommand{\without}[1]{w/o #1}
\renewcommand{\arraystretch}{1.25}
\center
  \resizebox{1.0\textwidth}{!}{
  \begin{tabular}{l c c c c c c c c c   c}
\toprule
Model & Step & Categorization & Localization & VQA & Refexp & Segmentation & Keypoint & Normal & MNLI \\
\midrule
\uio$_{\texttt{LARGE}}$ & 250k & 50.3 & 63.4 & 65.7 & 73.4 & 51.8 & 69.2 & 40.7 & 85.1\\
\midrule
\without{Image Synthesis} & 200k & 52.7 & 62.9 & 64.2 & 72.0 & 53.6 & 18.3 & \textbf{42.2} & 84.3\\
\without{Sparse} & 220k & 52.6 & - & 64.1 & - & 51.3 & - & 38.5 & 83.8\\
\without{Dense} & 235k & 49.5 & 62.4 & 65.6 & 72.9 & - & 66.7 & - &  84.8\\
\without{Classification} & 220k & - & 63.1 & 64.0 & 73.7 & 52.1 & 66.8 & 39.1 & 84.6 \\
\without{Captioning} & 220k & 49.7 & \textbf{65.0} & \textbf{68.0} & \textbf{74.7} & \textbf{54.2} & 67.4 & 39.2 & \textbf{85.3}\\
\without{V\&L} & 220k & 50.9 & - & - & 72.5 & 51.9 & 70.0 & 38.2 &  84.4\\
\without{NLP} & 220k & \textbf{56.1} & 64.3 & 65.9 & 74.6 & 52.0 & 69.3 & 39.9 & - \\ 
\without{Language Modelling} & 220k & 52.9 & 64.7 & 66.7 & \textbf{74.7} & 52.7 & \textbf{70.2} & 39.9 & 83.5 \\
\bottomrule
\end{tabular}}
\caption{\small{Ablation study on holding out tasks groups and evaluating on GRIT and MNLI~\citep{mnli}}}
\label{tab:grit_ablations}
\end{table}

To better understand how multi-tasking affects learning, we perform ablations by leaving out individual task groups from multi-task training. Due to computational constraints, we ablate \uio${_\texttt{LARGE}}$ and train for 250k steps. 
If ablating a task group, we reduce the number of training steps so that all models are trained on approximately the same number of examples for each of the remaining task groups. 
Results are shown in Table~\ref{tab:grit_ablations} on GRIT and MNLI \citep{mnli}. 

In spite of supporting a large number of heterogeneous tasks, Unified-IO is able to perform well across all tasks. Reducing this heterogeneity by removing task groups does not impact the 
performance of individual tasks significantly.
This is notable since removing a task group significantly reduces the scope of what a model needs to learn while keeping the model capacity fixed. This empirically demonstrates the effectiveness of the proposed unified architecture for massive heterogeneous task support. 

An exception is that removing the NLP group significantly boosts categorization, which might indicate that the sentence classification task interferes with image classification. Removing captioning also boosts performances on VQA and a few other tasks, which might be caused by captioning requiring a relatively large amount of model capacity to learn free-form text generation, in contrast to VQA that requires short answer phrases from a limited vocabulary. Removing image synthesis causes a major regression in keypoint. Manual inspection shows that the model predicts standing-human shaped keypoints even for people in very different postures, suggesting the model learned to rely on priors instead of the image content. We also see minor regressions in localization and referring expression, suggesting that image synthesis tasks, possibly image in-painting in particular, had a surprising positive transfer to understanding sparse structured outputs. It is possible that an ablation analysis on the \texttt{XL} model may yield different outcomes, but we are unable to perform an \texttt{XL}-based analysis due to limited compute.

\subsection{Results on Additional Tasks}
\label{sect:other-results}
\begin{table}[tb]
\setlength\tabcolsep{3pt}
\renewcommand{\arraystretch}{1.25}
\center
  \resizebox{\textwidth}{!}{
  \begin{tabular}{l c  c c c c c  c c  c c c c c c  c c c c c}
\toprule
 & \rotatebox{75}{\raisebox{0.5pt}{NYUv2}}
 & \rotatebox{75}{\raisebox{0.5pt}{ImageNet}} 
 & \rotatebox{75}{\raisebox{0.5pt}{Place365}} 
 & \rotatebox{75}{\raisebox{0.5pt}{VQAv2}} 
 & \rotatebox{75}{\raisebox{0.5pt}{OkVQA}} 
 & \rotatebox{75}{\raisebox{0.5pt}{A-OkVQA}} 
 & \rotatebox{75}{\raisebox{0.5pt}{VizWizQA}} 
 & \rotatebox{75}{\raisebox{0.5pt}{VizWizG}} 
 & \rotatebox{75}{\raisebox{0.5pt}{Swig}} 
 & \rotatebox{75}{\raisebox{0.5pt}{SNLI-VE}} 
 & \rotatebox{75}{\raisebox{0.5pt}{VisComet}} 
 & \rotatebox{75}{\raisebox{0.5pt}{Nocaps}}
 & \rotatebox{75}{\raisebox{0.5pt}{COCO}} 
 & \rotatebox{75}{\raisebox{0.5pt}{COCO}} 
 & \rotatebox{75}{\raisebox{0.5pt}{MRPC}} 
 & \rotatebox{75}{\raisebox{0.5pt}{BoolQ}} 
 & \rotatebox{75}{\raisebox{0.5pt}{SciTail}} 

 \\

 \midrule
 Split & val & val & val & test-dev & test & test & test-dev & test-std & test & val & val & val & val & test & val & val & test\\
 Metric & RMSE & Acc. & Acc. & Acc. & Acc. & Acc. & Acc. & IOU & Acc. & Acc. & CIDEr & CIDEr & CIDEr & CIDEr & F1 & Acc & Acc\\
 \midrule 
 Unified SOTA & UViM & - & - & - & Flamingo & - & Flamingo & - & - & - & - & - & - & - & T5 & PaLM & - \\
   & 0.467 & - & - & - & 57.8 & - & 49.8 & - & - & - & - & - & - & - & 92.20 & 92.2 & - \\
 \midrule
 \uio$_{\texttt{SMALL}}$ & 0.649 & 42.8 & 38.2 & 57.7 & 31.0 & 24.3 & 42.4 & 35.5 & 17.3 & 76.5 & - & 45.1  & 80.1 & - & 84.9 & 65.9 & 87.4\\
 \uio$_{\texttt{BASE}}$ & 0.469 & 63.3 & 43.2 & 61.8 & 37.8 & 28.5 & 45.8 & 50.0 & 29.7 & 85.6 & - & 66.9 & 104.0 & - & 87.9 & 70.8 & 90.8 \\
 \uio$_{\texttt{LARGE}}$ & 0.402 & 71.8 & 50.5 & 67.8 & 42.7 & 33.4 & 47.7 & 54.7 & 40.4 & 86.1 & - & 87.2 & 117.5 & - & 87.5 & 73.1 & 93.1\\
 \uio$_{\texttt{XL}}$ & 0.385 & 79.1 & 53.2 & 77.9 & 54.0 & 45.2 & 57.4 & 65.0 & 49.8 & 91.1 & 21.2 & 100.0 & 126.8 & 122.3 & 89.2 & 79.7 & 95.7\\
 \midrule
 Single or fine- & BinsFormer & CoCa & MAE & CoCa & KAT & GPV2 & Flamingo & MAC-Caps & JSL & OFA & SVT & CoCa & - & OFA & Turing NLR & ST-MOE & DeBERTa  \\
 tuned SOTA  & 0.330 & 91.00 & 60.3 & 82.3 & 54.4 & 38.1 & 65.7 & 27.3 & 39.6 & 91.0 & 18.3 & 122.4 & - & 145.3 & 93.8 & 92.4 & 97.7 \\

\bottomrule
\end{tabular}
}
\caption{\small{Comparing the jointly trained \uio\ to specialized and benchmark fine-tuned state of the art models across Vision, V\&L and Language tasks. Benchmarks used for evaluation are: 
NYUv2~\citep{nyu_depth}, ImageNet~\citep{imagenet_cvpr09}, Places365 \citep{zhou2017places}, VQA 2.0~\citep{balanced_vqa_v2}, A-OKVQA~\citep{schwenk2022okvqa}, VizWizVQA~\citep{vizwiz_vqa}, VizWizG~\citep{whiz_viz_answer_grounded_vqa}, Swig~\citep{pratt2020grounded}, SNLI-VE~\citep{xie2019visual},
VisComet~\citep{park2020visualcomet}, Nocaps~\citep{agrawal2019nocaps}, COCO Captions~\citep{coco_captions}, MRPC~\citep{mrpc}, BoolQ~\citep{clark2019boolq}, and SciTail~\citep{khot2018scitail}.
}}
\label{tab:multi_task_1}

\end{table}

We report results on 16 additional tasks used in our training setup. For these tasks, we do not expect to get state-of-the-art results since specialized models are usually designed and hyper-parameter tuned for a single task, while we are evaluating a single jointly trained model. We also avoid extensive task-specific tricks like color jittering, horizontal flipping, CIDEr optimization, and label smoothing, which are often responsible for considerable gains in individual task performance. We leave such task-specific tuning for future work.
See Table~\ref{tab:multi_task_1} for the results. When possible, we additionally report the best prior result on these tasks from a unified model, meaning a model that is trained in a multi-task setting and a unified architecture (no task-specific head or customizations) with at least three other tasks.

\uio~provides strong performance on all these tasks despite being massively multi-tasked. We review more fine-grained results below.

\boldheader{Depth Estimation} On depth estimation, \uio~achieves 0.385 rmse, which is behind SOTA~\citep{li2022binsformer} but ahead of the recently proposed unified model, UViM~\citep{kolesnikov2022uvim}, despite being trained to do far more tasks.

\boldheader{Image Classification} \uio~achieves 79.1 on ImageNet and 53.2 on Places365, showing the model was able to retain the knowledge of many fine-grained classes despite being massively multi-tasked. Notably, we achieve this without the extensive data augmentations methods typically used by SOTA models~\citep{yu2022coca,he2022masked}.

\boldheader{Visual Question Answering} \uio~is competitive with fine-tuned models on VQA~\citep{flamingo, Kamath2022WeblySC,gui2021kat}, and achieves SOTA results on A-OKVQA. Relative to Flamingo, \uio~performs better on VizWiz-QA but worse on OK-VQA.

\boldheader{Image Captioning} Despite the lack of CIDEr optimization, \uio~is a strong captioning model and generalizes well to nocaps. Since \uio~is trained on many captioning datasets, it is likely the use of style tags following~\cite{cornia2021universal} would offer additional improvement by signaling \uio~to specifically generate COCO-style captions during inference.

\boldheader{NLP tasks}: \uio~achieves respectable results on three NLP tasks but lags behind SOTA models~\citep{smith2022using,zoph2022designing,he2020deberta}. This can partly be attributed to scale. Modern NLP models contain 100 billion+ parameters and with more extensive NLP pre-training.

\subsection{Prompt Generalization Case Study}
\label{appendix:prompt_generalization}
\begin{table}[tb]
\setlength\tabcolsep{10pt}
\newcommand{\without}[1]{w/o #1}
\renewcommand{\arraystretch}{1.25}
\center
  \resizebox{0.7\textwidth}{!}{
  \begin{tabular}{c l c}
 \toprule
 & \textbf{Prompt} & \textbf{Refexp Score} \\
 \midrule
 \small\texttt{0} & Which region does the text `` \texttt{REFEXP} " describe ? & 78.9 \\
 \midrule
 \small\texttt{1} & Which region does the text ``\texttt{REFEXP}" describe? & 76.7\\
 \small\texttt{2} & Which region matches the text `` \texttt{REFEXP} " ? & 77.4 \\
 \small\texttt{3} & Locate the `` \texttt{REFEXP} " . & 65.6 \\
 \small\texttt{4} & Which region can be described as `` \texttt{REFEXP} " ? & 64.8 \\
 \small\texttt{5} & Locate the region described by `` \texttt{REFEXP} " . & 43.2 \\
 \small\texttt{6} & Where is the `` \texttt{REFEXP} " ? & 41.5 \\
 \small\texttt{7} & Where is the ``\texttt{REFEXP}"? & 0.1 \\

\bottomrule
\end{tabular}}
\caption{Case study on GRIT referring expressions using different prompts. The first prompt is the one used during training, the others are paraphrases. \texttt{REFEXP} is replaced by the referring expression text of individual examples during evaluation.}
\label{tab:refexp_prompting}
\end{table}
To better understand how different prompts affect \uio, we do a case study on referring expressions. In particular, we re-evaluate \uio\ on the GRIT referring expression ablation set while replacing the prompt used during training (first row in the table) with a paraphrase (following rows). Results are shown in Table~\ref{tab:refexp_prompting}. 

Overall, we find that the model has some capacity to generalize to paraphrases of the prompt (\eg, row 3 works reasonably well despite using completely different words), but there are paraphrases that result in very significant performance decrease (\eg rows 5, 6, and 8). We also find removing the spaces around the punctuation sometimes results in minor regressions (row 0 vs row 1) and sometimes in sharply reduced performance (row 6 vs row 7), showing \uio\ can be sensitive to formatting details. We hypothesize that this caused by the SentencePiece tokenizer changing the tokenization of the referring expressing if the quotes are not separated from it by spaces.
Building multi-task models that can generalize to different prompts, and ideally to prompts for completely new tasks, is an exciting avenue for future work.

\subsection{Limitations}

%

For object detection, while \uio\ generally produces accurate outputs (see Appendix~\ref{sect:qual_examples}), we find the recall is often poor in cluttered images. Prior work~\citep{chen2021pix2seq} has shown this can be overcome with extensive data augmentation techniques, but these methods are not currently integrated into \uio.
Our use of a pre-trained VQ-GAN greatly simplifies our training and is surprisingly effective for dense prediction tasks. However, it does mean \uio~has limited image generation capabilities (recent works~\citep{parti} have shown this method can be greatly improved but was not available at the time of development).
We also found in a small-scale study that our model does not always understand prompts not in the training data (see Appendix~\ref{appendix:prompt_generalization}).

\section{Related Work}

Vision and language pre-training has become standard practice for multi-modal models, including unified and non-unified models requiring task-specific heads to train from scratch during fine-tuning. Many initial pre-training strategies were inspired by BERT~\citep{devlin2018bert} and included masked-language-modeling, image-text-matching, or mask-region-modeling objectives, often supplemented with objectives using the predictions of a strong object detector model (e.g, VILBERT~\citep{lu2019vilbert}, LXMERT~\citep{tan2019lxmert}, VisualBERT~\citep{li2019visualbert}). More recently contrastive-image-text losses~\citep{radford2021learning,li2022blip,li2021align} or auto-regressive generation losses~\citep{wang2021simvlm,wang2022git,yu2022coca}, have become common. Several works have also directly used object detection or segmentation datasets for pre-training~\cite{yuan2021florence,wang2022OFA,sun2022gppf}. The generalized masked-data-modeling pretraining objective used in \uio\ is similar to ones used in several recent works~\citep{wang2022image,beitv2,singh2022flava}.

Constructing models that can learn to solve many different tasks has been of long-standing interest to researchers. A traditional approach to this problem is to build models with task-specialized heads on top of shared backbones~\citep{He17Mask, liu2019mt-dnn, Lu202012in1MV}. However, this requires manually designing a specialized head for each task, potentially limiting transfer across tasks. An alternative is to build \textit{unified} models -- models that can complete many different tasks without task-specialized components. In NLP, this approach has achieved great success using pre-trained generative models~\citep{2020t5,gpt3,chowdhery2022palm}. 

Inspired by this success, there has been a recent trend to build unified models that can apply to tasks with visual or structured inputs and outputs. Many models have been proposed for tasks with text and/or image input and text output~\citep{cho2021vlt5,wang2021simvlm,li2022blip,wang2021vlmo,kaiser2017one,sun2022gppf,chen2022pali,wang2022image}. However, these models can not produce any structured or visual output. 

More recent unified models can additionally support image locations, which allows tasks like object detection or region captioning. This can be done by using bounding boxes proposed by an object detector~\citep{cho2021vlt5,Kamath2022WeblySC} or including a bounding box output head~\citep{Gupta2021GPV,dou2022coarse,chen2022unified,kamath2021mdetr,li2022grounded}. Alternatively, image locations can be encoded as special tokens in the input/output text~\citep{Yang2021UniTABUT,yao2022pevl,zhu2022seqtr} following~\cite{chen2021pix2seq}. \uio\ follows this design but applies it to a wider set of tasks than previous works, including keypoint estimation, image in-painting, and region captioning.

Some recent unified models have extended these capabilities in other directions~\citep{li2022lavender,li2022mplug,wang2022git,flamingo, jaegle2021perceiver, gato, liang2022highmmt}. Gato \citep{gato} supports additional modalities, including button presses in Atari games or joint torques for
a robot arm, and Flamingo \citep{flamingo} supports interleaved sequences of text, images,
and videos as input. However, neither of these models support image outputs or image location
references limiting the computer vision tasks they can support. Perceiver-IO \citep{jaegle2021perceiver} supports a range of
modalities and proposes a non-auto-regressive decoding approach using task-specific latent query
vectors. While effective for some tasks, this method is not as effective as
auto-regressive decoding on classic generative tasks like captioning or image generation. Uni-Perceiver \citep{zhu2022uni} also supports images, text, and videos and shows good zero-shot performance but does not support generative tasks.

Concurrent to our work, OFA~\citep{wang2022OFA}
proposes a similar approach that also supports image locations and text-to-image synthesis. However, OFA does not support dense labeling tasks such as depth estimation, segmentation, and surface normal estimation. 
Other closely related models include UViM~\citep{kolesnikov2022uvim} which generates a discrete guiding code for a \dvae\ to build an autoregressive model for panoptic segmentation, depth prediction, and colorization, and Pix2Seq v2 \citep{chen2022unified} which extends Pix2Seq to segmentation, keypoint estimation, and image captioning.
\uio\ covers all these tasks and more and focuses on multi-tasking rather than task-specific fine-tuning.

\section{Conclusion}
We have presented \uio, a unified architecture that supports a large variety of computer vision and NLP tasks with diverse inputs and outputs, including images, continuous maps, binary masks, segmentation masks, text, bounding boxes, and keypoints. This unification is made possible by homogenizing each of these modalities into a sequence of discrete tokens. The 2.9B parameter \uio\ XL model is jointly trained on 90+ datasets, is the first model to perform all 7 tasks on the GRIT benchmark and obtains impressive results across 16 other vision and NLP benchmarks, with no benchmark fine-tuning or task-specific modifications.

\section*{Acknowledgements}
\small{Research supported with Cloud TPUs from Google's TPU Research Cloud (TRC). We thank Zak Stone and the Google Cloud TPU team for providing access to the TPU machines used for conducting experiments and Keisuke Sakaguchi for providing early feedback on the project. We thank Amita Kamath for conducting the GRIT referring expression experiment using different prompts. We also thank the ReVIZ team at the Allen Institute for AI, including Sam Stuesser, Sam Skjonsberg, Jon Borchardt, Carissa Schoenick and Michael Schmitz for helping setup the demo website: \href{https://unified-io.allenai.org}{\texttt{unified-io.allenai.org}}}

{\small
\bibliography{iclr2023_conference}

\begin{thebibliography}{130}
\providecommand{\natexlab}[1]{#1}
\providecommand{\url}[1]{\texttt{#1}}
\expandafter\ifx\csname urlstyle\endcsname\relax
  \providecommand{\doi}[1]{doi: #1}\else
  \providecommand{\doi}{doi: \begingroup \urlstyle{rm}\Url}\fi

\bibitem[Agrawal et~al.(2019)Agrawal, Desai, Wang, Chen, Jain, Johnson, Batra,
  Parikh, Lee, and Anderson]{agrawal2019nocaps}
Harsh Agrawal, Karan Desai, Yufei Wang, Xinlei Chen, Rishabh Jain, Mark
  Johnson, Dhruv Batra, Devi Parikh, Stefan Lee, and Peter Anderson.
\newblock nocaps: novel object captioning at scale.
\newblock In \emph{ICCV}, 2019.

\bibitem[Alayrac et~al.(2022)Alayrac, Donahue, Luc, Miech, Barr, Hasson, Lenc,
  Mensch, Millican, Reynolds, Ring, Rutherford, Cabi, Han, Gong, Samangooei,
  Monteiro, Menick, Borgeaud, Brock, Nematzadeh, Sharifzadeh, Binkowski,
  Barreira, Vinyals, Zisserman, and Simonyan]{flamingo}
Jean-Baptiste Alayrac, Jeff Donahue, Pauline Luc, Antoine Miech, Iain Barr,
  Yana Hasson, Karel Lenc, Arthur Mensch, Katie Millican, Malcolm Reynolds,
  Roman Ring, Eliza Rutherford, Serkan Cabi, Tengda Han, Zhitao Gong, Sina
  Samangooei, Marianne Monteiro, Jacob Menick, Sebastian Borgeaud, Andrew
  Brock, Aida Nematzadeh, Sahand Sharifzadeh, Mikolaj Binkowski, Ricardo
  Barreira, Oriol Vinyals, Andrew Zisserman, and Karen Simonyan.
\newblock Flamingo: a visual language model for few-shot learning.
\newblock \emph{arXiv}, 2022.

\bibitem[Antol et~al.(2015)Antol, Agrawal, Lu, Mitchell, Batra, Zitnick, and
  Parikh]{VQA}
Stanislaw Antol, Aishwarya Agrawal, Jiasen Lu, Margaret Mitchell, Dhruv Batra,
  C.~Lawrence Zitnick, and Devi Parikh.
\newblock {VQA}: {V}isual {Q}uestion {A}nswering.
\newblock In \emph{International Conference on Computer Vision (ICCV)}, 2015.

\bibitem[Bae et~al.(2021)Bae, Budvytis, and Cipolla]{bae2021estimating}
Gwangbin Bae, Ignas Budvytis, and Roberto Cipolla.
\newblock Estimating and exploiting the aleatoric uncertainty in surface normal
  estimation.
\newblock In \emph{ICCV}, 2021.

\bibitem[Bao et~al.(2022)Bao, Dong, and Wei]{bao2021beit}
Hangbo Bao, Li~Dong, and Furu Wei.
\newblock {BEiT}: {BERT} pre-training of image transformers.
\newblock In \emph{ICLR}, 2022.

\bibitem[Bar-Haim et~al.(2006)Bar-Haim, Dagan, Dolan, Ferro, Giampiccolo,
  Magnini, and Szpektor]{rte2}
Roy Bar-Haim, Ido Dagan, Bill Dolan, Lisa Ferro, Danilo Giampiccolo, Bernardo
  Magnini, and Idan Szpektor.
\newblock The second pascal recognising textual entailment challenge.
\newblock In \emph{Proceedings of the second PASCAL challenges workshop on
  recognising textual entailment}, volume~6, pp.\  6--4. Venice, 2006.

\bibitem[Bentivogli et~al.(2009)Bentivogli, Clark, Dagan, and
  Giampiccolo]{rte5}
Luisa Bentivogli, Peter Clark, Ido Dagan, and Danilo Giampiccolo.
\newblock The fifth pascal recognizing textual entailment challenge.
\newblock In \emph{TAC}, 2009.

\bibitem[Bowman et~al.(2015)Bowman, Angeli, Potts, and Manning]{bowman2015snli}
Samuel~R Bowman, Gabor Angeli, Christopher Potts, and Christopher~D Manning.
\newblock The {SNLI} corpus.
\newblock 2015.

\bibitem[Brown et~al.(2020)Brown, Mann, Ryder, Subbiah, Kaplan, Dhariwal,
  Neelakantan, Shyam, Sastry, Askell, Agarwal, Herbert{-}Voss, Krueger,
  Henighan, Child, Ramesh, Ziegler, Wu, Winter, Hesse, Chen, Sigler, Litwin,
  Gray, Chess, Clark, Berner, McCandlish, Radford, Sutskever, and Amodei]{gpt3}
Tom~B. Brown, Benjamin Mann, Nick Ryder, Melanie Subbiah, Jared Kaplan,
  Prafulla Dhariwal, Arvind Neelakantan, Pranav Shyam, Girish Sastry, Amanda
  Askell, Sandhini Agarwal, Ariel Herbert{-}Voss, Gretchen Krueger, Tom
  Henighan, Rewon Child, Aditya Ramesh, Daniel~M. Ziegler, Jeffrey Wu, Clemens
  Winter, Christopher Hesse, Mark Chen, Eric Sigler, Mateusz Litwin, Scott
  Gray, Benjamin Chess, Jack Clark, Christopher Berner, Sam McCandlish, Alec
  Radford, Ilya Sutskever, and Dario Amodei.
\newblock Language models are few-shot learners.
\newblock In \emph{NeurIPS}, 2020.

\bibitem[Cer et~al.(2017)Cer, Diab, Agirre, Lopez-Gazpio, and Specia]{stsb}
Daniel Cer, Mona Diab, Eneko Agirre, Inigo Lopez-Gazpio, and Lucia Specia.
\newblock Semeval-2017 task 1: Semantic textual similarity-multilingual and
  cross-lingual focused evaluation.
\newblock \emph{arXiv preprint arXiv:1708.00055}, 2017.

\bibitem[Changpinyo et~al.(2021)Changpinyo, Sharma, Ding, and Soricut]{cc12cm}
Soravit Changpinyo, Piyush Sharma, Nan Ding, and Radu Soricut.
\newblock Conceptual 12m: Pushing web-scale image-text pre-training to
  recognize long-tail visual concepts.
\newblock In \emph{CVPR}, 2021.

\bibitem[Chen et~al.(2022{\natexlab{a}})Chen, Anjum, and
  Gurari]{whiz_viz_answer_grounded_vqa}
Chongyan Chen, Samreen Anjum, and Danna Gurari.
\newblock Grounding answers for visual questions asked by visually impaired
  people.
\newblock In \emph{CVPR}, 2022{\natexlab{a}}.

\bibitem[Chen et~al.(2022{\natexlab{b}})Chen, Saxena, Li, Fleet, and
  Hinton]{chen2021pix2seq}
Ting Chen, Saurabh Saxena, Lala Li, David~J Fleet, and Geoffrey Hinton.
\newblock Pix2seq: A language modeling framework for object detection.
\newblock In \emph{ICLR}, 2022{\natexlab{b}}.

\bibitem[Chen et~al.(2022{\natexlab{c}})Chen, Saxena, Li, Lin, Fleet, and
  Hinton]{chen2022unified}
Ting Chen, Saurabh Saxena, Lala Li, Tsung-Yi Lin, David~J Fleet, and Geoffrey
  Hinton.
\newblock A unified sequence interface for vision tasks.
\newblock \emph{arXiv preprint arXiv:2206.07669}, 2022{\natexlab{c}}.

\bibitem[Chen et~al.(2022{\natexlab{d}})Chen, Wang, Changpinyo, Piergiovanni,
  Padlewski, Salz, Goodman, Grycner, Mustafa, Beyer, et~al.]{chen2022pali}
Xi~Chen, Xiao Wang, Soravit Changpinyo, AJ~Piergiovanni, Piotr Padlewski,
  Daniel Salz, Sebastian Goodman, Adam Grycner, Basil Mustafa, Lucas Beyer,
  et~al.
\newblock Pali: A jointly-scaled multilingual language-image model.
\newblock \emph{arXiv preprint arXiv:2209.06794}, 2022{\natexlab{d}}.

\bibitem[Chen et~al.(2015)Chen, Fang, Lin, Vedantam, Gupta, Doll{\'a}r, and
  Zitnick]{coco_captions}
Xinlei Chen, Hao Fang, Tsung-Yi Lin, Ramakrishna Vedantam, Saurabh Gupta, Piotr
  Doll{\'a}r, and C~Lawrence Zitnick.
\newblock Microsoft coco captions: Data collection and evaluation server.
\newblock \emph{arXiv}, 2015.

\bibitem[Cho et~al.(2021)Cho, Lei, Tan, and Bansal]{cho2021vlt5}
Jaemin Cho, Jie Lei, Hao Tan, and Mohit Bansal.
\newblock Unifying vision-and-language tasks via text generation.
\newblock In \emph{ICML}, 2021.

\bibitem[Chowdhery et~al.(2022)Chowdhery, Narang, Devlin, Bosma, Mishra,
  Roberts, Barham, Chung, Sutton, Gehrmann, et~al.]{chowdhery2022palm}
Aakanksha Chowdhery, Sharan Narang, Jacob Devlin, Maarten Bosma, Gaurav Mishra,
  Adam Roberts, Paul Barham, Hyung~Won Chung, Charles Sutton, Sebastian
  Gehrmann, et~al.
\newblock Palm: Scaling language modeling with pathways.
\newblock \emph{arXiv preprint arXiv:2204.02311}, 2022.

\bibitem[Clark et~al.(2019)Clark, Lee, Chang, Kwiatkowski, Collins, and
  Toutanova]{clark2019boolq}
Christopher Clark, Kenton Lee, Ming-Wei Chang, Tom Kwiatkowski, Michael
  Collins, and Kristina Toutanova.
\newblock Boolq: Exploring the surprising difficulty of natural yes/no
  questions.
\newblock In \emph{NAACL}, 2019.

\bibitem[Cornia et~al.(2021)Cornia, Baraldi, Fiameni, and
  Cucchiara]{cornia2021universal}
Marcella Cornia, Lorenzo Baraldi, Giuseppe Fiameni, and Rita Cucchiara.
\newblock Universal captioner: long-tail vision-and-language model training
  through content-style separation.
\newblock \emph{arXiv}, 2021.

\bibitem[Dagan et~al.(2005)Dagan, Glickman, and Magnini]{rte1}
Ido Dagan, Oren Glickman, and Bernardo Magnini.
\newblock The pascal recognising textual entailment challenge.
\newblock In \emph{Machine Learning Challenges Workshop}, pp.\  177--190.
  Springer, 2005.

\bibitem[De~Marneff et~al.(2019)De~Marneff, Simons, and
  Tonhauser]{commitment_bank}
Marie-Catherine De~Marneff, Mandy Simons, and Judith Tonhauser.
\newblock The commitmentbank: Investigating projection in naturally occurring
  discourse.
\newblock \emph{proceedings of Sinn und Bedeutung 23}, 2019.

\bibitem[Deng et~al.(2009)Deng, Dong, Socher, Li, Li, and
  Fei-Fei]{imagenet_cvpr09}
J.~Deng, W.~Dong, R.~Socher, L.-J. Li, K.~Li, and L.~Fei-Fei.
\newblock {I}mage{N}et: A large-scale hierarchical image database.
\newblock In \emph{CVPR}, 2009.

\bibitem[Desai et~al.(2021)Desai, Kaul, Aysola, and Johnson]{redcaps}
Karan Desai, Gaurav Kaul, Zubin Aysola, and Justin Johnson.
\newblock {R}ed{C}aps: Web-curated image-text data created by the people, for
  the people.
\newblock \emph{arXiv}, 2021.

\bibitem[Devlin et~al.(2019)Devlin, Chang, Lee, and Toutanova]{devlin2018bert}
Jacob Devlin, Ming-Wei Chang, Kenton Lee, and Kristina Toutanova.
\newblock {BERT}: Pre-training of deep bidirectional transformers for language
  understanding.
\newblock In \emph{NAACL}, 2019.

\bibitem[Dolan \& Brockett(2005)Dolan and Brockett]{mrpc}
William~B. Dolan and Chris Brockett.
\newblock Automatically constructing a corpus of sentential paraphrases.
\newblock In \emph{IWP}, 2005.

\bibitem[Dosovitskiy et~al.(2021)Dosovitskiy, Beyer, Kolesnikov, Weissenborn,
  Zhai, Unterthiner, Dehghani, Minderer, Heigold, Gelly, Uszkoreit, and
  Houlsby]{dosovitskiy2020vit}
Alexey Dosovitskiy, Lucas Beyer, Alexander Kolesnikov, Dirk Weissenborn,
  Xiaohua Zhai, Thomas Unterthiner, Mostafa Dehghani, Matthias Minderer, Georg
  Heigold, Sylvain Gelly, Jakob Uszkoreit, and Neil Houlsby.
\newblock An image is worth 16x16 words: Transformers for image recognition at
  scale.
\newblock In \emph{ICLR}, 2021.

\bibitem[Dou et~al.(2022)Dou, Kamath, Gan, Zhang, Wang, Li, Liu, Liu, LeCun,
  Peng, et~al.]{dou2022coarse}
Zi-Yi Dou, Aishwarya Kamath, Zhe Gan, Pengchuan Zhang, Jianfeng Wang, Linjie
  Li, Zicheng Liu, Ce~Liu, Yann LeCun, Nanyun Peng, et~al.
\newblock Coarse-to-fine vision-language pre-training with fusion in the
  backbone.
\newblock \emph{arXiv preprint arXiv:2206.07643}, 2022.

\bibitem[Dunn et~al.(2017)Dunn, Sagun, Higgins, Guney, Cirik, and
  Cho]{dunn2017searchqa}
Matthew Dunn, Levent Sagun, Mike Higgins, V~Ugur Guney, Volkan Cirik, and
  Kyunghyun Cho.
\newblock Searchqa: A new q\&a dataset augmented with context from a search
  engine.
\newblock \emph{arXiv preprint arXiv:1704.05179}, 2017.

\bibitem[Esser et~al.(2021)Esser, Rombach, and Ommer]{esser2021taming}
Patrick Esser, Robin Rombach, and Bjorn Ommer.
\newblock Taming transformers for high-resolution image synthesis.
\newblock In \emph{CVPR}, 2021.

\bibitem[Fisch et~al.(2019)Fisch, Talmor, Jia, Seo, Choi, and
  Chen]{fisch2019mrqa}
Adam Fisch, Alon Talmor, Robin Jia, Minjoon Seo, Eunsol Choi, and Danqi Chen.
\newblock Mrqa 2019 shared task: Evaluating generalization in reading
  comprehension.
\newblock \emph{arXiv preprint arXiv:1910.09753}, 2019.

\bibitem[Foundation()]{wikidump}
Wikimedia Foundation.
\newblock Wikimedia downloads.
\newblock URL \url{https://dumps.wikimedia.org}.

\bibitem[Giampiccolo et~al.(2007)Giampiccolo, Magnini, Dagan, and Dolan]{rte3}
Danilo Giampiccolo, Bernardo Magnini, Ido Dagan, and Bill Dolan.
\newblock The third pascal recognizing textual entailment challenge.
\newblock In \emph{Proceedings of the ACL-PASCAL workshop on textual entailment
  and paraphrasing}, pp.\  1--9. Association for Computational Linguistics,
  2007.

\bibitem[Goyal et~al.(2017)Goyal, Khot, Summers{-}Stay, Batra, and
  Parikh]{balanced_vqa_v2}
Yash Goyal, Tejas Khot, Douglas Summers{-}Stay, Dhruv Batra, and Devi Parikh.
\newblock Making the {V} in {VQA} matter: Elevating the role of image
  understanding in visual question answering.
\newblock In \emph{CVPR}, 2017.

\bibitem[Graff et~al.(2003)Graff, Kong, Chen, and Maeda]{graff2003english}
David Graff, Junbo Kong, Ke~Chen, and Kazuaki Maeda.
\newblock English gigaword.
\newblock \emph{Linguistic Data Consortium, Philadelphia}, 4\penalty0
  (1):\penalty0 34, 2003.

\bibitem[Gui et~al.(2021)Gui, Wang, Huang, Hauptmann, Bisk, and
  Gao]{gui2021kat}
Liangke Gui, Borui Wang, Qiuyuan Huang, Alex Hauptmann, Yonatan Bisk, and
  Jianfeng Gao.
\newblock Kat: A knowledge augmented transformer for vision-and-language.
\newblock \emph{arXiv}, 2021.

\bibitem[Gupta et~al.(2019)Gupta, Dollar, and Girshick]{lvis}
Agrim Gupta, Piotr Dollar, and Ross Girshick.
\newblock {LVIS}: A dataset for large vocabulary instance segmentation.
\newblock In \emph{CVPR}, 2019.

\bibitem[Gupta et~al.(2022{\natexlab{a}})Gupta, Kamath, Kembhavi, and
  Hoiem]{Gupta2021GPV}
Tanmay Gupta, Amita Kamath, Aniruddha Kembhavi, and Derek Hoiem.
\newblock Towards general purpose vision systems: An end-to-end task-agnostic
  vision-language architecture.
\newblock In \emph{CVPR}, 2022{\natexlab{a}}.

\bibitem[Gupta et~al.(2022{\natexlab{b}})Gupta, Marten, Kembhavi, and
  Hoiem]{gupta2022grit}
Tanmay Gupta, Ryan Marten, Aniruddha Kembhavi, and Derek Hoiem.
\newblock {GRIT}: General robust image task benchmark.
\newblock \emph{arXiv}, 2022{\natexlab{b}}.

\bibitem[Gurari et~al.(2018)Gurari, Li, Stangl, Guo, Lin, Grauman, Luo, and
  Bigham]{vizwiz_vqa}
Danna Gurari, Qing Li, Abigale~J Stangl, Anhong Guo, Chi Lin, Kristen Grauman,
  Jiebo Luo, and Jeffrey~P Bigham.
\newblock Vizwiz grand challenge: Answering visual questions from blind people.
\newblock In \emph{CVPR}, 2018.

\bibitem[He et~al.(2017)He, Gkioxari, Dollar, and Girshick]{He17Mask}
Kaiming He, Georgia Gkioxari, Piotr Dollar, and Ross Girshick.
\newblock Mask {R-CNN}.
\newblock In \emph{ICCV}, 2017.

\bibitem[He et~al.(2022)He, Chen, Xie, Li, Doll{\'a}r, and
  Girshick]{he2022masked}
Kaiming He, Xinlei Chen, Saining Xie, Yanghao Li, Piotr Doll{\'a}r, and Ross
  Girshick.
\newblock Masked autoencoders are scalable vision learners.
\newblock In \emph{CVPR}, 2022.

\bibitem[He et~al.(2021)He, Liu, Gao, and Chen]{he2020deberta}
Pengcheng He, Xiaodong Liu, Jianfeng Gao, and Weizhu Chen.
\newblock {DeBERTa}: Decoding-enhanced bert with disentangled attention.
\newblock In \emph{ICLR}, 2021.

\bibitem[Hu \& Singh(2021)Hu and Singh]{hu2021unit}
Ronghang Hu and Amanpreet Singh.
\newblock Unit: Multimodal multitask learning with a unified transformer.
\newblock In \emph{Proceedings of the IEEE/CVF International Conference on
  Computer Vision}, pp.\  1439--1449, 2021.

\bibitem[Huang et~al.(2019{\natexlab{a}})Huang, Zhou, Funkhouser, and
  Guibas]{huang2019framenet}
Jingwei Huang, Yichao Zhou, Thomas Funkhouser, and Leonidas~J Guibas.
\newblock Framenet: Learning local canonical frames of 3d surfaces from a
  single rgb image.
\newblock In \emph{ICCV}, 2019{\natexlab{a}}.

\bibitem[Huang et~al.(2019{\natexlab{b}})Huang, Bras, Bhagavatula, and
  Choi]{huang2019cosmos}
Lifu Huang, Ronan~Le Bras, Chandra Bhagavatula, and Yejin Choi.
\newblock Cosmos qa: Machine reading comprehension with contextual commonsense
  reasoning.
\newblock \emph{arXiv}, 2019{\natexlab{b}}.

\bibitem[Iyer et~al.(2017)Iyer, Dandekar, and Csernai]{qqp}
Shankar Iyer, Nikhil Dandekar, and Kornel Csernai.
\newblock First quora dataset release: Question pairs, 2017.
\newblock URL
  \url{https://data.quora.com/First-Quora-Dataset-Release-Question-Pairs}.

\bibitem[Jaegle et~al.(2022)Jaegle, Borgeaud, Alayrac, Doersch, Ionescu, Ding,
  Koppula, Zoran, Brock, Shelhamer, Henaff, Botvinick, Zisserman, Vinyals, and
  Carreira]{jaegle2021perceiver}
Andrew Jaegle, Sebastian Borgeaud, Jean-Baptiste Alayrac, Carl Doersch, Catalin
  Ionescu, David Ding, Skanda Koppula, Daniel Zoran, Andrew Brock, Evan
  Shelhamer, Olivier~J Henaff, Matthew Botvinick, Andrew Zisserman, Oriol
  Vinyals, and Joao Carreira.
\newblock Perceiver io: A general architecture for structured inputs \&
  outputs.
\newblock In \emph{ICLR}, 2022.

\bibitem[Joshi et~al.(2017)Joshi, Choi, Weld, and
  Zettlemoyer]{joshi-etal-2017-triviaqa}
Mandar Joshi, Eunsol Choi, Daniel Weld, and Luke Zettlemoyer.
\newblock {T}rivia{QA}: A large scale distantly supervised challenge dataset
  for reading comprehension.
\newblock In \emph{Proceedings of the 55th Annual Meeting of the Association
  for Computational Linguistics (Volume 1: Long Papers)}, pp.\  1601--1611,
  Vancouver, Canada, July 2017. Association for Computational Linguistics.
\newblock \doi{10.18653/v1/P17-1147}.
\newblock URL \url{https://aclanthology.org/P17-1147}.

\bibitem[Kaiser et~al.(2017)Kaiser, Gomez, Shazeer, Vaswani, Parmar, Jones, and
  Uszkoreit]{kaiser2017one}
Lukasz Kaiser, Aidan~N Gomez, Noam Shazeer, Ashish Vaswani, Niki Parmar, Llion
  Jones, and Jakob Uszkoreit.
\newblock One model to learn them all.
\newblock \emph{arXiv preprint arXiv:1706.05137}, 2017.

\bibitem[Kamath et~al.(2021)Kamath, Singh, LeCun, Synnaeve, Misra, and
  Carion]{kamath2021mdetr}
Aishwarya Kamath, Mannat Singh, Yann LeCun, Gabriel Synnaeve, Ishan Misra, and
  Nicolas Carion.
\newblock Mdetr-modulated detection for end-to-end multi-modal understanding.
\newblock In \emph{Proceedings of the IEEE/CVF International Conference on
  Computer Vision}, pp.\  1780--1790, 2021.

\bibitem[Kamath et~al.(2022)Kamath, Clark, Gupta, Kolve, Hoiem, and
  Kembhavi]{Kamath2022WeblySC}
Amita Kamath, Christopher Clark, Tanmay Gupta, Eric Kolve, Derek Hoiem, and
  Aniruddha Kembhavi.
\newblock Webly supervised concept expansion for general purpose vision models.
\newblock \emph{arXiv}, 2022.

\bibitem[Kazemzadeh et~al.(2014)Kazemzadeh, Ordonez, Matten, and
  Berg]{referitgame}
Sahar Kazemzadeh, Vicente Ordonez, Mark Matten, and Tamara Berg.
\newblock Referitgame: Referring to objects in photographs of natural scenes.
\newblock In \emph{EMNLP}, 2014.

\bibitem[Khashabi et~al.(2018)Khashabi, Chaturvedi, Roth, Upadhyay, and
  Roth]{multirc}
Daniel Khashabi, Snigdha Chaturvedi, Michael Roth, Shyam Upadhyay, and Dan
  Roth.
\newblock Looking beyond the surface:a challenge set for reading comprehension
  over multiple sentences.
\newblock In \emph{Proceedings of North American Chapter of the Association for
  Computational Linguistics (NAACL)}, 2018.

\bibitem[Khot et~al.(2018)Khot, Sabharwal, and Clark]{khot2018scitail}
Tushar Khot, Ashish Sabharwal, and Peter Clark.
\newblock Scitail: A textual entailment dataset from science question
  answering.
\newblock In \emph{AAAI}, 2018.

\bibitem[Kolesnikov et~al.(2022)Kolesnikov, Pinto, Beyer, Zhai, Harmsen, and
  Houlsby]{kolesnikov2022uvim}
Alexander Kolesnikov, Andr{\'e}~Susano Pinto, Lucas Beyer, Xiaohua Zhai,
  Jeremiah Harmsen, and Neil Houlsby.
\newblock Uvim: A unified modeling approach for vision with learned guiding
  codes.
\newblock \emph{arXiv}, 2022.

\bibitem[Krishna et~al.(2017)Krishna, Zhu, Groth, Johnson, Hata, Kravitz, Chen,
  Kalantidis, Li, Shamma, Bernstein, and Li]{visual_genome}
Ranjay Krishna, Yuke Zhu, Oliver Groth, Justin Johnson, Kenji Hata, Joshua
  Kravitz, Stephanie Chen, Yannis Kalantidis, Li-Jia Li, David~A. Shamma,
  Michael~S. Bernstein, and Fei-Fei Li.
\newblock Visual genome: Connecting language and vision using crowdsourced
  dense image annotations.
\newblock \emph{IJCV}, 2017.

\bibitem[Kudo \& Richardson(2018)Kudo and Richardson]{kudo2018sentencepiece}
Taku Kudo and John Richardson.
\newblock Sentencepiece: A simple and language independent subword tokenizer
  and detokenizer for neural text processing.
\newblock In \emph{EMNLP}, 2018.

\bibitem[Kuznetsova et~al.(2020)Kuznetsova, Rom, Alldrin, Uijlings, Krasin,
  Pont-Tuset, Kamali, Popov, Malloci, Kolesnikov, Duerig, and
  Ferrari]{open_images}
Alina Kuznetsova, Hassan Rom, Neil Alldrin, Jasper Uijlings, Ivan Krasin, Jordi
  Pont-Tuset, Shahab Kamali, Stefan Popov, Matteo Malloci, Alexander
  Kolesnikov, Tom Duerig, and Vittorio Ferrari.
\newblock The open images dataset v4: Unified image classification, object
  detection, and visual relationship detection at scale.
\newblock \emph{IJCV}, 2020.

\bibitem[Kwiatkowski et~al.(2019)Kwiatkowski, Palomaki, Redfield, Collins,
  Parikh, Alberti, Epstein, Polosukhin, Devlin, Lee, Toutanova, Jones, Kelcey,
  Chang, Dai, Uszkoreit, Le, and Petrov]{kwiatkowski-etal-2019-natural}
Tom Kwiatkowski, Jennimaria Palomaki, Olivia Redfield, Michael Collins, Ankur
  Parikh, Chris Alberti, Danielle Epstein, Illia Polosukhin, Jacob Devlin,
  Kenton Lee, Kristina Toutanova, Llion Jones, Matthew Kelcey, Ming-Wei Chang,
  Andrew~M. Dai, Jakob Uszkoreit, Quoc Le, and Slav Petrov.
\newblock Natural questions: A benchmark for question answering research.
\newblock \emph{Transactions of the Association for Computational Linguistics},
  7:\penalty0 452--466, 2019.
\newblock \doi{10.1162/tacl_a_00276}.
\newblock URL \url{https://aclanthology.org/Q19-1026}.

\bibitem[Levesque et~al.(2012)Levesque, Davis, and
  Morgenstern]{levesque2012winograd}
Hector Levesque, Ernest Davis, and Leora Morgenstern.
\newblock The winograd schema challenge.
\newblock In \emph{Thirteenth International Conference on the Principles of
  Knowledge Representation and Reasoning}, 2012.

\bibitem[Li et~al.(2022{\natexlab{a}})Li, Xu, Tian, Wang, Yan, Bi, Ye, Chen,
  Xu, Cao, et~al.]{li2022mplug}
Chenliang Li, Haiyang Xu, Junfeng Tian, Wei Wang, Ming Yan, Bin Bi, Jiabo Ye,
  Hehong Chen, Guohai Xu, Zheng Cao, et~al.
\newblock mplug: Effective and efficient vision-language learning by
  cross-modal skip-connections.
\newblock \emph{arXiv preprint arXiv:2205.12005}, 2022{\natexlab{a}}.

\bibitem[Li et~al.(2021)Li, Selvaraju, Gotmare, Joty, Xiong, and
  Hoi]{li2021align}
Junnan Li, Ramprasaath Selvaraju, Akhilesh Gotmare, Shafiq Joty, Caiming Xiong,
  and Steven Chu~Hong Hoi.
\newblock Align before fuse: Vision and language representation learning with
  momentum distillation.
\newblock \emph{Advances in neural information processing systems},
  34:\penalty0 9694--9705, 2021.

\bibitem[Li et~al.(2022{\natexlab{b}})Li, Li, Xiong, and Hoi]{li2022blip}
Junnan Li, Dongxu Li, Caiming Xiong, and Steven Hoi.
\newblock Blip: Bootstrapping language-image pre-training for unified
  vision-language understanding and generation.
\newblock In \emph{ICML}, 2022{\natexlab{b}}.

\bibitem[Li et~al.(2022{\natexlab{c}})Li, Gan, Lin, Lin, Liu, Liu, and
  Wang]{li2022lavender}
Linjie Li, Zhe Gan, Kevin Lin, Chung-Ching Lin, Zicheng Liu, Ce~Liu, and Lijuan
  Wang.
\newblock Lavender: Unifying video-language understanding as masked language
  modeling.
\newblock \emph{arXiv preprint arXiv:2206.07160}, 2022{\natexlab{c}}.

\bibitem[Li et~al.(2019)Li, Yatskar, Yin, Hsieh, and Chang]{li2019visualbert}
Liunian~Harold Li, Mark Yatskar, Da~Yin, Cho-Jui Hsieh, and Kai-Wei Chang.
\newblock Visualbert: A simple and performant baseline for vision and language.
\newblock In \emph{arXiv}, 2019.

\bibitem[Li et~al.(2022{\natexlab{d}})Li, Zhang, Zhang, Yang, Li, Zhong, Wang,
  Yuan, Zhang, Hwang, et~al.]{li2022grounded}
Liunian~Harold Li, Pengchuan Zhang, Haotian Zhang, Jianwei Yang, Chunyuan Li,
  Yiwu Zhong, Lijuan Wang, Lu~Yuan, Lei Zhang, Jenq-Neng Hwang, et~al.
\newblock Grounded language-image pre-training.
\newblock In \emph{Proceedings of the IEEE/CVF Conference on Computer Vision
  and Pattern Recognition}, pp.\  10965--10975, 2022{\natexlab{d}}.

\bibitem[Li et~al.(2022{\natexlab{e}})Li, Wang, Liu, and
  Jiang]{li2022binsformer}
Zhenyu Li, Xuyang Wang, Xianming Liu, and Junjun Jiang.
\newblock Binsformer: Revisiting adaptive bins for monocular depth estimation.
\newblock \emph{arXiv}, 2022{\natexlab{e}}.

\bibitem[Liang et~al.(2022)Liang, Lyu, Fan, Mo, Yogatama, Morency, and
  Salakhutdinov]{liang2022highmmt}
Paul~Pu Liang, Yiwei Lyu, Xiang Fan, Shengtong Mo, Dani Yogatama,
  Louis-Philippe Morency, and Ruslan Salakhutdinov.
\newblock Highmmt: Towards modality and task generalization for high-modality
  representation learning.
\newblock \emph{arXiv preprint arXiv:2203.01311}, 2022.

\bibitem[Lin et~al.(2014)Lin, Maire, Belongie, Hays, Perona, Ramanan,
  Doll{\'a}r, and Zitnick]{coco}
Tsung-Yi Lin, Michael Maire, Serge Belongie, James Hays, Pietro Perona, Deva
  Ramanan, Piotr Doll{\'a}r, and C~Lawrence Zitnick.
\newblock Microsoft coco: Common objects in context.
\newblock In \emph{ECCV}, 2014.

\bibitem[Liu et~al.(2019)Liu, He, Chen, and Gao]{liu2019mt-dnn}
Xiaodong Liu, Pengcheng He, Weizhu Chen, and Jianfeng Gao.
\newblock Multi-task deep neural networks for natural language understanding.
\newblock In \emph{Proceedings of the 57th Annual Meeting of the Association
  for Computational Linguistics}, pp.\  4487--4496. Association for
  Computational Linguistics, 2019.
\newblock URL \url{https://www.aclweb.org/anthology/P19-1441}.

\bibitem[Lu et~al.(2016)Lu, Krishna, Bernstein, and Fei-Fei]{lu2016visual}
Cewu Lu, Ranjay Krishna, Michael Bernstein, and Li~Fei-Fei.
\newblock Visual relationship detection with language priors.
\newblock In \emph{European Conference on Computer Vision}, 2016.

\bibitem[Lu et~al.(2019)Lu, Batra, Parikh, and Lee]{lu2019vilbert}
Jiasen Lu, Dhruv Batra, Devi Parikh, and Stefan Lee.
\newblock Vilbert: Pretraining task-agnostic visiolinguistic representations
  for vision-and-language tasks.
\newblock In \emph{NeurIPS}, 2019.

\bibitem[Lu et~al.(2020)Lu, Goswami, Rohrbach, Parikh, and Lee]{Lu202012in1MV}
Jiasen Lu, Vedanuj Goswami, Marcus Rohrbach, D.~Parikh, and Stefan Lee.
\newblock 12-in-1: Multi-task vision and language representation learning.
\newblock In \emph{CVPR}, 2020.

\bibitem[Maas et~al.(2011)Maas, Daly, Pham, Huang, Ng, and Potts]{imdb_reviews}
Andrew~L. Maas, Raymond~E. Daly, Peter~T. Pham, Dan Huang, Andrew~Y. Ng, and
  Christopher Potts.
\newblock Learning word vectors for sentiment analysis.
\newblock In \emph{Proceedings of the 49th Annual Meeting of the Association
  for Computational Linguistics: Human Language Technologies}, pp.\  142--150,
  Portland, Oregon, USA, June 2011. Association for Computational Linguistics.
\newblock URL \url{http://www.aclweb.org/anthology/P11-1015}.

\bibitem[Mao et~al.(2016)Mao, Huang, Toshev, Camburu, Yuille, and
  Murphy]{google_refcoco}
Junhua Mao, Jonathan Huang, Alexander Toshev, Oana Camburu, Alan Yuille, and
  Kevin Murphy.
\newblock Generation and comprehension of unambiguous object descriptions.
\newblock In \emph{CVPR}, 2016.

\bibitem[Marino et~al.(2019)Marino, Rastegari, Farhadi, and Mottaghi]{okvqa}
Kenneth Marino, Mohammad Rastegari, Ali Farhadi, and Roozbeh Mottaghi.
\newblock Ok-vqa: A visual question answering benchmark requiring external
  knowledge.
\newblock In \emph{Conference on Computer Vision and Pattern Recognition
  (CVPR)}, 2019.

\bibitem[McCann et~al.(2018)McCann, Keskar, Xiong, and
  Socher]{McCann2018decaNLP}
Bryan McCann, Nitish~Shirish Keskar, Caiming Xiong, and Richard Socher.
\newblock The natural language decathlon: Multitask learning as question
  answering.
\newblock \emph{arXiv}, 2018.

\bibitem[Mihaylov et~al.(2018)Mihaylov, Clark, Khot, and
  Sabharwal]{mihaylov2018can}
Todor Mihaylov, Peter Clark, Tushar Khot, and Ashish Sabharwal.
\newblock Can a suit of armor conduct electricity? a new dataset for open book
  question answering.
\newblock In \emph{EMNLP}, 2018.

\bibitem[Nathan~Silberman \& Fergus(2012)Nathan~Silberman and
  Fergus]{nyu_depth}
Pushmeet~Kohli Nathan~Silberman, Derek~Hoiem and Rob Fergus.
\newblock Indoor segmentation and support inference from {RGBD} images.
\newblock In \emph{ECCV}, 2012.

\bibitem[Park et~al.(2020)Park, Bhagavatula, Mottaghi, Farhadi, and
  Choi]{park2020visualcomet}
Jae~Sung Park, Chandra Bhagavatula, Roozbeh Mottaghi, Ali Farhadi, and Yejin
  Choi.
\newblock Visualcomet: Reasoning about the dynamic context of a still image.
\newblock In \emph{In Proceedings of the European Conference on Computer Vision
  (ECCV)}, 2020.

\bibitem[Peng et~al.(2022)Peng, Dong, Bao, Ye, and Wei]{beitv2}
Zhiliang Peng, Li~Dong, Hangbo Bao, Qixiang Ye, and Furu Wei.
\newblock Beit v2: Masked image modeling with vector-quantized visual
  tokenizers.
\newblock \emph{ArXiv}, abs/2208.06366, 2022.

\bibitem[Pilehvar \& os{'{e} } Camacho{-}Collados(2018)Pilehvar and os{'{e} }
  Camacho{-}Collados]{wic}
Mohammad~Taher Pilehvar and os{'{e} } Camacho{-}Collados.
\newblock Wic: 10, 000 example pairs for evaluating context-sensitive
  representations.
\newblock \emph{CoRR}, abs/1808.09121, 2018.
\newblock URL \url{http://arxiv.org/abs/1808.09121}.

\bibitem[Pinz et~al.(2006)]{pinz2006object}
Axel Pinz et~al.
\newblock Object categorization.
\newblock \emph{Foundations and Trends{\textregistered} in Computer Graphics
  and Vision}, 1\penalty0 (4):\penalty0 255--353, 2006.

\bibitem[Pratt et~al.(2020)Pratt, Yatskar, Weihs, Farhadi, and
  Kembhavi]{pratt2020grounded}
Sarah Pratt, Mark Yatskar, Luca Weihs, Ali Farhadi, and Aniruddha Kembhavi.
\newblock Grounded situation recognition.
\newblock In \emph{ECCV}, 2020.

\bibitem[Radford et~al.(2021)Radford, Kim, Hallacy, Ramesh, Goh, Agarwal,
  Sastry, Askell, Mishkin, Clark, Krueger, and Sutskever]{radford2021learning}
Alec Radford, Jong~Wook Kim, Chris Hallacy, Aditya Ramesh, Gabriel Goh,
  Sandhini Agarwal, Girish Sastry, Amanda Askell, Pamela Mishkin, Jack Clark,
  Gretchen Krueger, and Ilya Sutskever.
\newblock Learning transferable visual models from natural language
  supervision.
\newblock In \emph{ICML}, 2021.

\bibitem[Raffel et~al.(2020)Raffel, Shazeer, Roberts, Lee, Narang, Matena,
  Zhou, Li, and Liu]{2020t5}
Colin Raffel, Noam Shazeer, Adam Roberts, Katherine Lee, Sharan Narang, Michael
  Matena, Yanqi Zhou, Wei Li, and Peter~J. Liu.
\newblock Exploring the limits of transfer learning with a unified text-to-text
  transformer.
\newblock \emph{JMLR}, 2020.

\bibitem[Rajpurkar et~al.(2016)Rajpurkar, Zhang, Lopyrev, and
  Liang]{rajpurkar2016squad}
Pranav Rajpurkar, Jian Zhang, Konstantin Lopyrev, and Percy Liang.
\newblock Squad: 100,000+ questions for machine comprehension of text.
\newblock In \emph{EMNLP}, 2016.

\bibitem[Ramesh et~al.(2022)Ramesh, Dhariwal, Nichol, Chu, and Chen]{dalle2}
Aditya Ramesh, Prafulla Dhariwal, Alex Nichol, Casey Chu, and Mark Chen.
\newblock Hierarchical text-conditional image generation with clip latents.
\newblock \emph{arXiv}, 2022.

\bibitem[Reed et~al.(2022)Reed, Zolna, Parisotto, Colmenarejo, Novikov,
  Barth-Maron, Gimenez, Sulsky, Kay, Springenberg, Eccles, Bruce, Razavi,
  Edwards, Heess, Chen, Hadsell, Vinyals, Bordbar, and de~Freitas]{gato}
Scott Reed, Konrad Zolna, Emilio Parisotto, Sergio~Gomez Colmenarejo, Alexander
  Novikov, Gabriel Barth-Maron, Mai Gimenez, Yury Sulsky, Jackie Kay,
  Jost~Tobias Springenberg, Tom Eccles, Jake Bruce, Ali Razavi, Ashley Edwards,
  Nicolas Heess, Yutian Chen, Raia Hadsell, Oriol Vinyals, Mahyar Bordbar, and
  Nando de~Freitas.
\newblock A generalist agent.
\newblock \emph{arXiv}, 2022.

\bibitem[Rhodes et~al.(2017)Rhodes, Quinn, and Mitchell]{rhodes2017fast}
Anthony~D Rhodes, Max~H Quinn, and Melanie Mitchell.
\newblock Fast on-line kernel density estimation for active object
  localization.
\newblock In \emph{2017 international joint conference on neural networks
  (IJCNN)}, pp.\  454--462. IEEE, 2017.

\bibitem[Ridnik et~al.(2021)Ridnik, Ben-Baruch, Noy, and
  Zelnik-Manor]{ridnik2021imagenet}
Tal Ridnik, Emanuel Ben-Baruch, Asaf Noy, and Lihi Zelnik-Manor.
\newblock Imagenet-21k pretraining for the masses.
\newblock \emph{arXiv}, 2021.

\bibitem[Roemmele et~al.(2011)Roemmele, Bejan, and
  Gordon]{choice_of_plausible_alternatives}
Melissa Roemmele, Cosmin~Adrian Bejan, and Andrew~S Gordon.
\newblock Choice of plausible alternatives: An evaluation of commonsense causal
  reasoning.
\newblock In \emph{2011 AAAI Spring Symposium Series}, 2011.

\bibitem[Rush et~al.(2015)Rush, Chopra, and Weston]{Rush_2015}
Alexander~M. Rush, Sumit Chopra, and Jason Weston.
\newblock A neural attention model for abstractive sentence summarization.
\newblock \emph{Proceedings of the 2015 Conference on Empirical Methods in
  Natural Language Processing}, 2015.
\newblock \doi{10.18653/v1/d15-1044}.
\newblock URL \url{http://dx.doi.org/10.18653/v1/D15-1044}.

\bibitem[Schwenk et~al.(2022)Schwenk, Khandelwal, Clark, Marino, and
  Mottaghi]{schwenk2022okvqa}
Dustin Schwenk, Apoorv Khandelwal, Christopher Clark, Kenneth Marino, and
  Roozbeh Mottaghi.
\newblock A-okvqa: A benchmark for visual question answering using world
  knowledge.
\newblock \emph{arXiv}, 2022.

\bibitem[Shazeer \& Stern(2018)Shazeer and Stern]{shazeer2018adafactor}
Noam Shazeer and Mitchell Stern.
\newblock Adafactor: Adaptive learning rates with sublinear memory cost.
\newblock In \emph{International Conference on Machine Learning}, pp.\
  4596--4604. PMLR, 2018.

\bibitem[Singh et~al.(2022)Singh, Hu, Goswami, Couairon, Galuba, Rohrbach, and
  Kiela]{singh2022flava}
Amanpreet Singh, Ronghang Hu, Vedanuj Goswami, Guillaume Couairon, Wojciech
  Galuba, Marcus Rohrbach, and Douwe Kiela.
\newblock Flava: A foundational language and vision alignment model.
\newblock In \emph{Proceedings of the IEEE/CVF Conference on Computer Vision
  and Pattern Recognition}, pp.\  15638--15650, 2022.

\bibitem[Smith et~al.(2022)Smith, Patwary, Norick, LeGresley, Rajbhandari,
  Casper, Liu, Prabhumoye, Zerveas, Korthikanti, Zheng, Child, Aminabadi,
  Bernauer, Song, Shoeybi, He, Houston, Tiwary, and Catanzaro]{smith2022using}
Shaden Smith, Mostofa Patwary, Brandon Norick, Patrick LeGresley, Samyam
  Rajbhandari, Jared Casper, Zhun Liu, Shrimai Prabhumoye, George Zerveas,
  Vijay Korthikanti, Elton Zheng, Rewon Child, Reza~Yazdani Aminabadi, Julie
  Bernauer, Xia Song, Mohammad Shoeybi, Yuxiong He, Michael Houston, Saurabh
  Tiwary, and Bryan Catanzaro.
\newblock Using deepspeed and megatron to train megatron-turing nlg 530b, a
  large-scale generative language model.
\newblock \emph{arXiv}, 2022.

\bibitem[Socher et~al.(2013)Socher, Perelygin, Wu, Chuang, Manning, Ng, and
  Potts]{sst2}
Richard Socher, Alex Perelygin, Jean Wu, Jason Chuang, Christopher~D Manning,
  Andrew Ng, and Christopher Potts.
\newblock Recursive deep models for semantic compositionality over a sentiment
  treebank.
\newblock In \emph{Proceedings of the 2013 conference on empirical methods in
  natural language processing}, pp.\  1631--1642, 2013.

\bibitem[Sun et~al.(2022)Sun, Dai, Liang, Liu, Yang, and Bai]{sun2022gppf}
Benyuan Sun, Jin Dai, Zihao Liang, Congying Liu, Yi~Yang, and Bo~Bai.
\newblock Gppf: A general perception pre-training framework via sparsely
  activated multi-task learning.
\newblock \emph{arXiv preprint arXiv:2208.02148}, 2022.

\bibitem[Tan \& Bansal(2019)Tan and Bansal]{tan2019lxmert}
Hao Tan and Mohit Bansal.
\newblock Lxmert: Learning cross-modality encoder representations from
  transformers.
\newblock In \emph{EMNLP}, 2019.

\bibitem[Trischler et~al.(2017)Trischler, Wang, Yuan, Harris, Sordoni, Bachman,
  and Suleman]{trischler-etal-2017-newsqa}
Adam Trischler, Tong Wang, Xingdi Yuan, Justin Harris, Alessandro Sordoni,
  Philip Bachman, and Kaheer Suleman.
\newblock {N}ews{QA}: A machine comprehension dataset.
\newblock In \emph{Proceedings of the 2nd Workshop on Representation Learning
  for {NLP}}, pp.\  191--200, Vancouver, Canada, August 2017. Association for
  Computational Linguistics.
\newblock \doi{10.18653/v1/W17-2623}.
\newblock URL \url{https://aclanthology.org/W17-2623}.

\bibitem[Van~Horn et~al.(2018)Van~Horn, Mac~Aodha, Song, Cui, Sun, Shepard,
  Adam, Perona, and Belongie]{van2018inaturalist}
Grant Van~Horn, Oisin Mac~Aodha, Yang Song, Yin Cui, Chen Sun, Alex Shepard,
  Hartwig Adam, Pietro Perona, and Serge Belongie.
\newblock The inaturalist species classification and detection dataset.
\newblock In \emph{CVPR}, 2018.

\bibitem[Wang et~al.(2018)Wang, Singh, Michael, Hill, Levy, and
  Bowman]{wang2018glue}
Alex Wang, Amanpreet Singh, Julian Michael, Felix Hill, Omer Levy, and Samuel~R
  Bowman.
\newblock Glue: A multi-task benchmark and analysis platform for natural
  language understanding.
\newblock \emph{arXiv}, 2018.

\bibitem[Wang et~al.(2019)Wang, Pruksachatkun, Nangia, Singh, Michael, Hill,
  Levy, and Bowman]{wang2019superglue}
Alex Wang, Yada Pruksachatkun, Nikita Nangia, Amanpreet Singh, Julian Michael,
  Felix Hill, Omer Levy, and Samuel Bowman.
\newblock Superglue: A stickier benchmark for general-purpose language
  understanding systems.
\newblock In \emph{NeurIPS}, 2019.

\bibitem[Wang et~al.(2022{\natexlab{a}})Wang, Yang, Hu, Li, Lin, Gan, Liu, Liu,
  and Wang]{wang2022git}
Jianfeng Wang, Zhengyuan Yang, Xiaowei Hu, Linjie Li, Kevin Lin, Zhe Gan,
  Zicheng Liu, Ce~Liu, and Lijuan Wang.
\newblock Git: A generative image-to-text transformer for vision and language.
\newblock \emph{arXiv preprint arXiv:2205.14100}, 2022{\natexlab{a}}.

\bibitem[Wang et~al.(2022{\natexlab{b}})Wang, Yang, Men, Lin, Bai, Li, Ma,
  Zhou, Zhou, and Yang]{wang2022OFA}
Peng Wang, An~Yang, Rui Men, Junyang Lin, Shuai Bai, Zhikang Li, Jianxin Ma,
  Chang Zhou, Jingren Zhou, and Hongxia Yang.
\newblock Unifying architectures, tasks, and modalities through a simple
  sequence-to-sequence learning framework.
\newblock \emph{arXiv}, 2022{\natexlab{b}}.

\bibitem[Wang et~al.(2021)Wang, Bao, Dong, and Wei]{wang2021vlmo}
Wenhui Wang, Hangbo Bao, Li~Dong, and Furu Wei.
\newblock Vlmo: Unified vision-language pre-training with
  mixture-of-modality-experts.
\newblock \emph{arXiv preprint arXiv:2111.02358}, 2021.

\bibitem[Wang et~al.(2022{\natexlab{c}})Wang, Bao, Dong, Bjorck, Peng, Liu,
  Aggarwal, Mohammed, Singhal, Som, et~al.]{wang2022image}
Wenhui Wang, Hangbo Bao, Li~Dong, Johan Bjorck, Zhiliang Peng, Qiang Liu, Kriti
  Aggarwal, Owais~Khan Mohammed, Saksham Singhal, Subhojit Som, et~al.
\newblock Image as a foreign language: Beit pretraining for all vision and
  vision-language tasks.
\newblock \emph{arXiv preprint arXiv:2208.10442}, 2022{\natexlab{c}}.

\bibitem[Wang et~al.(2022{\natexlab{d}})Wang, Yu, Yu, Dai, Tsvetkov, and
  Cao]{wang2021simvlm}
Zirui Wang, Jiahui Yu, Adams~Wei Yu, Zihang Dai, Yulia Tsvetkov, and Yuan Cao.
\newblock Simvlm: Simple visual language model pretraining with weak
  supervision.
\newblock In \emph{ICLR}, 2022{\natexlab{d}}.

\bibitem[Warstadt et~al.(2018)Warstadt, Singh, and Bowman]{cola}
Alex Warstadt, Amanpreet Singh, and Samuel~R Bowman.
\newblock Neural network acceptability judgments.
\newblock \emph{arXiv preprint arXiv:1805.12471}, 2018.

\bibitem[Welinder et~al.(2010)Welinder, Branson, Mita, Wah, Schroff, Belongie,
  and Perona]{WelinderEtal2010}
P.~Welinder, S.~Branson, T.~Mita, C.~Wah, F.~Schroff, S.~Belongie, and
  P.~Perona.
\newblock {Caltech-UCSD Birds 200}.
\newblock Technical Report CNS-TR-2010-001, California Institute of Technology,
  2010.

\bibitem[Williams et~al.(2018)Williams, Nangia, and Bowman]{mnli}
Adina Williams, Nikita Nangia, and Samuel Bowman.
\newblock A broad-coverage challenge corpus for sentence understanding through
  inference.
\newblock In \emph{Proceedings of the 2018 Conference of the North American
  Chapter of the Association for Computational Linguistics: Human Language
  Technologies, Volume 1 (Long Papers)}, pp.\  1112--1122. Association for
  Computational Linguistics, 2018.
\newblock URL \url{http://aclweb.org/anthology/N18-1101}.

\bibitem[Xiao et~al.(2010)Xiao, Hays, Ehinger, Oliva, and
  Torralba]{xiao2010sun}
Jianxiong Xiao, James Hays, Krista~A Ehinger, Aude Oliva, and Antonio Torralba.
\newblock Sun database: Large-scale scene recognition from abbey to zoo.
\newblock In \emph{CVPR}, 2010.

\bibitem[Xie et~al.(2019)Xie, Lai, Doran, and Kadav]{xie2019visual}
Ning Xie, Farley Lai, Derek Doran, and Asim Kadav.
\newblock Visual entailment: A novel task for fine-grained image understanding.
\newblock \emph{arXiv}, 2019.

\bibitem[Yang et~al.(2021)Yang, Gan, Wang, Hu, Ahmed, Liu, Lu, and
  Wang]{Yang2021UniTABUT}
Zhengyuan Yang, Zhe Gan, Jianfeng Wang, Xiaowei Hu, Faisal Ahmed, Zicheng Liu,
  Yumao Lu, and Lijuan Wang.
\newblock Unitab: Unifying text and box outputs for grounded vision-language
  modeling.
\newblock 2021.

\bibitem[Yang et~al.(2018)Yang, Qi, Zhang, Bengio, Cohen, Salakhutdinov, and
  Manning]{yang-etal-2018-hotpotqa}
Zhilin Yang, Peng Qi, Saizheng Zhang, Yoshua Bengio, William Cohen, Ruslan
  Salakhutdinov, and Christopher~D. Manning.
\newblock {H}otpot{QA}: A dataset for diverse, explainable multi-hop question
  answering.
\newblock In \emph{Proceedings of the 2018 Conference on Empirical Methods in
  Natural Language Processing}, pp.\  2369--2380, Brussels, Belgium,
  October-November 2018. Association for Computational Linguistics.
\newblock \doi{10.18653/v1/D18-1259}.
\newblock URL \url{https://aclanthology.org/D18-1259}.

\bibitem[Yao et~al.(2020)Yao, Luo, Li, Zhang, Ren, Zhou, Fang, and
  Quan]{yao2020blendedmvs}
Yao Yao, Zixin Luo, Shiwei Li, Jingyang Zhang, Yufan Ren, Lei Zhou, Tian Fang,
  and Long Quan.
\newblock Blendedmvs: A large-scale dataset for generalized multi-view stereo
  networks.
\newblock In \emph{CVPR}, 2020.

\bibitem[Yao et~al.(2022)Yao, Chen, Zhang, Ji, Liu, Chua, and Sun]{yao2022pevl}
Yuan Yao, Qianyu Chen, Ao~Zhang, Wei Ji, Zhiyuan Liu, Tat-Seng Chua, and
  Maosong Sun.
\newblock Pevl: Position-enhanced pre-training and prompt tuning for
  vision-language models.
\newblock \emph{arXiv preprint arXiv:2205.11169}, 2022.

\bibitem[Yatskar et~al.(2016)Yatskar, Zettlemoyer, and
  Farhadi]{yatskar2016situation}
Mark Yatskar, Luke Zettlemoyer, and Ali Farhadi.
\newblock Situation recognition: Visual semantic role labeling for image
  understanding.
\newblock In \emph{Proceedings of the IEEE conference on computer vision and
  pattern recognition}, pp.\  5534--5542, 2016.

\bibitem[Yu et~al.(2022{\natexlab{a}})Yu, Wang, Vasudevan, Yeung,
  Seyedhosseini, and Wu]{yu2022coca}
Jiahui Yu, Zirui Wang, Vijay Vasudevan, Legg Yeung, Mojtaba Seyedhosseini, and
  Yonghui Wu.
\newblock Coca: Contrastive captioners are image-text foundation models.
\newblock \emph{arXiv preprint arXiv:2205.01917}, 2022{\natexlab{a}}.

\bibitem[Yu et~al.(2022{\natexlab{b}})Yu, Xu, Koh, Luong, Baid, Wang,
  Vasudevan, Ku, Yang, Ayan, et~al.]{parti}
Jiahui Yu, Yuanzhong Xu, Jing~Yu Koh, Thang Luong, Gunjan Baid, Zirui Wang,
  Vijay Vasudevan, Alexander Ku, Yinfei Yang, Burcu~Karagol Ayan, et~al.
\newblock Scaling autoregressive models for content-rich text-to-image
  generation.
\newblock \emph{arXiv preprint arXiv:2206.10789}, 2022{\natexlab{b}}.

\bibitem[Yuan et~al.(2021)Yuan, Chen, Chen, Codella, Dai, Gao, Hu, Huang, Li,
  Li, et~al.]{yuan2021florence}
Lu~Yuan, Dongdong Chen, Yi-Ling Chen, Noel Codella, Xiyang Dai, Jianfeng Gao,
  Houdong Hu, Xuedong Huang, Boxin Li, Chunyuan Li, et~al.
\newblock Florence: A new foundation model for computer vision.
\newblock \emph{arXiv preprint arXiv:2111.11432}, 2021.

\bibitem[Zellers et~al.(2019{\natexlab{a}})Zellers, Bisk, Farhadi, and
  Choi]{zellers2019recognition}
Rowan Zellers, Yonatan Bisk, Ali Farhadi, and Yejin Choi.
\newblock From recognition to cognition: Visual commonsense reasoning.
\newblock In \emph{Proceedings of the IEEE/CVF conference on computer vision
  and pattern recognition}, pp.\  6720--6731, 2019{\natexlab{a}}.

\bibitem[Zellers et~al.(2019{\natexlab{b}})Zellers, Holtzman, Bisk, Farhadi,
  and Choi]{zellers2019hellaswag}
Rowan Zellers, Ari Holtzman, Yonatan Bisk, Ali Farhadi, and Yejin Choi.
\newblock Hellaswag: Can a machine really finish your sentence?
\newblock In \emph{Proceedings of the 57th Annual Meeting of the Association
  for Computational Linguistics}, 2019{\natexlab{b}}.

\bibitem[Zhang et~al.(2019)Zhang, Baldridge, and He]{paws2019naacl}
Yuan Zhang, Jason Baldridge, and Luheng He.
\newblock {PAWS: Paraphrase Adversaries from Word Scrambling}.
\newblock In \emph{Proc. of NAACL}, 2019.

\bibitem[Zhou et~al.(2017)Zhou, Lapedriza, Khosla, Oliva, and
  Torralba]{zhou2017places}
Bolei Zhou, Agata Lapedriza, Aditya Khosla, Aude Oliva, and Antonio Torralba.
\newblock Places: A 10 million image database for scene recognition.
\newblock \emph{TPAMI}, 2017.

\bibitem[Zhu et~al.(2022{\natexlab{a}})Zhu, Zhou, Shen, Luo, Pan, Lin, Chen,
  Cao, Sun, and Ji]{zhu2022seqtr}
Chaoyang Zhu, Yiyi Zhou, Yunhang Shen, Gen Luo, Xingjia Pan, Mingbao Lin, Chao
  Chen, Liujuan Cao, Xiaoshuai Sun, and Rongrong Ji.
\newblock Seqtr: A simple yet universal network for visual grounding.
\newblock \emph{arXiv preprint arXiv:2203.16265}, 2022{\natexlab{a}}.

\bibitem[Zhu et~al.(2022{\natexlab{b}})Zhu, Zhu, Li, Wu, Li, Wang, and
  Dai]{zhu2022uni}
Xizhou Zhu, Jinguo Zhu, Hao Li, Xiaoshi Wu, Hongsheng Li, Xiaohua Wang, and
  Jifeng Dai.
\newblock Uni-perceiver: Pre-training unified architecture for generic
  perception for zero-shot and few-shot tasks.
\newblock In \emph{CVPR}, 2022{\natexlab{b}}.

\bibitem[Zoph et~al.(2022)Zoph, Bello, Kumar, Du, Huang, Dean, Shazeer, and
  Fedus]{zoph2022designing}
Barret Zoph, Irwan Bello, Sameer Kumar, Nan Du, Yanping Huang, Jeff Dean, Noam
  Shazeer, and William Fedus.
\newblock Designing effective sparse expert models.
\newblock \emph{arXiv}, 2022.

\end{thebibliography}
\bibliographystyle{iclr2023_conference}
}
\newpage 
\appendix
\section{Appendix}
\subsection{Tasks Details}
\label{sect:task_details}

\uio\ is jointly trained on a large and diverse set of vision, language and vision \& language tasks. In this section, we describe these tasks in detail and show the prompts we use during training and inference (text on the left of example cards).  
We also provide qualitative examples of both the ground truth and the predictions made by \uio.

\vspace{-0.05in}
\subsubsection{Image Synthesis Tasks}
\boldheader{Image Synthesis from Text} 
This task requires generating an image that matches a sentence. 
Training data comes from 4 captioning datasets: COCO Caption~\citep{coco_captions}, Conceptual Captions 3M and 12M~\citep{cc12cm}, and RedCaps~\citep{redcaps} as well datasets used for image classification using the object class as the input caption.
Specialized image generation models like DALL·E 2~\citep{dalle2} use an order of magnitude more data, but we limit our sources to these sets for training efficiency.

\begin{minipage}{\linewidth}
    \centering
    \includegraphics[width=1\textwidth]{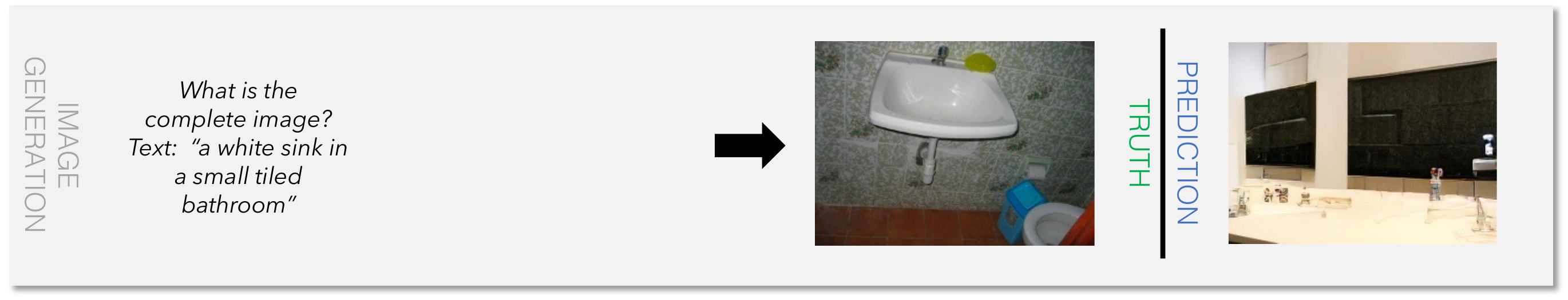}
    \label{fig:tasks_depth}
    \vspace{-0.02in}
\end{minipage}

\boldheader{Image Inpainting} 
This task requires filling in a region of an image with a target object. Training data for this task is built from object bounding box annotations from Open Images~\citep{open_images}, Visual Genome~\citep{visual_genome} and COCO~\citep{coco}. For each object, the input image becomes the source image with the object's bounding box blanked out. The input prompt provides the bounding box's location and the target category. The target output is the original image.

\begin{minipage}{\linewidth}
    \centering
    \includegraphics[width=1\textwidth]{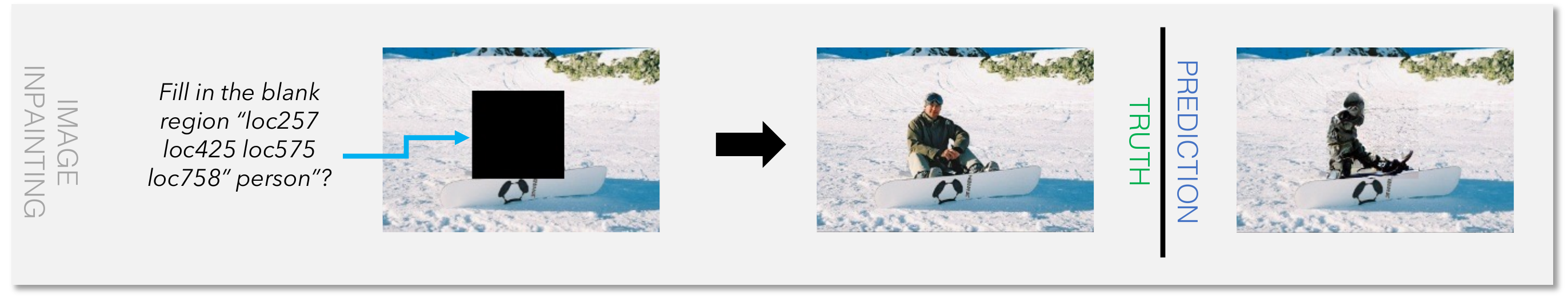}
    \label{fig:tasks_depth}
    \vspace{-0.02in}
\end{minipage}

\boldheader{Image Synthesis from Segmentation} 
This task involves generating an image that matches an input semantic segmentation, i.e., a set of class labels for some or all of the pixels in the image.
\uio~is trained for this task using segmentation annotations from COCO \citep{coco}, Open Images~\citep{open_images}, and LVIS~\citep{lvis} as input.
Following the method from Section~\ref{sect:task-representation} the segmentation input is converted into a RGB image paired with a prompt listing the color-to-class mapping, and the target output is the source image.

\begin{minipage}{\linewidth}
    \centering
    \includegraphics[width=1\textwidth]{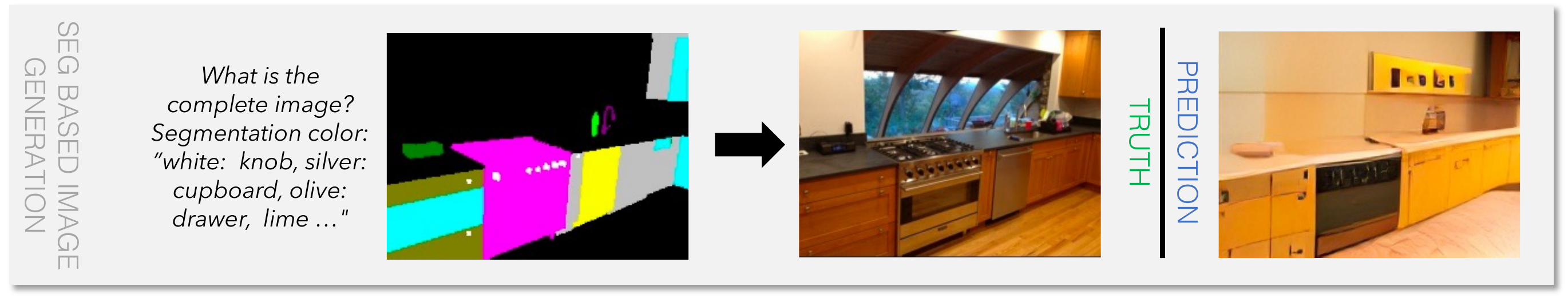}
    \label{fig:tasks_depth}
    \vspace{-0.02in}
\end{minipage}

\vspace{-0.05in}
\subsubsection{Sparse Labelling Tasks}

\boldheader{Object Detection}
\uio~is trained on object detection annotations from Visual Genome, Open Images, and COCO. For this task the input is a static prompt and an image, and the output text includes the bounding boxes and class names of all objects in the image. We randomize the order of the output objects during training, but for simplicity leave integrating more complex data-augmentation techniques~\citep{chen2021pix2seq} to future work.

\begin{minipage}{\linewidth}
    \centering
    \includegraphics[width=1\textwidth]{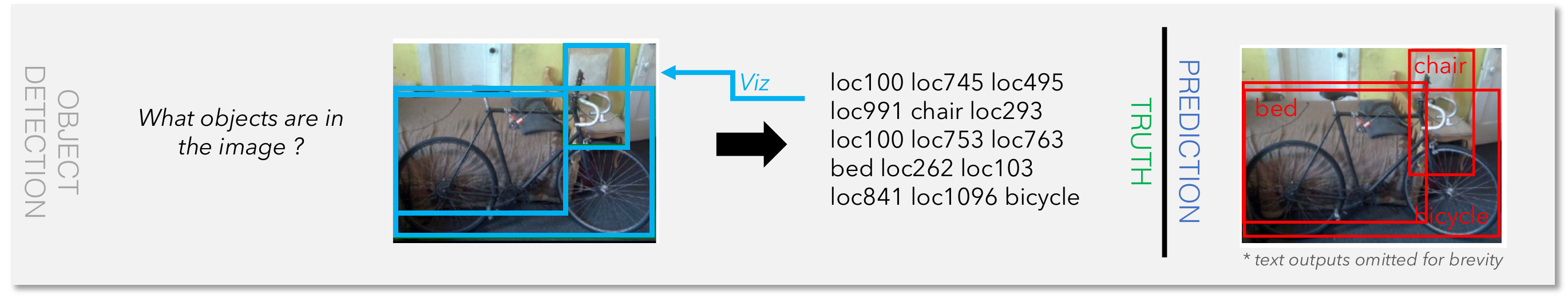}
    \vspace{-0.02in}
    \label{fig:tasks_depth}
\end{minipage}

\boldheader{Object Localization} Object localization requires returning bounding boxes around all objects of a given category. Training data is derived from our object detection training data by constructing a training example from each category of objects present in an image. 
The input is then the image, a prompt specifying the target class, and the output is a list of all boxes that contain an instance of that class.
The class for each box (which is always the class specified in the prompt) is included in the output for the sake of keeping the output format consistent with the object detection output. 
Object localization can use input categories which are not present in the image. To handle this, we construct negative samples by randomly selecting categories not present in the image to use as input, in which case the output is an empty sequence.

\begin{minipage}{\linewidth}
    \centering
    \includegraphics[width=1\textwidth]{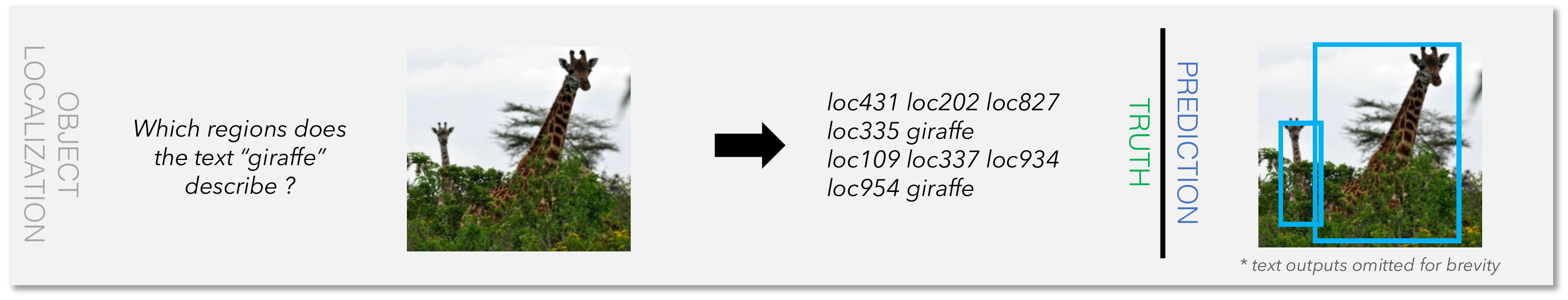}
    \vspace{-0.02in}
    \label{fig:tasks_depth}
\end{minipage}

\boldheader{Referring Expression Comprehension} The task requires the model to localize an image region described by a natural language expression. The annotation is similar to Object Localization, except that the target is specified with natural language expression instead of class name. Datasets for this task include RefCOCO~\citep{referitgame}, RefCOCO+~\citep{referitgame} and RefCOCOg~\citep{google_refcoco}.

\begin{minipage}{\linewidth}
    \centering
    \includegraphics[width=1\textwidth]{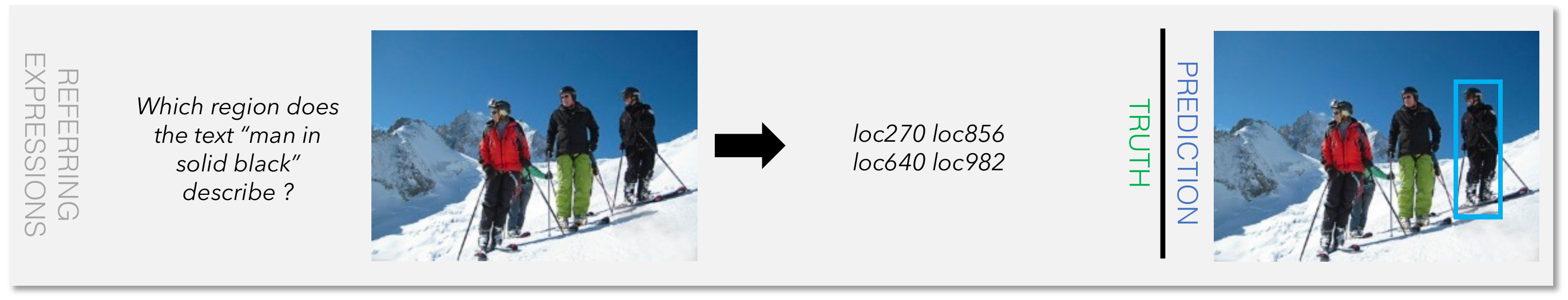}
    \vspace{-0.02in}
    \label{fig:tasks_depth}
\end{minipage}

\boldheader{Keypoint Estimation} Keypoint estimation requires returning the location of 17 keypoints on a human body (e.g., eyes, nose, feet, etc.) for each person in an image. While it is possible to perform this task in one pass by listing the keypoints of all people in the image in a single output sequence, this can result in an extremely long output sequence, so \uio\ uses a multi-step approach instead.
To do this \uio\ is trained to complete the subtask of detecting the keypoints for single a person in a given region.
For this subtask, the input prompt specifies the target region and and the output is a list of 17 points (a pair of locations tokens for the $x$ and $y$ coordinates) along with a visibility labels (1 for not visible, 2 for partly visible, 3 for fully visible). 
Non-visible points are preceded by two copies of a new special tokens that indicate there are no valid coordinates. The keypoint metric does not award points for correctly identifying non-visible points, so during inference we mask that special token so the model makes a best-effort guess for the coordinates of every single point.
Training data for this subtask comes from COCO human pose data~\citep{coco} with the ground-truth person regions as input.
During inference we locate person regions using the object localization prompt, then apply \uio\ again to find keypoints for each detected region.

\begin{minipage}{\linewidth}
    \centering
    \includegraphics[width=1\textwidth]{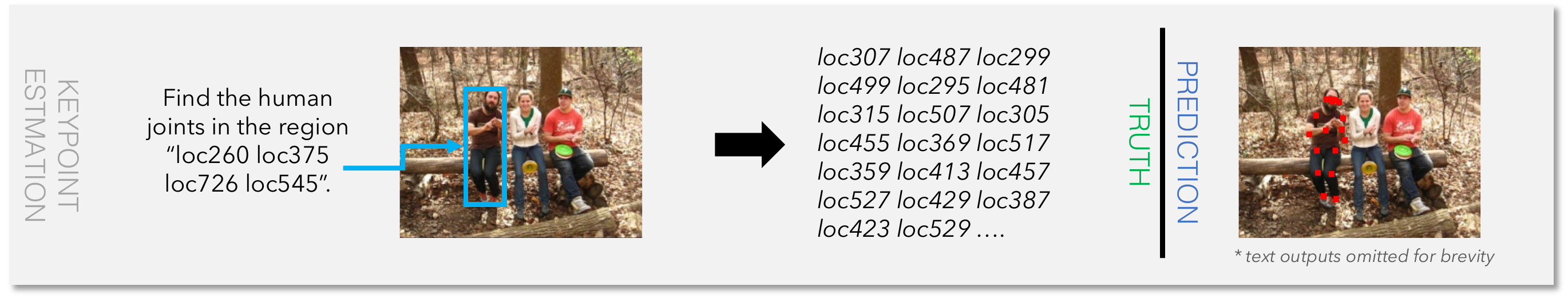}
    \vspace{-0.02in}
    \label{fig:tasks_depth}
\end{minipage}

\vspace{-0.05in}
\subsubsection{Dense Labelling Tasks}
\vspace{-0.05in}
\boldheader{Object Segmentation} Object segmentation requires finding the binary segmentation mask of each instance of a particular category in an image. The input is an image and a prompt that includes the target class, while the output is an RGB image with black background and instances of that class filled in with unique colors following the method in Section~\ref{sect:task-representation}. The output image is resized to match the input image if needed using a nearest-neighbor resizing method, and binary masks are built from each unique color. In practice the output image from \uio\ can have slightly non-uniform colors or extraneous background pixels, likely due to limitation in what the \dvae\ can decode/encode, so the output pixels are clustered by color and and connected components of less than 8 pixels are removed to build cleaned instance masks. Segmentation annotations come from Open Images\, LVIS, and COCO.

\begin{minipage}{\linewidth}
    \centering
    \includegraphics[width=1\textwidth]{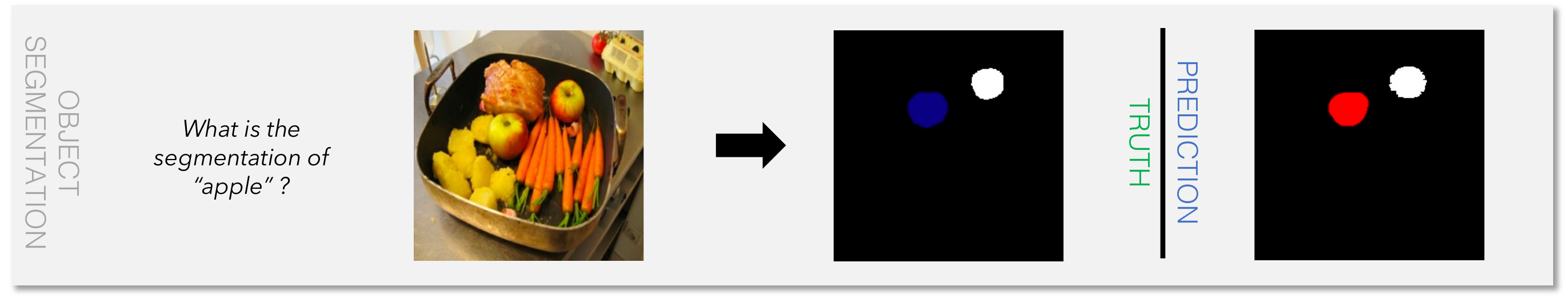}
    \vspace{-0.02in}
    \label{fig:tasks_depth}
\end{minipage}

\boldheader{Depth Estimation} Depth estimation requires assigning each pixel in an image a depth value. This task uses a static prompt as input, and the output is a grayscale image representing the normalized depth at each pixel. The generated output image is reiszed to the same size as the input image and then pixel values are rescaled to the maximum depth in the training to get an output depth map. Training data comes from the NYU Depth Dataset V2~\citep{nyu_depth}.  

\begin{minipage}{\linewidth}
    \centering
    \includegraphics[width=1\textwidth]{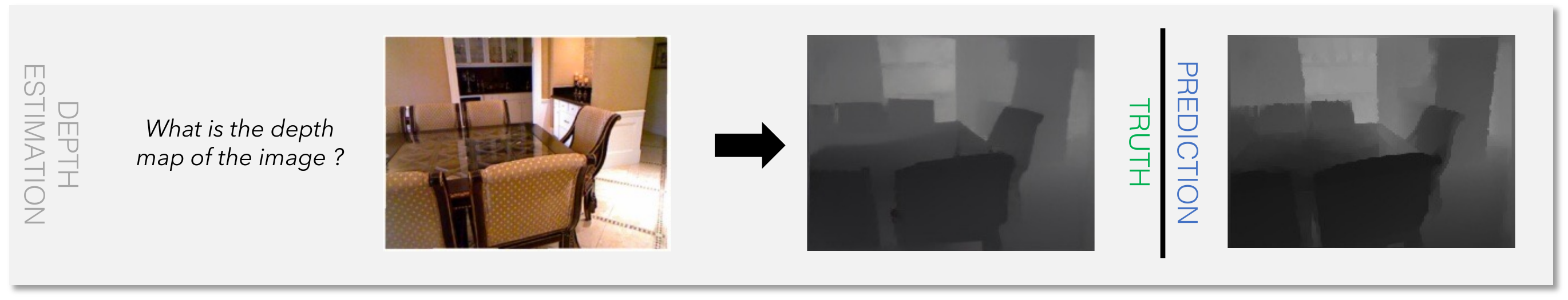}
    \vspace{-0.02in}
    \label{fig:tasks_depth}
\end{minipage}

\boldheader{Surface Normal Estimation} \uio\ is trained on FrameNet~\citep{huang2019framenet} and BlendedMVS~\citep{yao2020blendedmvs} surface normal estimation datasets.
For this task the input is a static prompt and an image and the output is an RGB representation of the $x/y/z$ orientation of the surface at each pixel. The generated output image is resized to match the input image and converted back to $x/y/z$ orientations to produce the final output.

\begin{minipage}{\linewidth}
    \centering
    \includegraphics[width=1\textwidth]{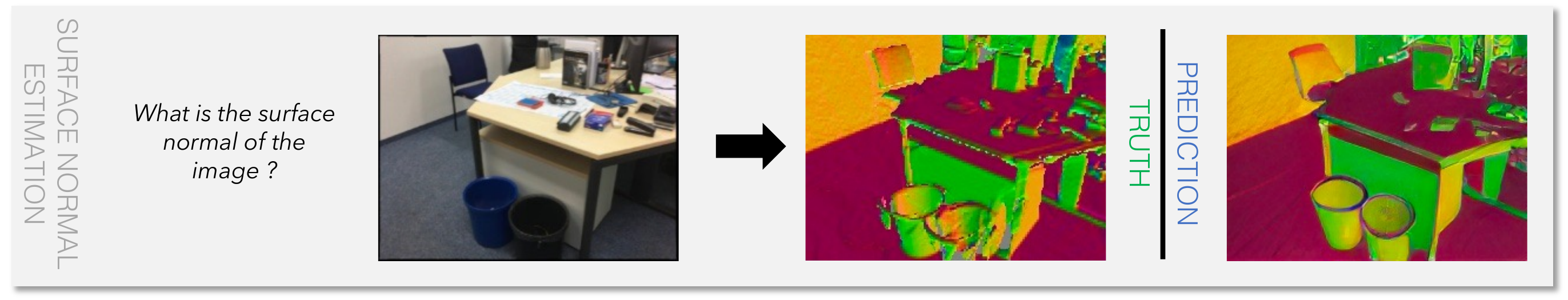}
    \label{fig:tasks_depth}
\end{minipage}

\subsubsection{Image Classification Tasks}

\boldheader{Image Classification}
\uio~is trained on 6 image classification datasets: ImageNet 2012~\citep{imagenet_cvpr09}, ImageNet21k \citep{ridnik2021imagenet}, Places365 \citep{zhou2017places}, Sun397 \citep{xiao2010sun}, iNaturalist \citep{van2018inaturalist} and Caltech birds 2011 \citep{WelinderEtal2010}.
For this task the input is an image and a static prompt, and the output is a class name. During inference we compute the log-probability of each class label in the dataset being evaluated and return the highest scoring one. This ensures~\uio~does not return a category from a different categorization dataset that is a synonym or hypernym of the correct label.

\begin{minipage}{\linewidth}
    \centering
    \includegraphics[width=1\textwidth]{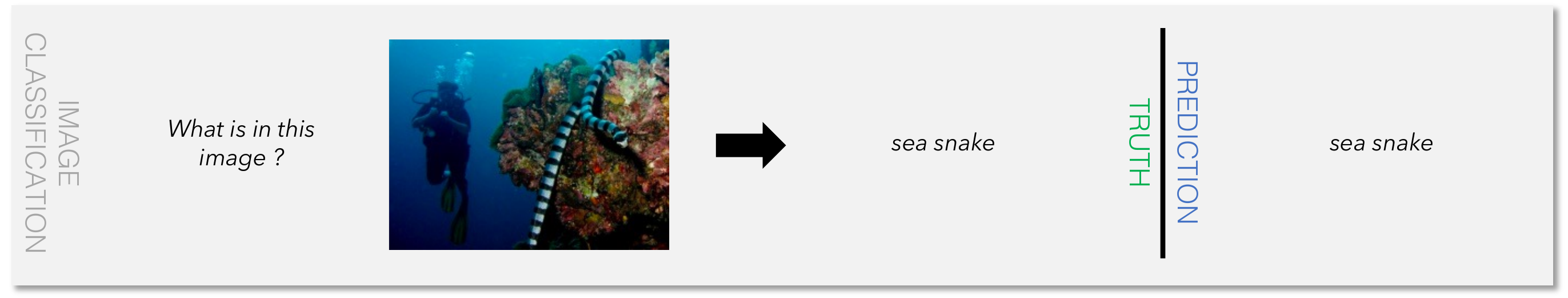}
    \vspace{-0.02in}
    \label{fig:tasks_depth}
\end{minipage}

\boldheader{Object Categorization} This task identifies which label, from a given set, best corresponds to an image region defined by an input image and bounding box. The input is the image, a prompt specifying the image region and the output is the target class name. We convert object detection annotations from Visual Genome, Open Images, and COCO for this task. Inference is constrained to return a valid label for the target label set just as with image classification.

\begin{minipage}{\linewidth}
    \centering
    \includegraphics[width=1\textwidth]{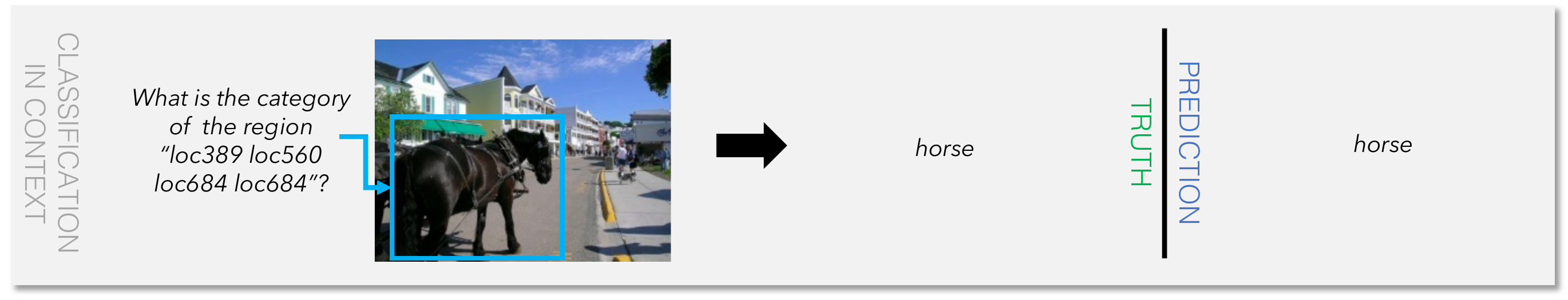}
    \label{fig:tasks_depth}
    \vspace{-0.02in}
\end{minipage}

\vspace{-0.02in}
\subsubsection{Image Captioning Tasks}
\boldheader{Image Captioning} 
Image captioning data comes from the same manually annotated and unsupervised sources used for Image Generation. In this case the inputs and output are reversed, the input is an image and the static prompt, and the output is a caption that matches the image.

\begin{minipage}{\linewidth}
    \centering
    \includegraphics[width=1\textwidth]{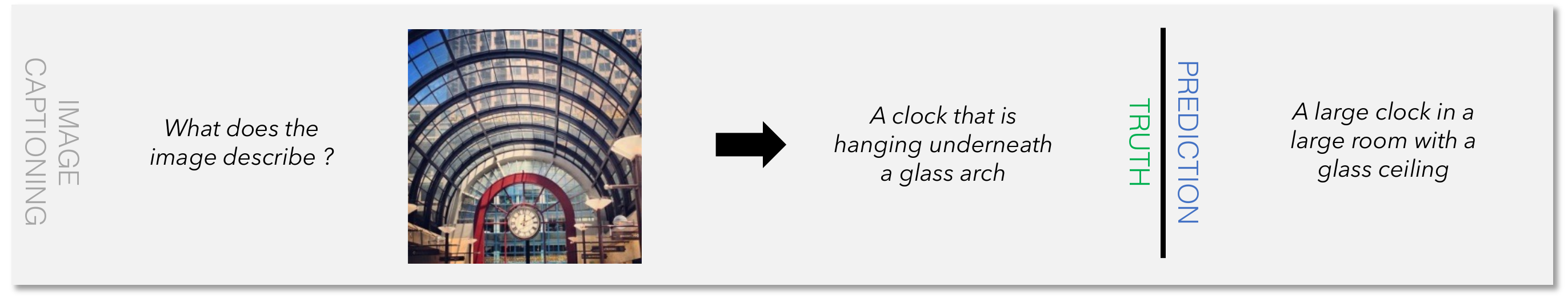}
    \label{fig:tasks_depth}
    \vspace{-0.02in}
\end{minipage}

\boldheader{Region Captioning} 
Region captioning tasks a model with generating a caption that describes a specific region in the image. Our format for this task is identical to Image Captioning except the region is included in the input prompt. Visual Genome~\citep{visual_genome} is used for the training data.

\begin{minipage}{\linewidth}
    \centering
    \includegraphics[width=1\textwidth]{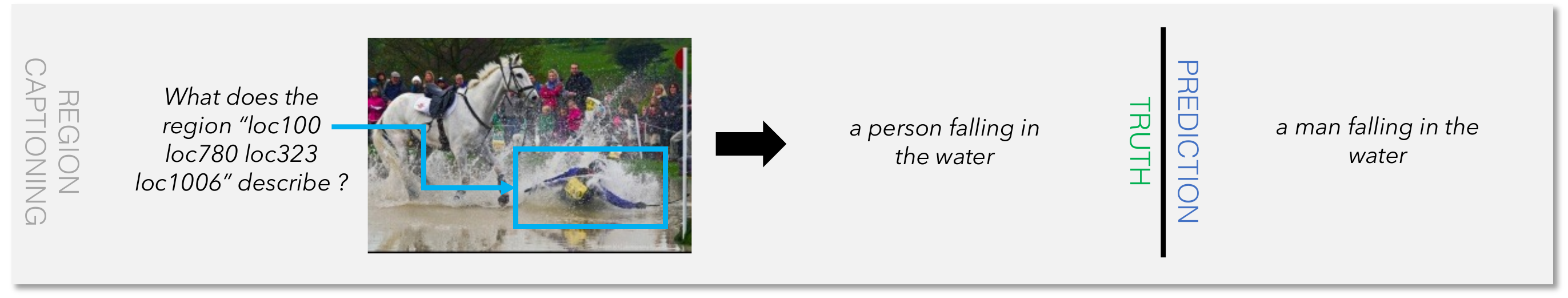}
    \label{fig:tasks_depth}
    \vspace{-0.02in}
\end{minipage}

\vspace{-0.02in}
\subsubsection{Vision \& Language Tasks}

\boldheader{Visual Question Answering} 
\uio~ is trained on a collection of VQA datasets including VQA 2.0~\citep{balanced_vqa_v2}, Visual Genome, VizWizVQA~\citep{vizwiz_vqa}, OKVQA~\citep{okvqa} and A-OKVQA~\citep{schwenk2022okvqa}. For VQA, the question is used as the prompt, and the output is the answer text. For VQA, it is common to constrain the model to predict an answer from a fixed last of common VQA answers~\citep{wang2022OFA,wang2021simvlm} during inference, but we avoid doing this since we find it does not benefit \uio\ in practice.

We additionally convert data from several other datasets in a VQA format, including imSitu~\citep{yatskar2016situation}, where we treat predicting the verb and then the related slots as separate VQA questions, VisualCOMMET~\citep{park2020visualcomet} where we convert the before/after/intent into questions by converting the input regions into location tokens, SNLI-VE~\citep{xie2019visual} where we integrate the entailed text into an input question, and VCR~\citep{zellers2019recognition} where we again integrate the input regions into the prompt by encoding them with location tokens and integrate the rationales into the target text for the answer justification task.

\begin{minipage}{\linewidth}
    \centering
    \includegraphics[width=1\textwidth]{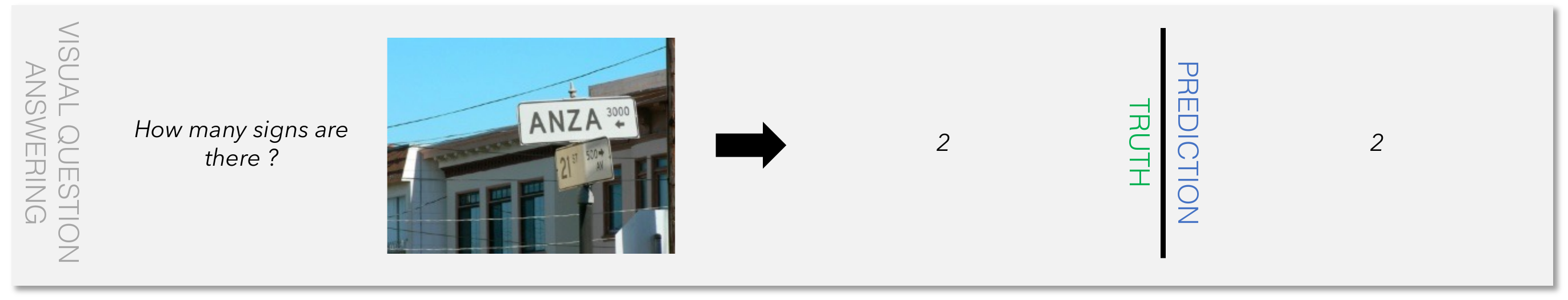}
    \label{fig:tasks_depth}
    \vspace{-0.02in}
\end{minipage}

\boldheader{Answer-Grounded Visual Question Answering} 
This task requires both answering a question and returning a binary mask specifying the region of the image used to answer the question. The format for this task follows the one for VQA except that a binary mask is also used as an additional output. Training data comes from VizWiz-VQA~\citep{whiz_viz_answer_grounded_vqa}, a dataset designed to train models that could benefit people with visual impairments.

\begin{minipage}{\linewidth}
    \centering
    \includegraphics[width=1\textwidth]{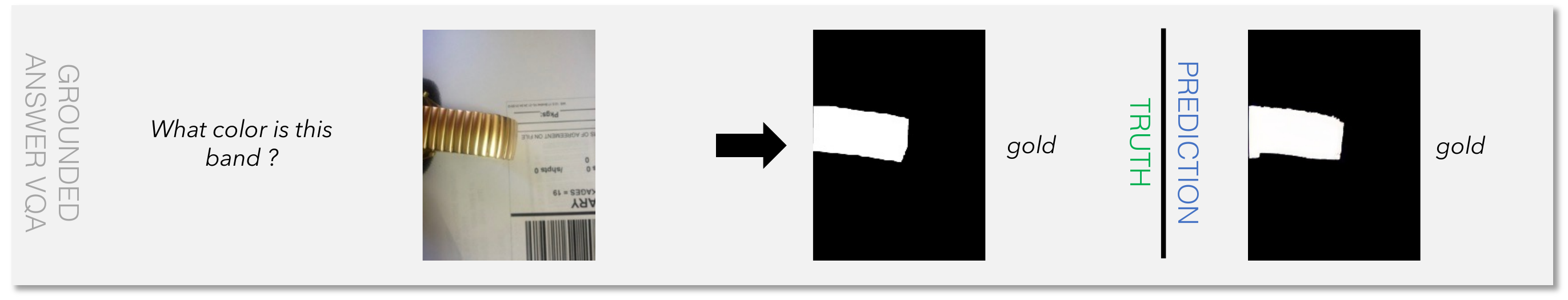}
    \label{fig:tasks_depth}
    \vspace{-0.02in}
\end{minipage}

\boldheader{Relationship Detection} This task requires predicating a relationship between a pair of objects which are grounded by bounding boxes. The prompt contains both the object regions, and the output is the predicted predicate.
There are 2 datasets in this tasks: Visual Genome~\citep{visual_genome} and Open Images~\citep{open_images}.

\begin{minipage}{\linewidth}
    \centering
    \includegraphics[width=1\textwidth]{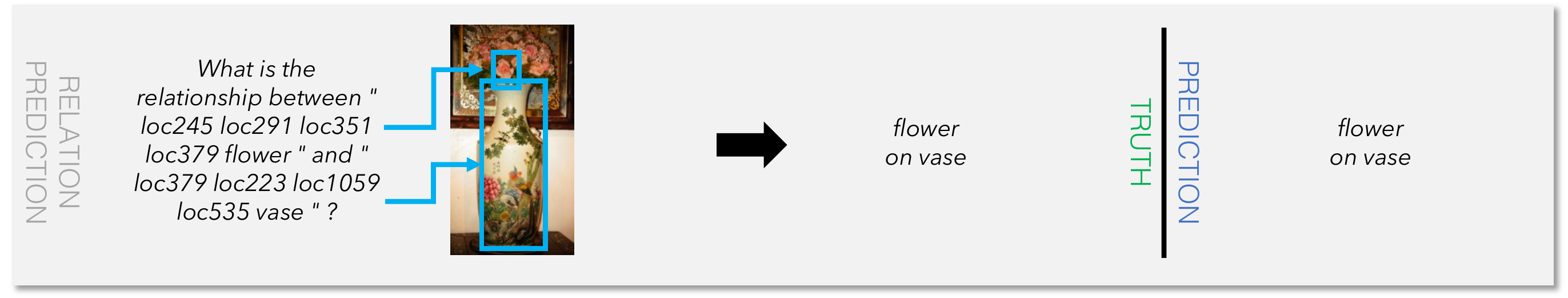}
    \label{fig:tasks_depth}
    \vspace{-0.02in}
\end{minipage}

\subsubsection{Natural Language Processing Tasks}

\boldheader{Question Answering} Following prior work in natural language processing~\citep{2020t5}, QA tasks are formatted by placing both the question and any text context (e.g., an paragraph containing the answer) into the prompt and training the model to generate the text answer. \uio\ is trained on several QA datasets including SQuAD 2.0~\citep{rajpurkar2016squad}, other training datasets from the MRQA~\citep{fisch2019mrqa} shared tasks~\citep{trischler-etal-2017-newsqa,joshi-etal-2017-triviaqa,dunn2017searchqa,yang-etal-2018-hotpotqa,kwiatkowski-etal-2019-natural}, QA datasets from SuperGLUE~\citep{wang2019superglue, clark2019boolq, multirc, choice_of_plausible_alternatives}, Cosmos QA~\citep{huang2019cosmos}, OpenBookQA~\citep{mihaylov2018can}, and HellaSwag~\citep{zellers2019hellaswag}. If the text context is longer then our maximum sequence length we use a sliding-window approach following~\citet{devlin2018bert} which exposes the model to different windows of text from the context and returns the highest-confidence answer.

\begin{minipage}{\linewidth}
    \centering
    \includegraphics[width=1\textwidth]{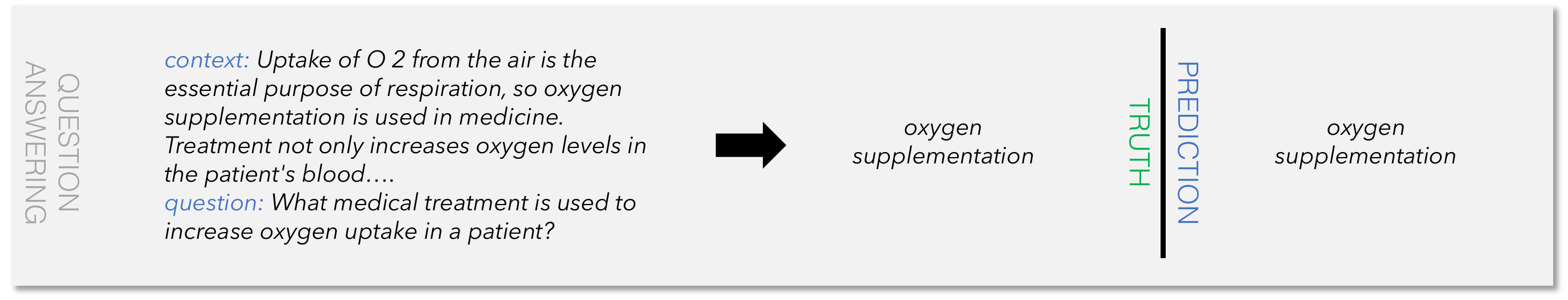}
    \label{fig:tasks_depth}
    \vspace{-0.02in}
\end{minipage}

\boldheader{Text Classification} Also following past work~\citep{2020t5}, text classification tasks are formatted by placing the input sentences and a query in the prompt and training the model to generate the target class. Datasets include tasks from GLUE and SuperGLUE~\citep{wang2018glue,wang2019superglue,cola,sst2,mrpc,qqp,stsb,mnli,rte1,rte2,rte3,rte5,levesque2012winograd,mnli,commitment_bank,wic}, as well as SNLI~\citep{bowman2015snli}, SciTail~\citep{khot2018scitail}, IMDB Reviews~\citep{imdb_reviews}, and PAWS~\citep{paws2019naacl}.

\begin{minipage}{\linewidth}
    \centering
    \includegraphics[width=1\textwidth]{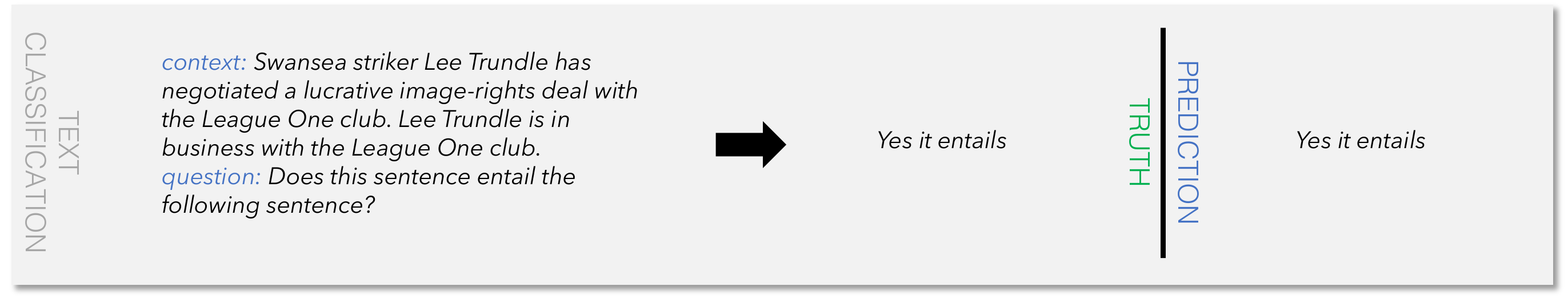}
    \label{fig:tasks_depth}
    \vspace{-0.02in}
\end{minipage}

\boldheader{Text Summerization}
Text summarization is done again by providing the input paragraph and a prompt as input and generating a summary as output. We use the Gigaword dataset~\citep{graff2003english,Rush_2015} for training data.

\begin{minipage}{\linewidth}
    \centering
    \includegraphics[width=1\textwidth]{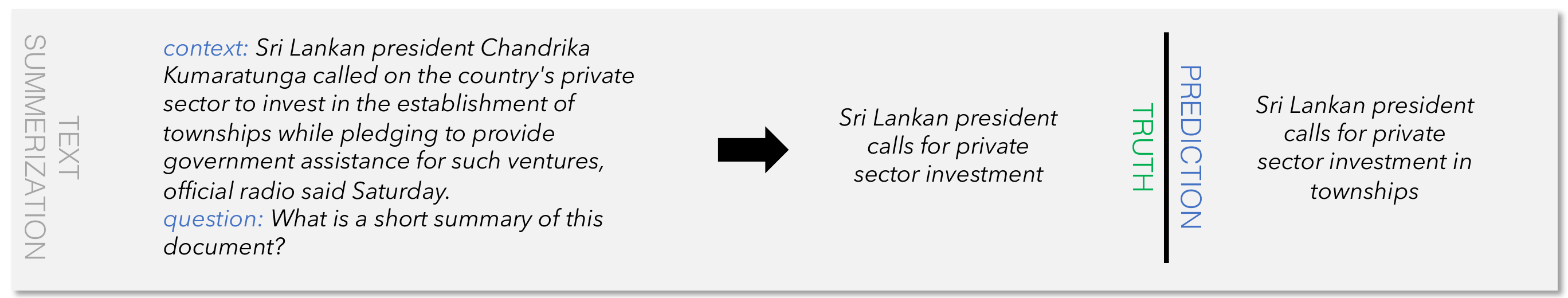}
    \label{fig:tasks_depth}
    \vspace{-0.02in}
\end{minipage}

\subsubsection{Language Modeling Tasks}

\boldheader{Mask Language Modeling} 
Following T5 \citep{2020t5}, the mask language modelling objective randomly samples and then drops out 15\% of tokens in the input sequence. All consecutive spans of dropped-out tokens are replaced by a single sentinel token. The target is to recover the dropped tokens given the sentinel token. We use C4 \citep{2020t5} and Wikipedia \citep{wikidump} datasets.

\begin{minipage}{\linewidth}
    \centering
    \includegraphics[width=1\textwidth]{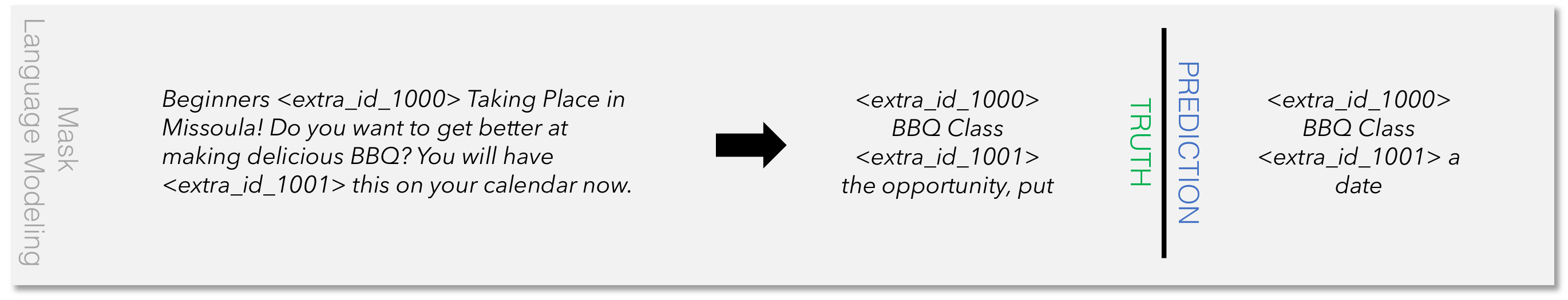}
    \label{fig:tasks_depth}
    \vspace{-0.02in}
\end{minipage}

\newpage

\subsection{Pre-Training Data Distribution}
\label{sect:pretraining_distribution}

\begin{figure}[!t]
    \centering
    \includegraphics[width=\textwidth]{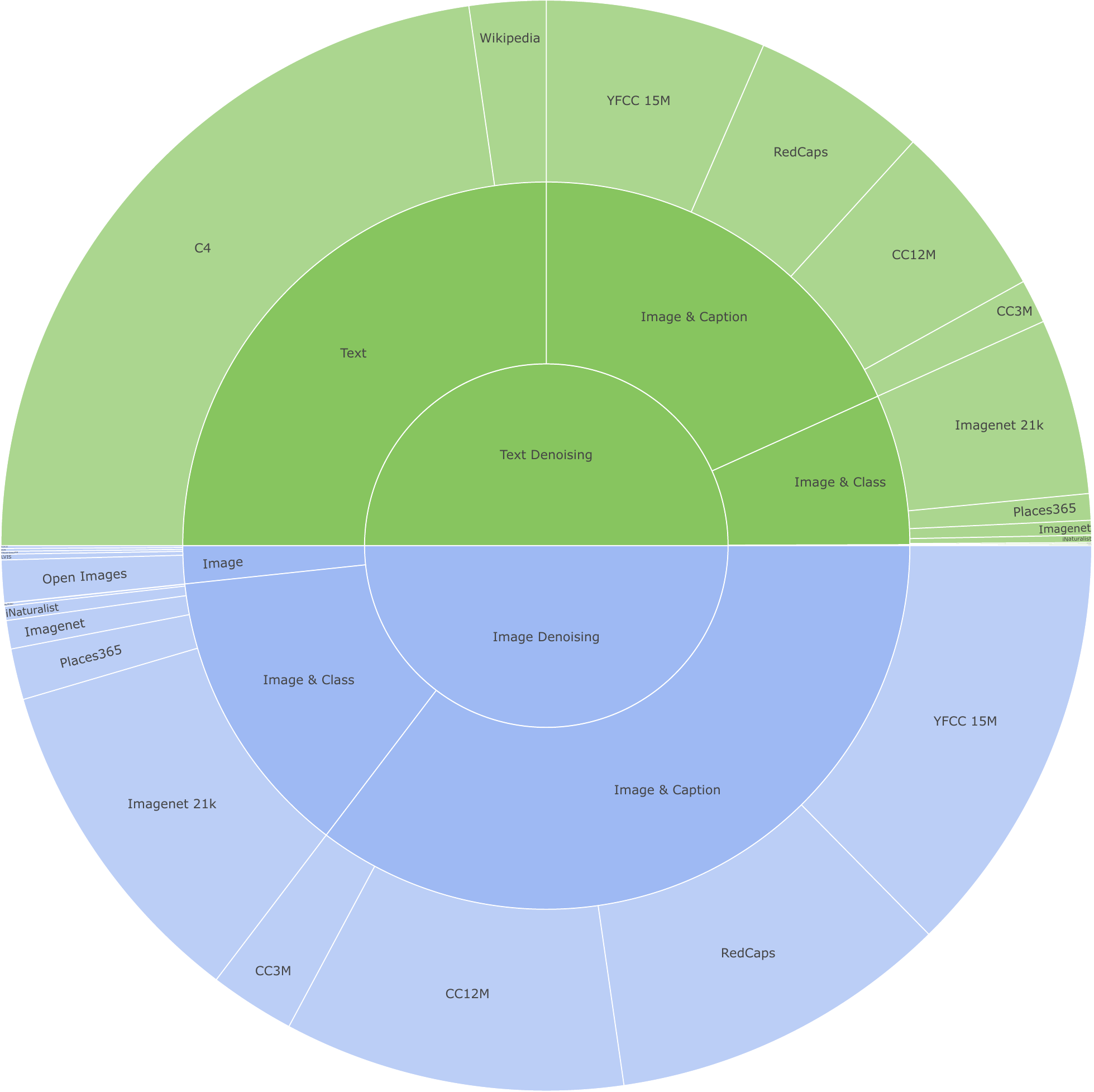}
    \caption{\small{Pre-training objectives (inner circle), annotation types (middle circle) and datasets (outer circle) used in pre-training of \uio. Sizes correspond to the sampling rate in the training distribution. Best viewed in color.}}
    \label{fig:pretrain_distribution}
\end{figure}

Figure~\ref{fig:pretrain_distribution} shows a visualization of pre-training data distribution used by \uio. As discussed in Section \ref{sect:training}, we equally sample data with the text denoising and image denoising objective (inner circle of Figure~\ref{fig:pretrain_distribution}). For text denoising, half of the samples are from pure text data, \ie C4 and Wikipedia. The other half is constructed from image and class, such as Imagenet21k \citep{ridnik2021imagenet} or image and caption, such as YFCC15M \citep{radford2021learning}. For image denoising, we use the text information when class and caption are present in the data source and sample the dataset proportional to the dataset size. For both text and image denoising, we random drop both modalities with 10\% of the time if both text and image as inputs.  

\newpage

\subsection{Multi-Tasking Data Distribution}
\label{sect:multi_task_distribution}

\begin{figure}[!t]
    \centering
    \includegraphics[width=\textwidth]{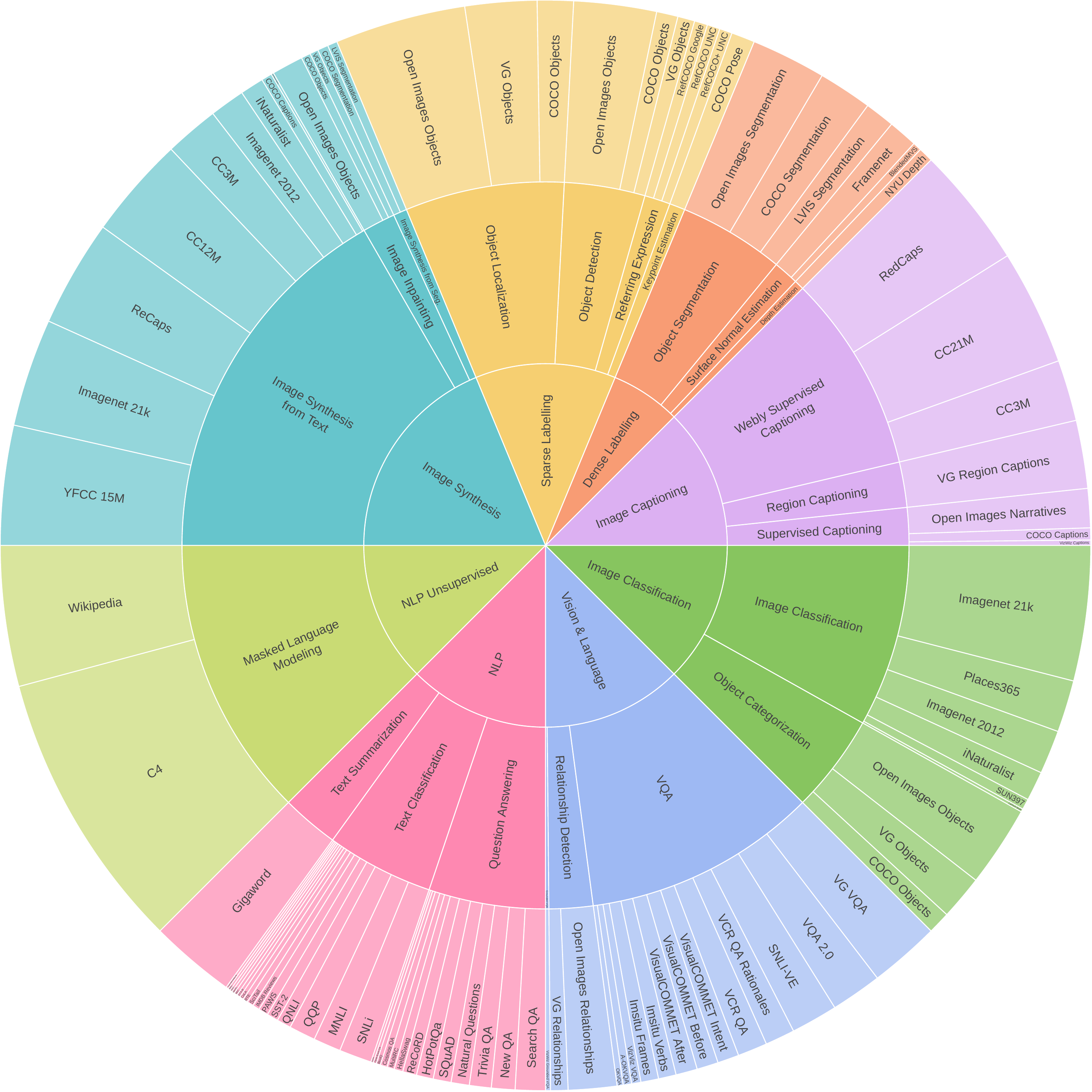}
    \caption{\small{Task groups (inner circle), tasks (middle circle) and datasets (outer circle) used in multi-task training of \uio. Sizes correspond to the sampling rate in the training distribution. Best viewed in color.}}
    \vspace{-0.2in}
    \label{fig:multi_task_distribution}
\end{figure}

Figure~\ref{fig:multi_task_distribution} shows a visualization of the multi-task training distribution used by \uio\ from Table~\ref{tab:tasks}. As discussed in Section \ref{sect:training}, we equally sample each group ($1/8$) except image synthesis ($3/16$) and dense labeling ($1/16$) since dense labeling has a much smaller sample size compared to image synthesis. We sample tasks and datasets (middle and outer circle) with a temperature-scaled mixing strategy to make sure the model is sufficiently exposed to underrepresented tasks. We raise each task’s mixing rate to the power of $1/T$ and then renormalize the rates so that they sum to 1. Following \cite{2020t5}, we use $T=2$ in our experiments. 

Due to the large variance in dataset size, some of the tasks are rarely sampled. For example, the depth estimation task only has the NYU Depth dataset source \citep{nyu_depth} and thus the sampling rate is only $0.43\%$. However, the model still works well for depth estimation tasks, even outperforming concurrent work \citep{kolesnikov2022uvim} (0.385 \vs 0.467 RMSE). We suspect the large model capacity and masked image denoising pre-training improves the performance. Similarly, Grounding VQA \citep{whiz_viz_answer_grounded_vqa} has $0.15\%$ sample rate, but the model can still achieve state-of-the-art performance on this task partly because it is trained on many related datasets for VQA and segmentation.

\newpage

\subsection{Qualitative Examples}
\label{sect:qual_examples}
Here we present qualitative examples of predictions from \uio\ for all training tasks. For brevity, if prompts are identical for each example we only show the prompt once, and if the prompt follows the same template for each example we show the template with parts that would be substituted with different words or location tokens underlined, and then show just the substitution with individual examples.

\begin{figure}
    \centering
    \includegraphics[width=\textwidth]{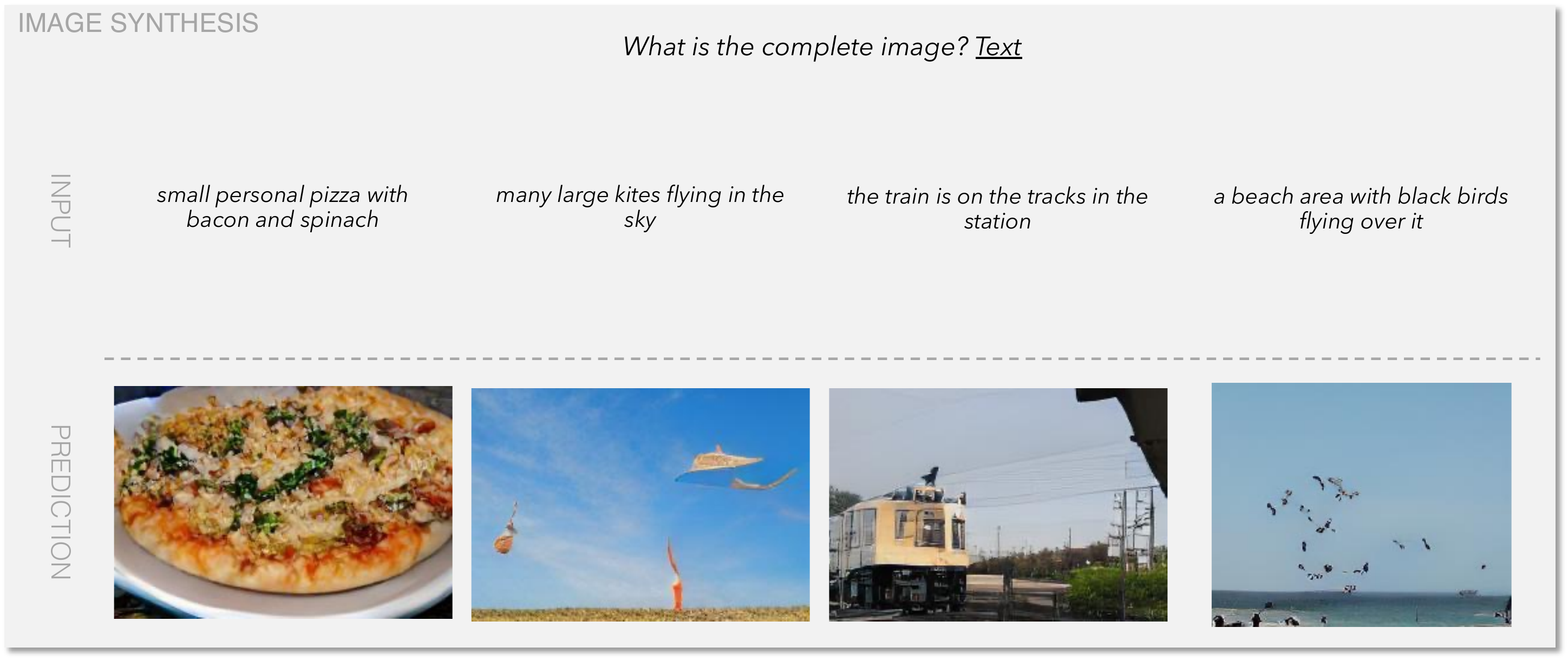}
    \includegraphics[width=\textwidth]{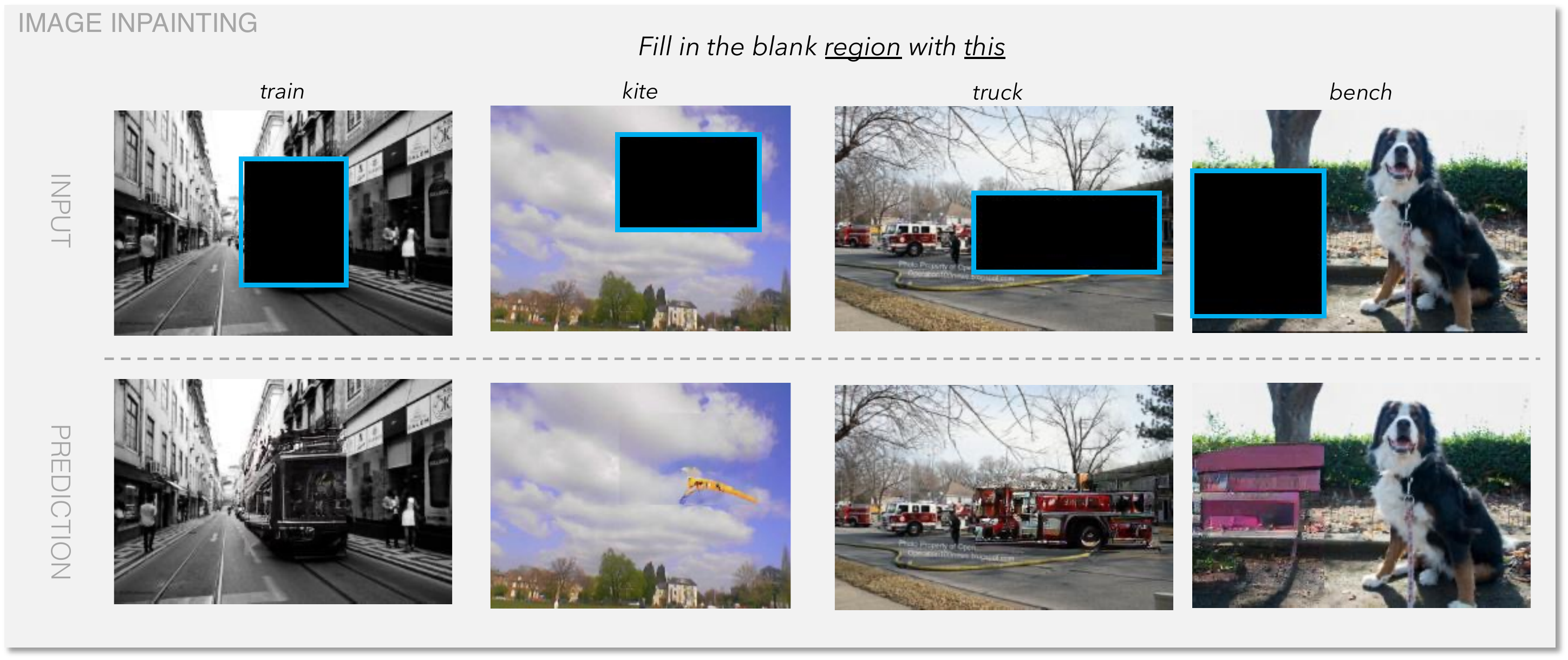}
    \includegraphics[width=\textwidth]{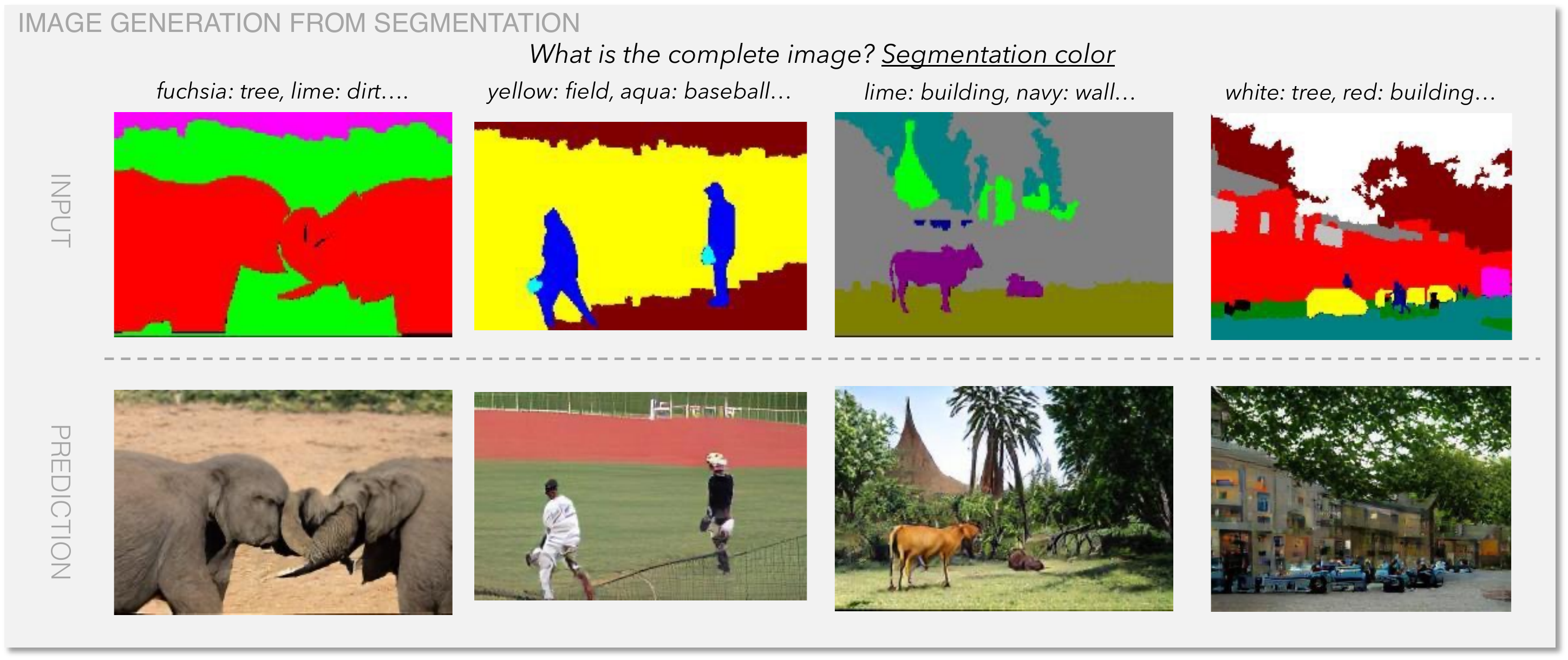}
    \caption{\small{Image synthesis qualitative examples.}}
\end{figure}

\begin{figure}
    \vspace{-0.35in}
    \centering
    \includegraphics[width=\textwidth]{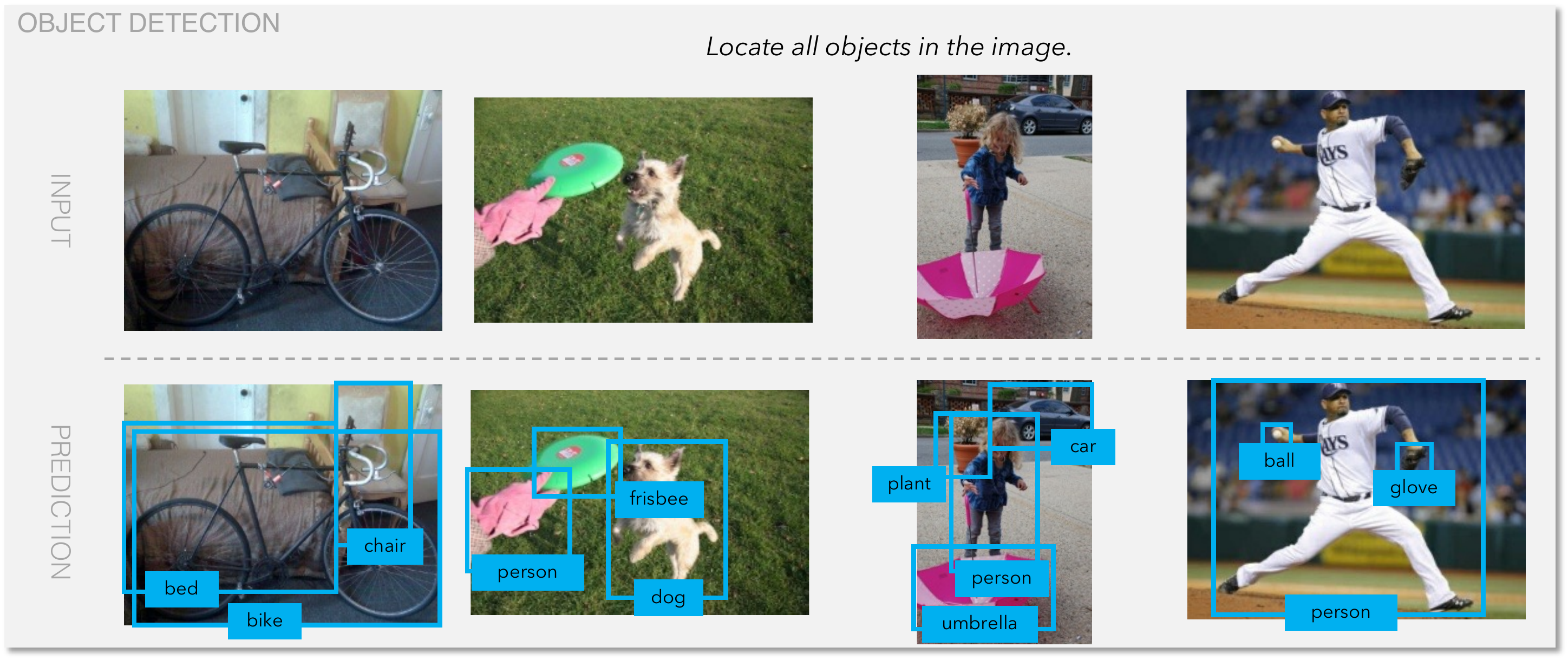}
    \includegraphics[width=\textwidth]{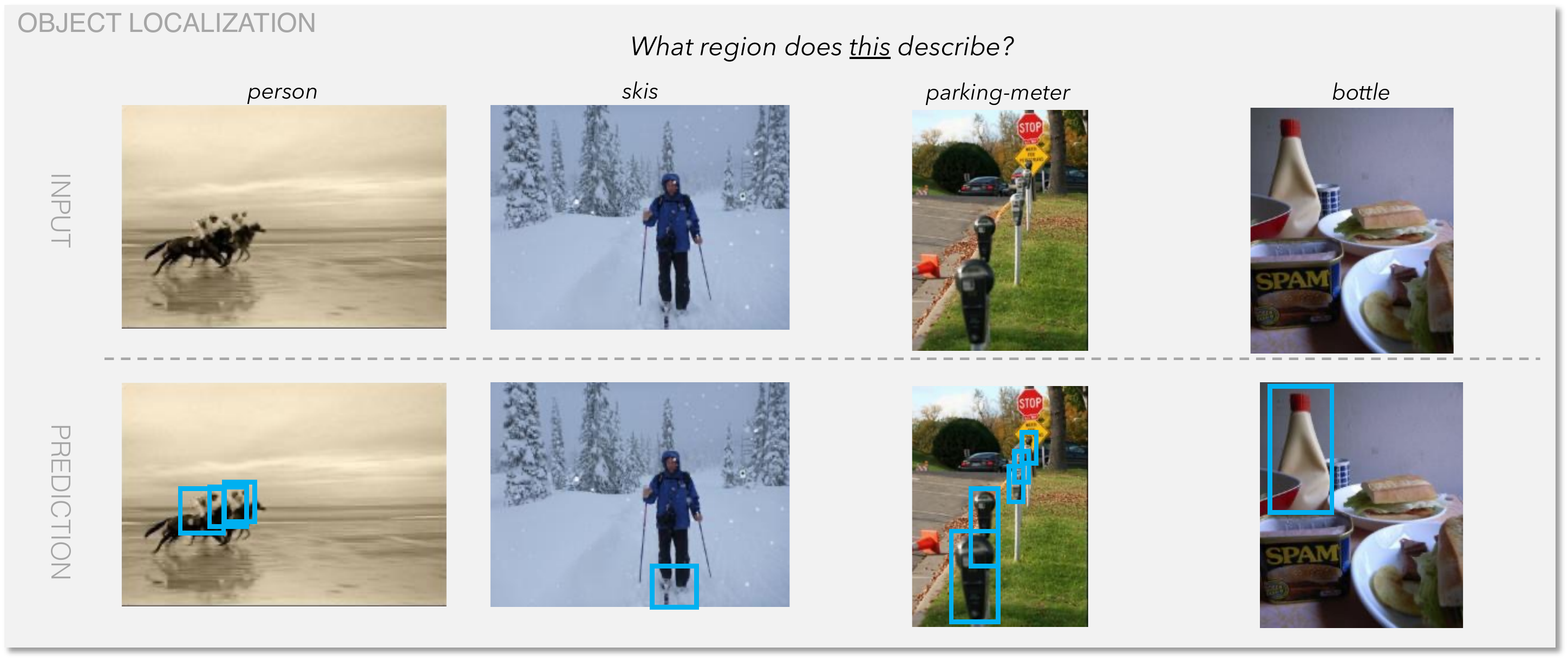}
    \includegraphics[width=\textwidth]{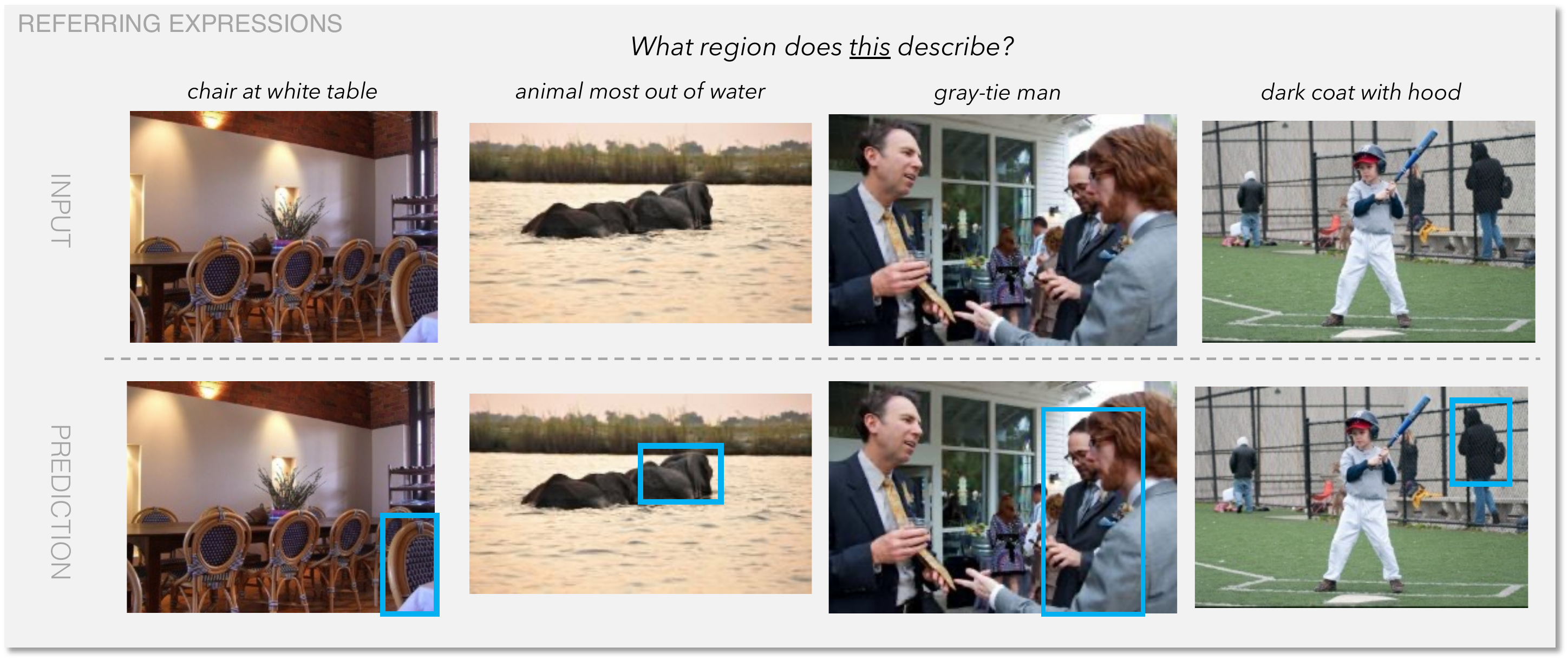}
    \includegraphics[width=\textwidth]{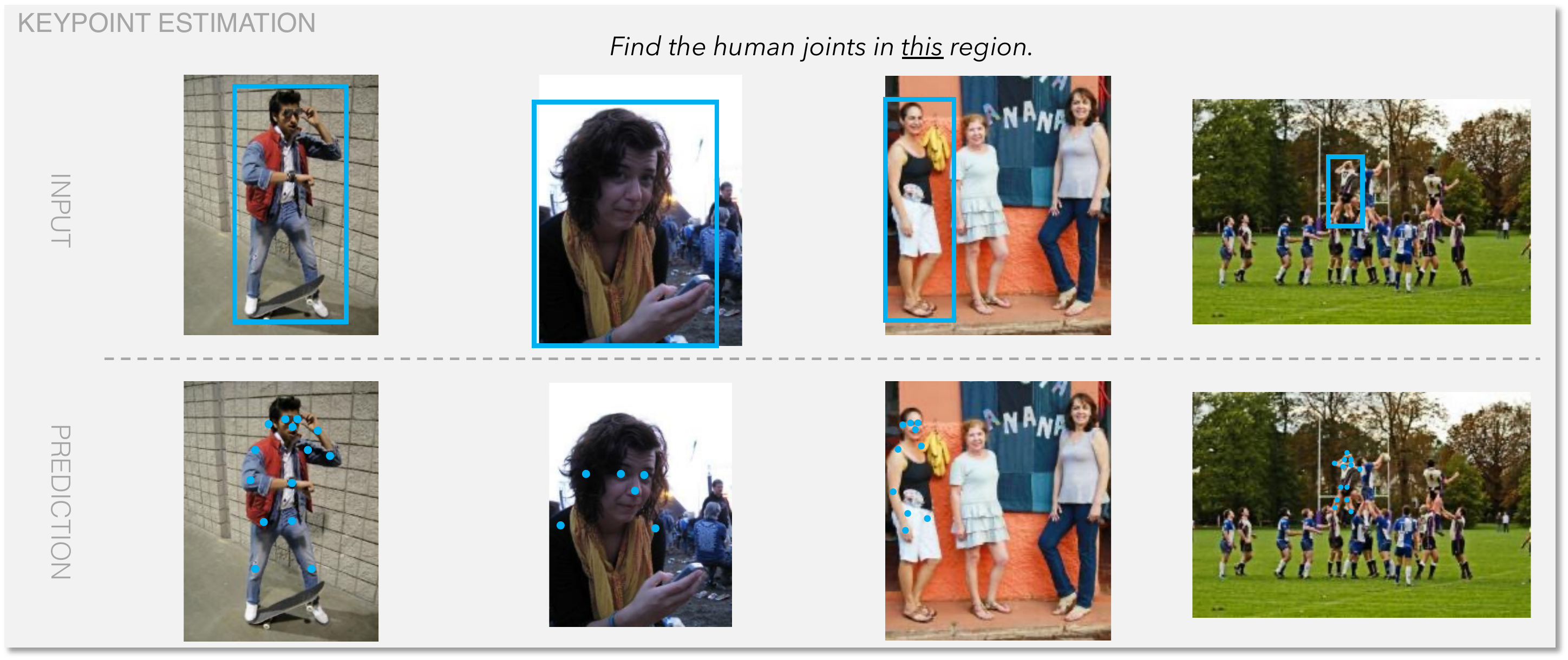}
    \caption{\small{Sparse labelling qualitative examples.}}
\end{figure}

\begin{figure}
    \centering
    \includegraphics[width=\textwidth]{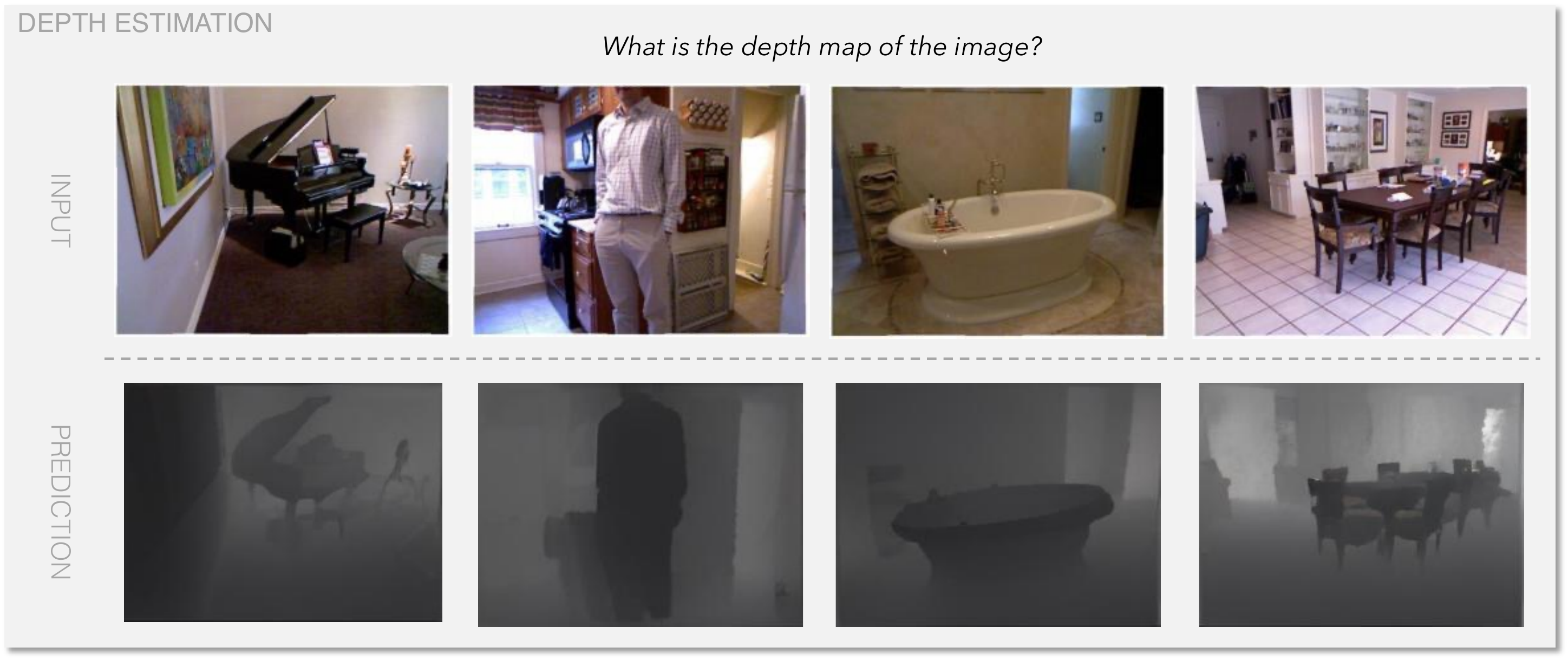}
    \includegraphics[width=\textwidth]{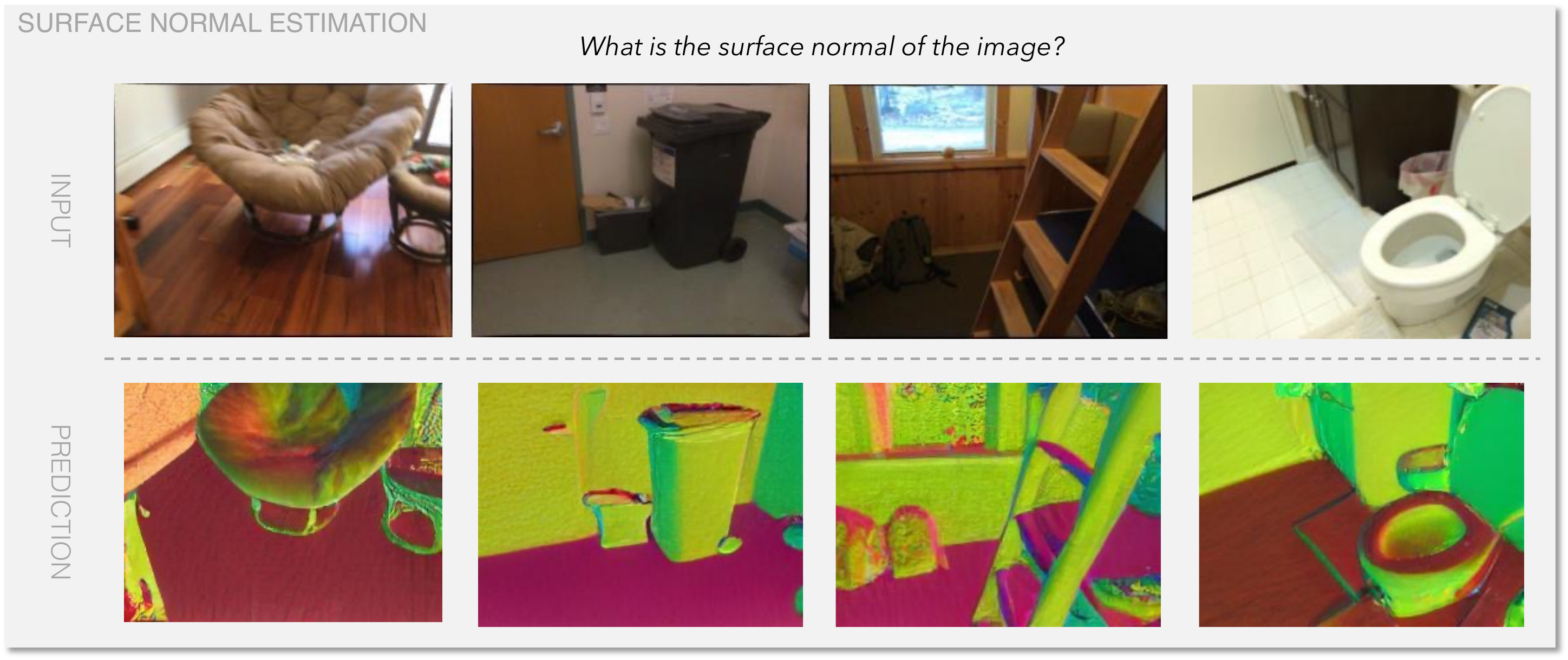}
    \includegraphics[width=\textwidth]{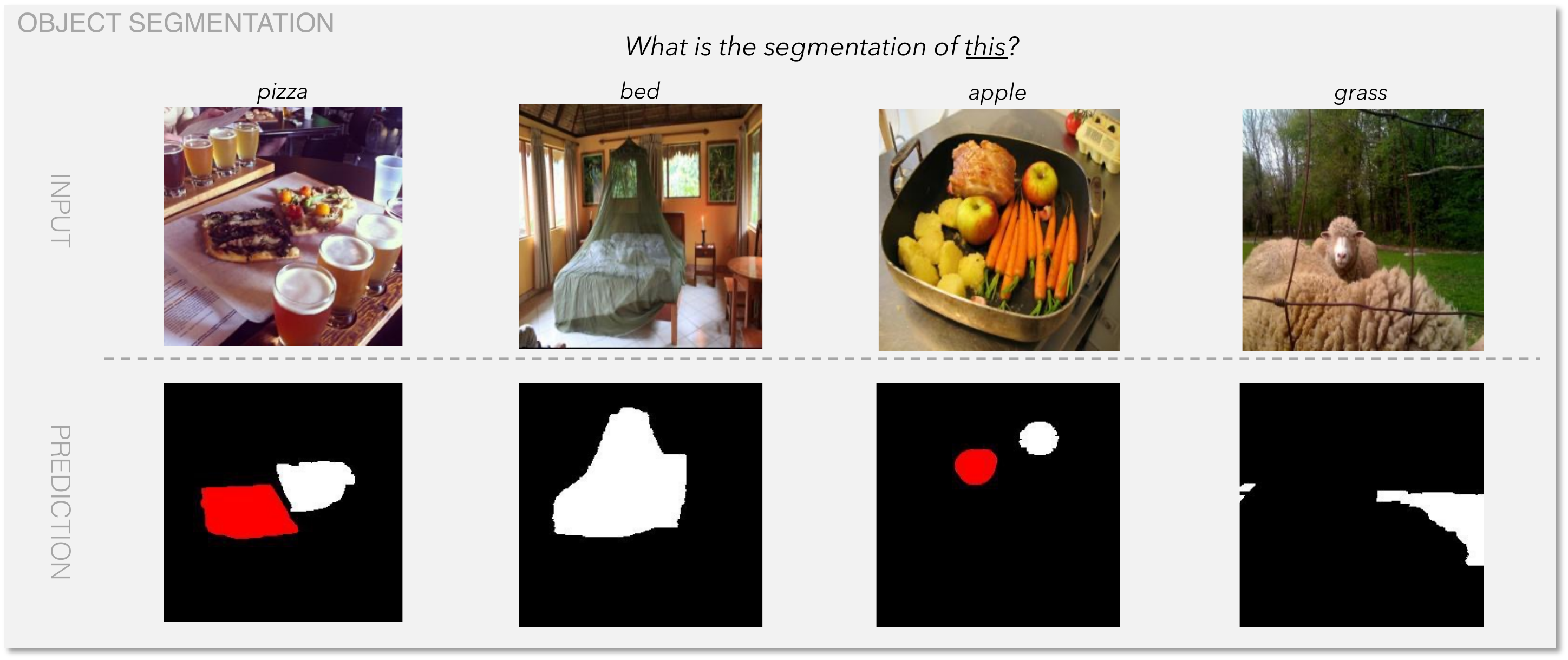}
    \caption{\small{Dense labelling qualitative examples.}}
\end{figure}

\begin{figure}
    \centering
    \includegraphics[width=\textwidth]{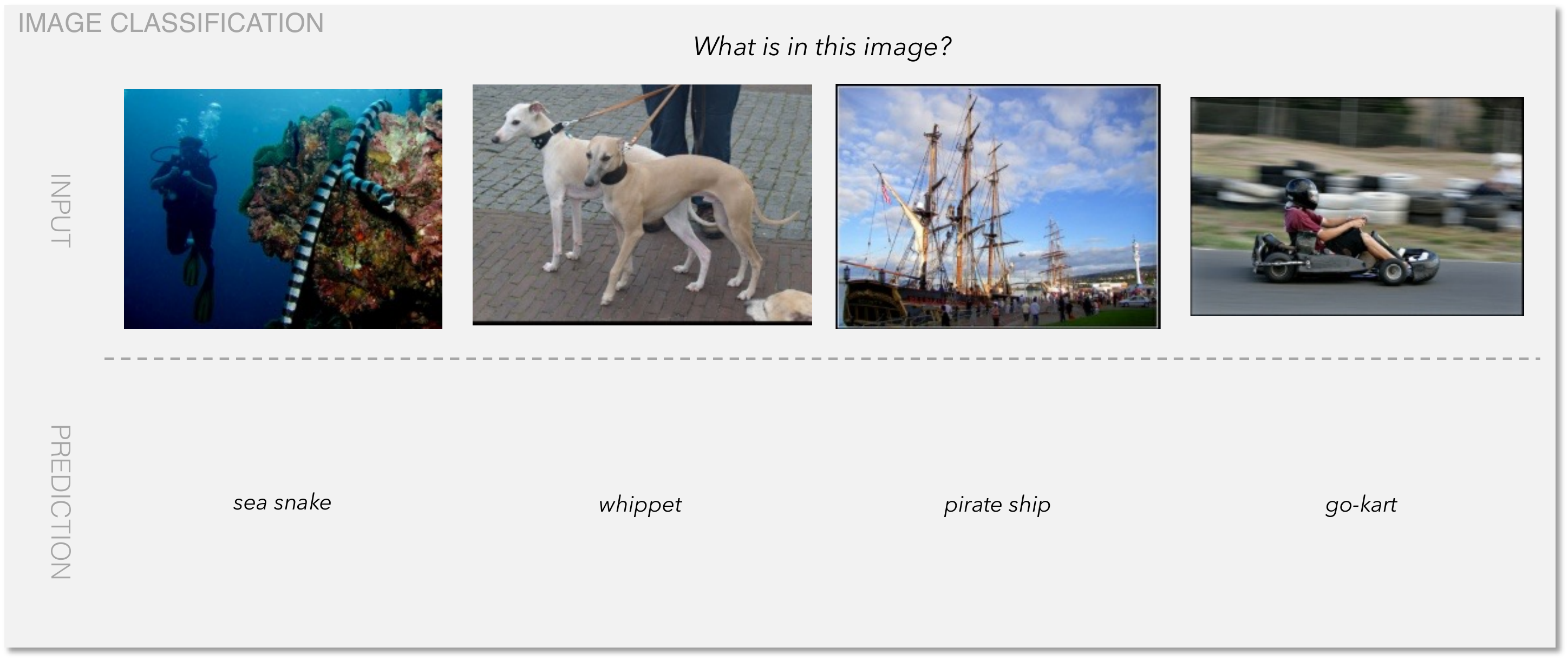}
    \includegraphics[width=\textwidth]{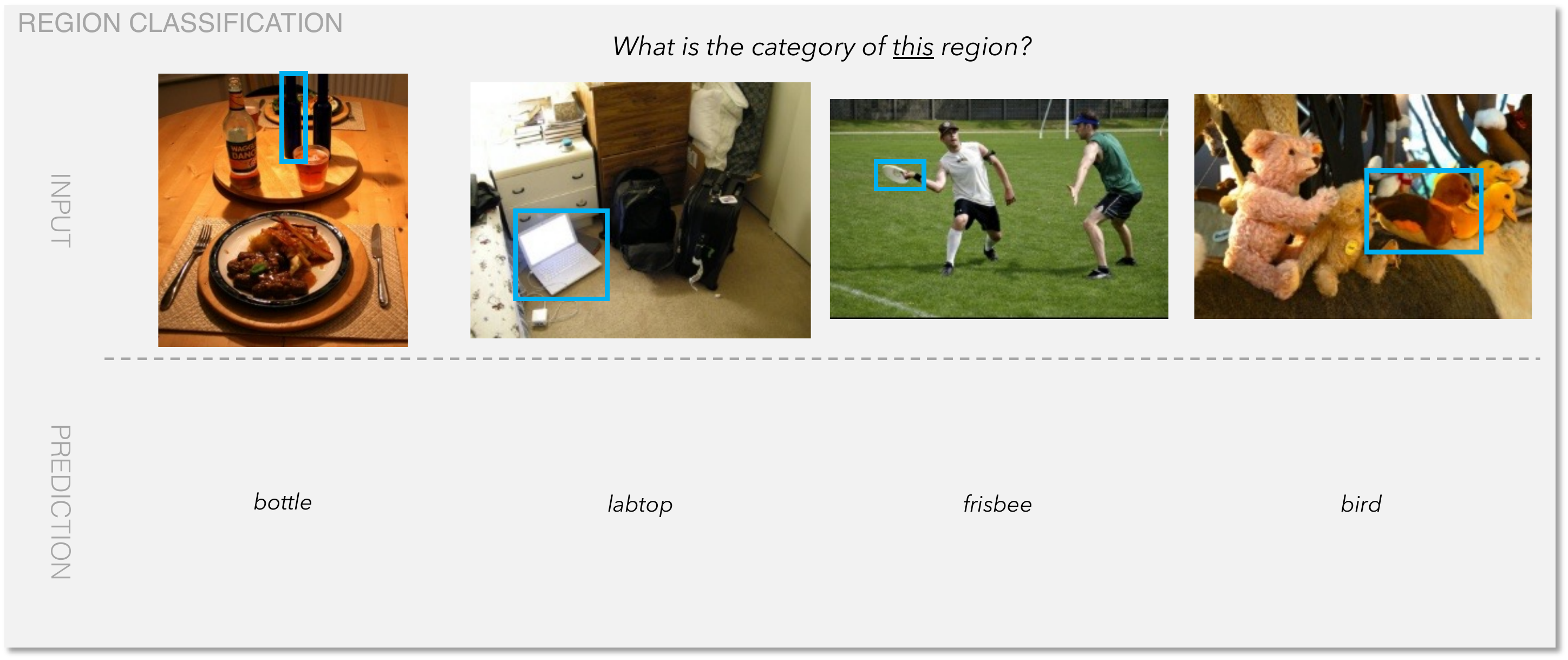}
    \caption{\small{Image classification qualitative examples.}}
\end{figure}

\begin{figure}
    \centering
    \includegraphics[width=\textwidth]{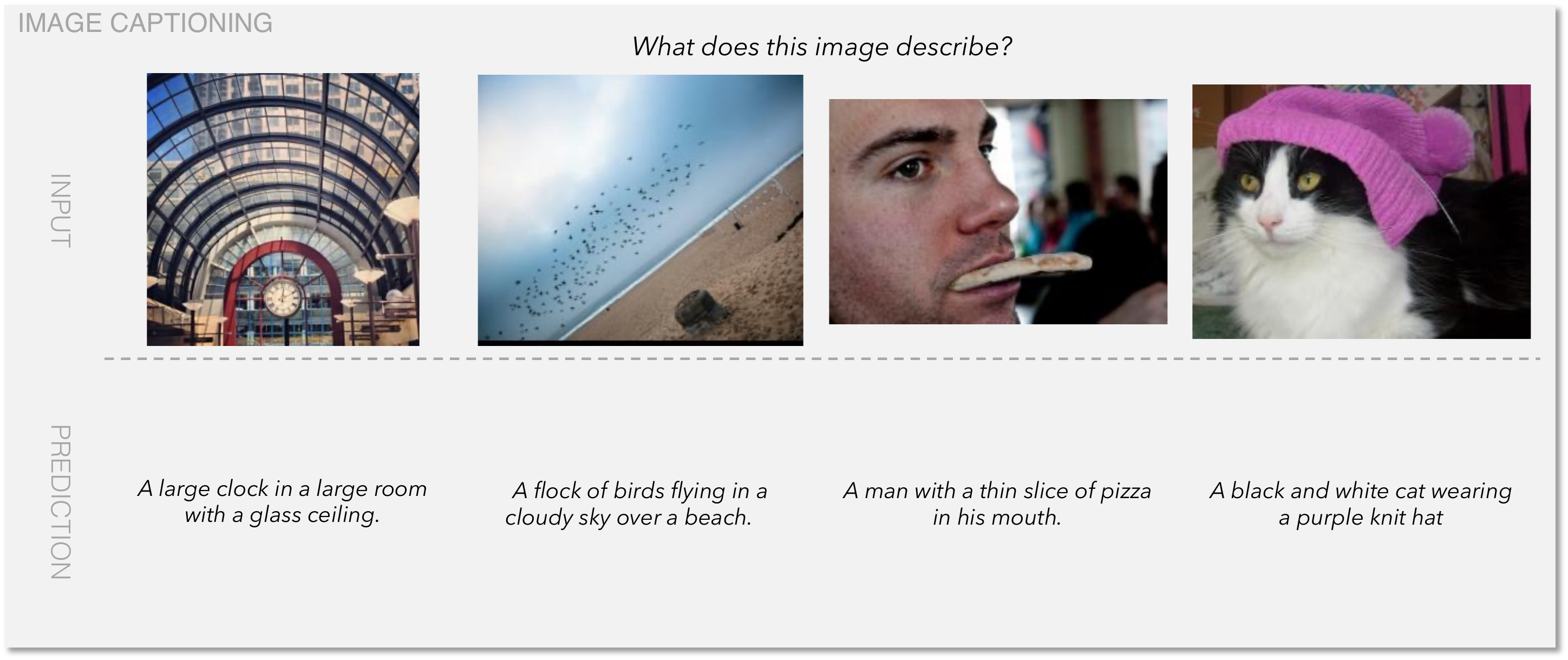}
    \includegraphics[width=\textwidth]{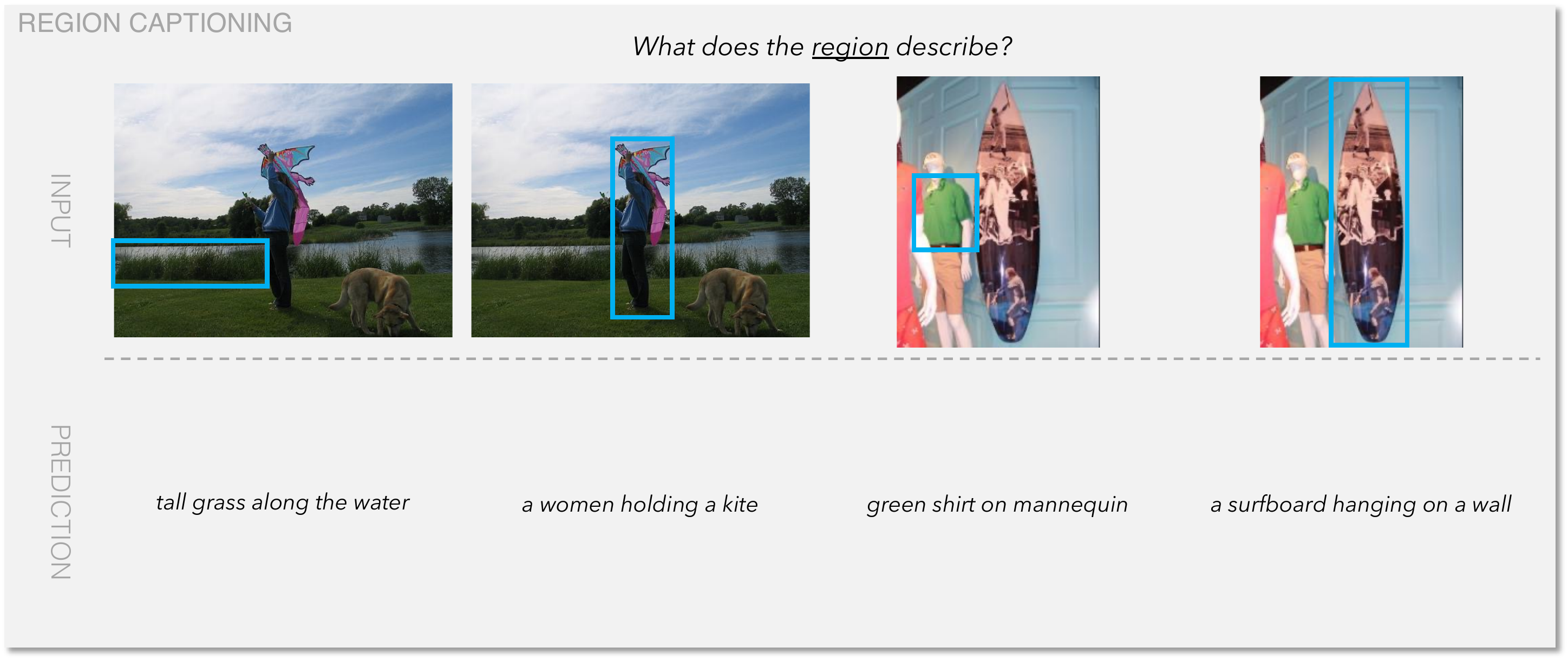}
    \caption{\small{Image captioning qualitative examples.}}
\end{figure}

\begin{figure}
    \centering
    \includegraphics[width=\textwidth]{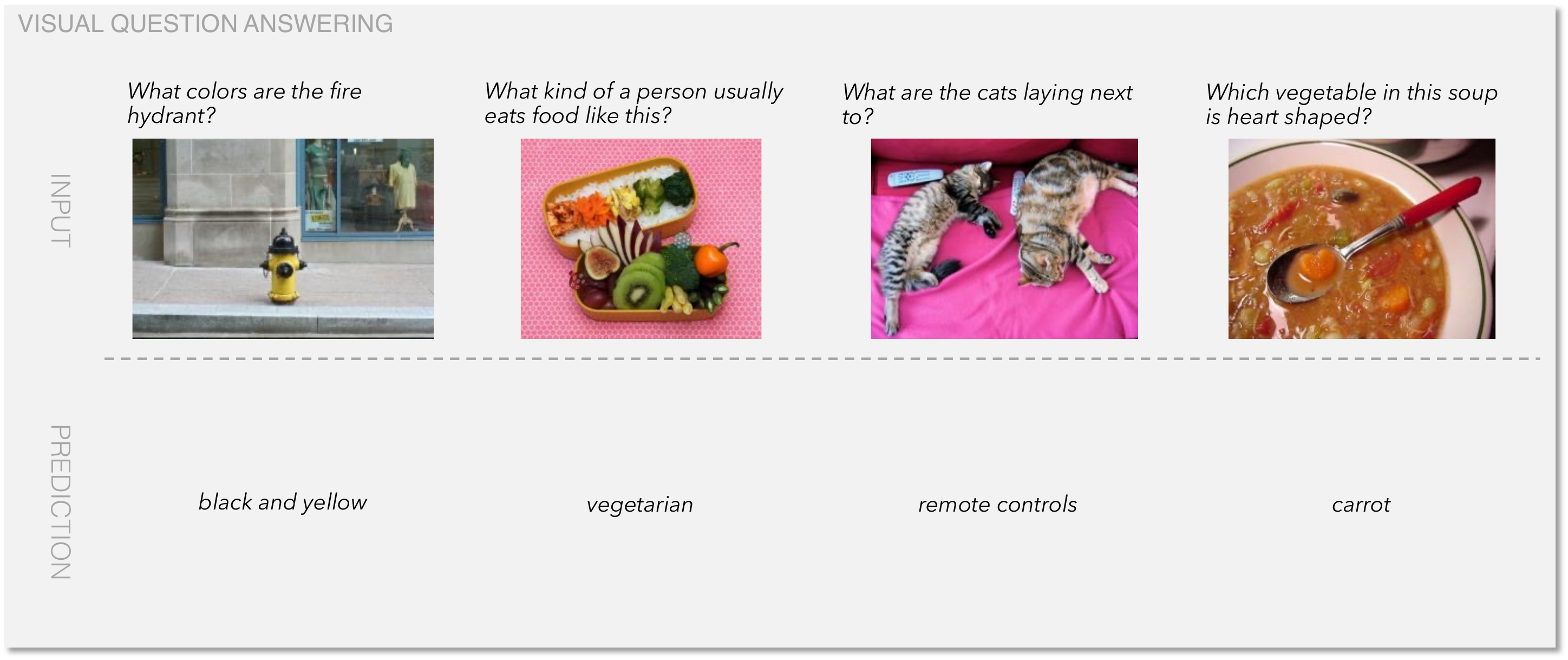}
    \includegraphics[width=\textwidth]{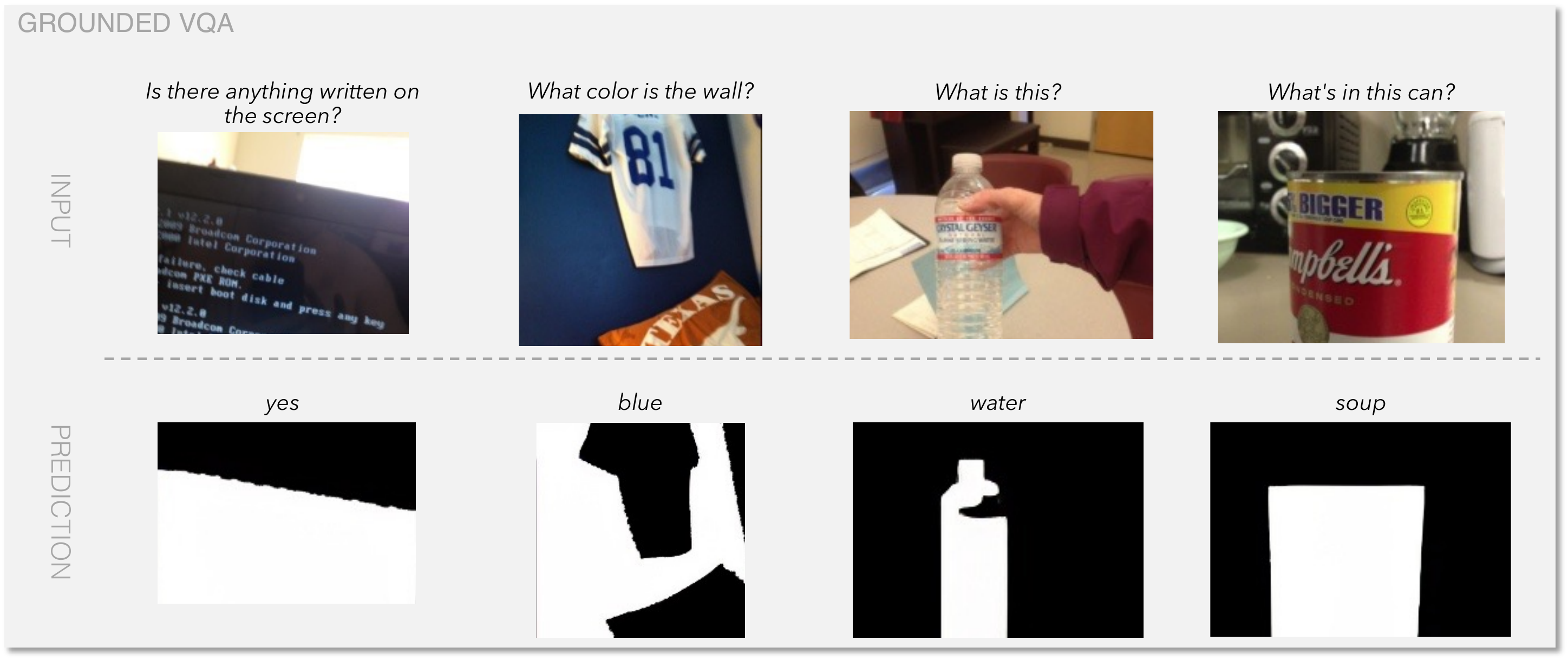}
    \includegraphics[width=\textwidth]{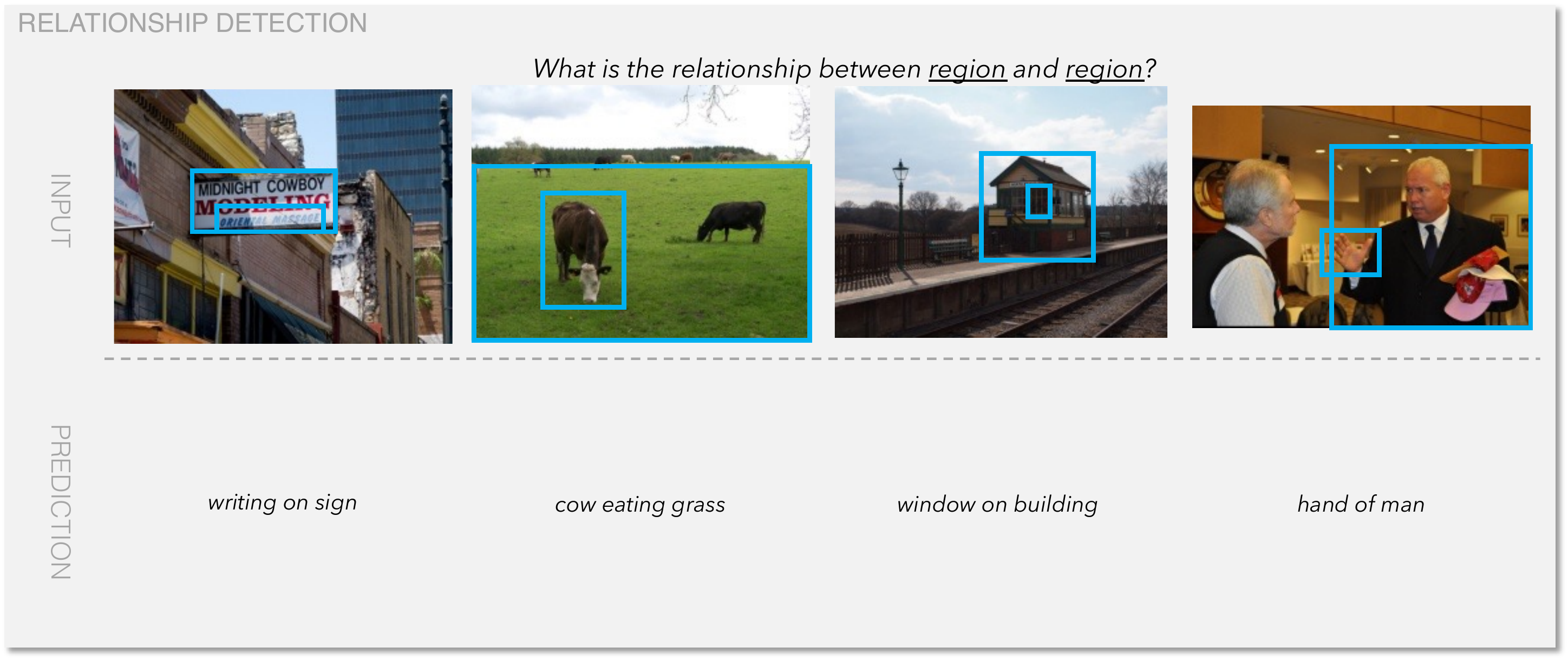}
    \caption{\small{Vision and language qualitative examples.}}
\end{figure}

\begin{figure}
    \centering
    \includegraphics[width=\textwidth]{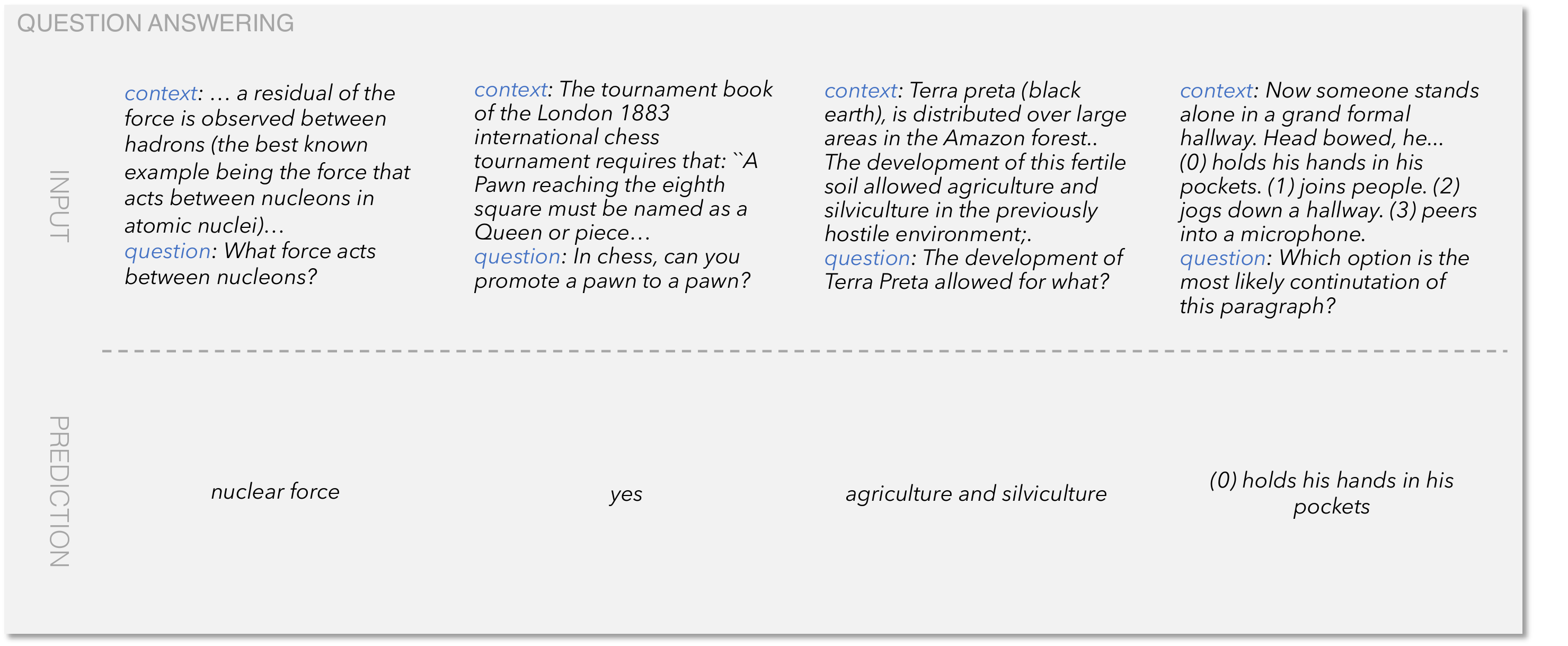}
    \includegraphics[width=\textwidth]{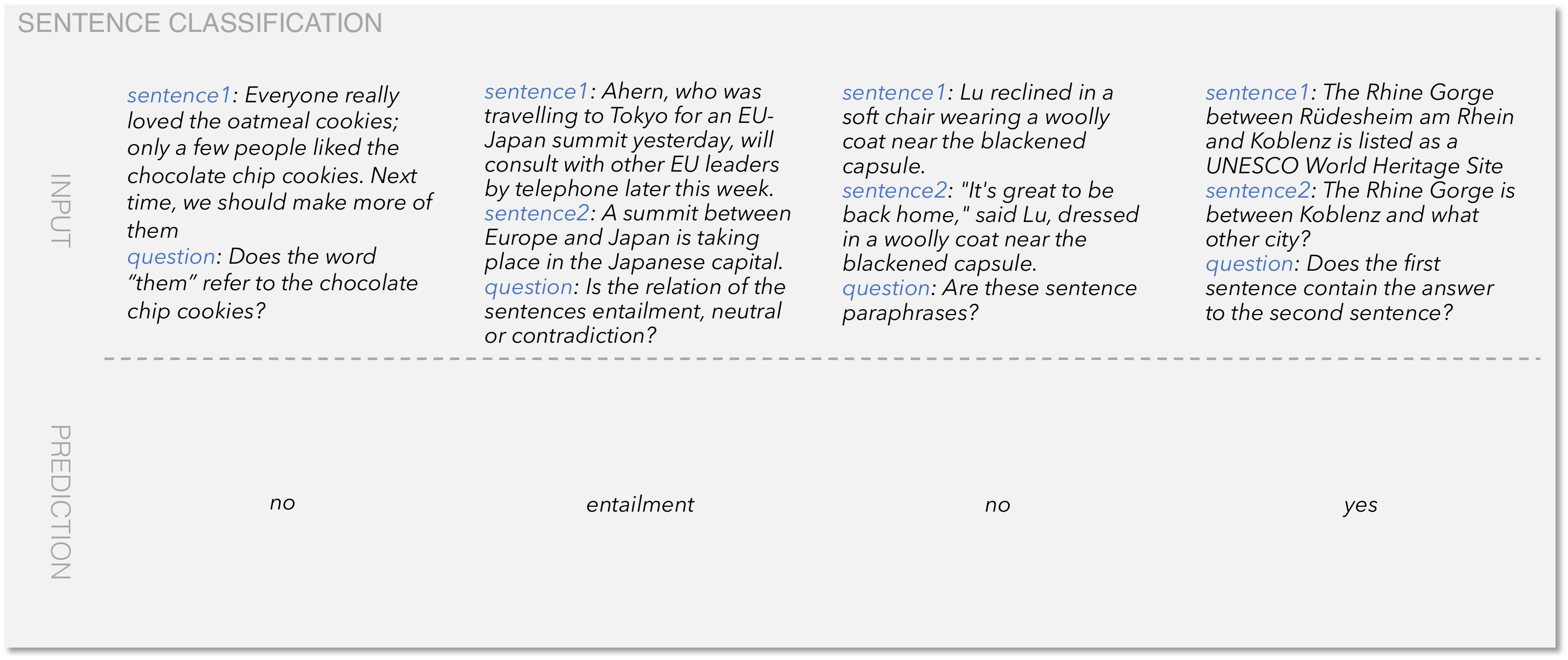}
    \includegraphics[width=\textwidth]{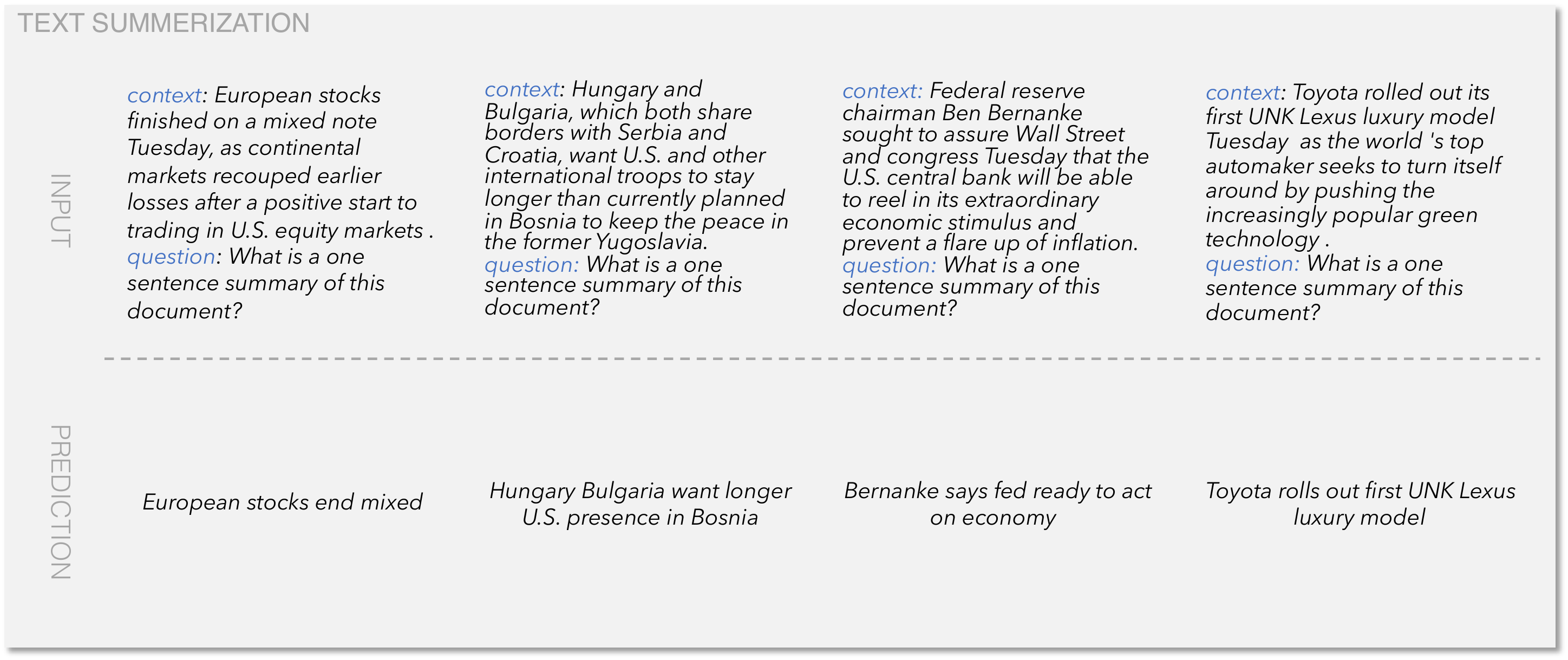}
    \caption{\small{Natural language processing  qualitative examples.}}
\end{figure}

    \centering



\end{document}